\documentclass[letterpaper]{article}
\usepackage[a4paper, total={6.5in, 10in}]{geometry}
\usepackage{biblatex}
\usepackage{csquotes}
\usepackage[english]{babel}
\usepackage{graphicx}
\usepackage{framed}

\usepackage{amsmath}
\usepackage{amssymb}
\usepackage{amsthm}
\usepackage[pro]{fontawesome5}
\usepackage{amsfonts}
\usepackage{color}
\usepackage{xcolor}

\usepackage{tikz}
\usetikzlibrary{shapes, arrows, decorations.pathreplacing, calligraphy, automata, positioning, fit, petri, backgrounds, calc, arrows.meta}
\usepackage{subcaption}

\usepackage[noend]{algpseudocode}
\usepackage{algorithm}

\usepackage{tabularx}
\usepackage{booktabs}
\usepackage{multirow}
\usepackage{longtable}
\usepackage{arydshln}
\usepackage{multicol}

\usepackage{enumitem}
\usepackage{url}
\usepackage{hyperref}
\usepackage{cleveref}
\usepackage[acronym, toc]{glossaries}
\usepackage{lscape}
\usepackage{pgfplots}
\usepackage{pgf-pie}
\usepackage{tikz-cd}
\usepackage{listings}

\pgfplotsset{compat=1.18}

\usetikzlibrary{intersections}
\newcounter{algline}
\newcommand{\lline}{\refstepcounter{algline}}

\definecolor{col1}{rgb}{0.283187, 0.125848, 0.44496}
\definecolor{textcol1}{gray}{1}
\definecolor{col2}{rgb}{0.227802, 0.326594, 0.546532}
\definecolor{textcol2}{gray}{1}
\definecolor{col3}{rgb}{0.132268, 0.655014, 0.519661}
\definecolor{textcol3}{gray}{0}
\definecolor{col4}{rgb}{0.404001, 0.800275, 0.362552}
\definecolor{textcol4}{gray}{0}
\definecolor{col5}{rgb}{0.83527, 0.886029, 0.102646}
\definecolor{textcol5}{gray}{0}

\newcommand*{\Truck}{\texttt{Truck}}
\newcommand*{\Item}{\texttt{Item}}
\newcommand*{\Package}{\texttt{Package}}
\newcommand*{\Cell}{\texttt{Cell}}

\newcommand{\CellVar}{c}
\newcommand{\TruckVar}{tr}
\newcommand{\LocationVar}{loc}
\newcommand{\LocationVarPrime}{\LocationVar'}
\newcommand{\PackageVar}{pk}
\newcommand{\PackageVarPrime}{pk'}
\newcommand{\Delivery}{\texttt{Delivery}}
\newcommand{\Childsnack}{\texttt{ChildSnack}}
\newcommand{\Ferry}{\texttt{Ferry}}

\newcommand*{\planningTask}{\mathcal{T}}
\newcommand*{\Predicates}{\mathcal{P}}
\newcommand*{\Objects}{\mathcal{O}}
\newcommand*{\ObjectTypeSet}{\mathcal{O}^{\ObjectType}}
\newcommand*{\ObjectLetter}{o}
\newcommand*{\ObjectType}{\tau}
\newcommand*{\InitialState}{\mathcal{I}}
\newcommand*{\Actions}{\mathcal{A}}
\newcommand*{\Goal}{\mathcal{G}}
\newcommand*{\plan}{\pi}
\newcommand*{\stateVar}{s}
\newcommand*{\stateVarPrime}{\stateVar'}
\newcommand*{\InitialStateVar}{\Trajectory_\InitialStateVarIndex}
\newcommand*{\InitialStateVarIndex}{0}
\newcommand*{\FinalStateVar}{\Trajectory_\planLength}
\newcommand*{\planLength}{n}

\newcommand*{\Trajectory}{\mathbf{t}}
\newcommand*{\Domain}{\mathcal{D}}

\newcommand{\SearchHeuristic}{h}
\newcommand{\SearchPathCost}{g}
\newcommand{\SearchEvalFunction}{f}
\newcommand{\timeVar}{t}
\newcommand{\graph}{G}
\newcommand{\nodes}{V}
\newcommand{\edges}{E}
\newcommand{\node}{v}
\newcommand{\edge}{e}
\newcommand*{\edgeDef}[3]{#1 = (#2,#3)}

\newcommand*{\TrainingTrajectories}{\mathcal{T}}
\newcommand*{\LoopTrajectories}{\TrainingTrajectories_{\LoopSymbol}}
\newcommand*{\GoalPredicates}{\Predicates^\Goal}
\newcommand{\goalPredicateFunc}[1]{{#1}^{\Goal}}
\newcommand*{\DomainGoal}{\mathcal{{\Goal}_{\Domain}}}
\newcommand*{\GeneralizedLandmark}{L}
\newcommand*{\Accepts}[2]{{#1}\vDash{#2}}
\newcommand*{\NotAccepts}[2]{{#1}\mathrel{\not\vDash}{#2}}
\newcommand*{\Guide}{\lambda}
\newcommand*{\GuideDistance}{{\delta}_{\Guide}}
\newcommand*{\GuideSet}{\Lambda}
\newcommand*{\lmNodeIndexOne}{x}
\newcommand*{\lmNodeIndexTwo}{y}
\newcommand*{\lmNodeIndexThree}{z}
\newcommand*{\StateDescriptor}{d}
\newcommand*{\StateValue}{\phi}
\newcommand*{\StateDescriptorSet}{\StateFunctions^\mathrm{\StateDescriptor}}
\newcommand*{\StateDescriptorSetDomain}{\StateDescriptorSet_{\pm}}
\newcommand*{\StateValueSet}{\StateFunctions^\mathrm{\StateValue}}
\newcommand*{\StateFunctions}{\mathcal{S}}
\newcommand*{\StateFunctionDef}{\StateFunctions = \{ \StateDescriptorSet, \StateValueSet\}}
\newcommand*{\FeatureSet}{\Phi}
\newcommand*{\FeaturePool}[1]{{\FeatureSet}_{#1}^{\Domain}(\TrainingTrajectories)}
\newcommand*{\FeatureCondition}{\mathrm{Cond}}
\newcommand*{\FeatureEffect}{\mathrm{Eff}}
\newcommand*{\BooleanFeature}{b}
\newcommand*{\NumericFeature}{q}
\newcommand*{\BooleanFeatureSet}{\FeaturePool{\BooleanFeature}}
\newcommand*{\NumericFeatureSet}{\FeaturePool{\NumericFeature}}
\newcommand*{\LoopLandmarkCounter}{\LoopConditions^{\mathrm{count}}}
\newcommand*{\LoopSymbol}{\ell}
\newcommand*{\LoopDef}[2]{\LoopSymbol = (#1,#2)}
\newcommand*{\LoopConditions}{\mathcal{L}}
\newcommand*{\LoopConditionsOfLoop}{\LoopConditions_\LoopSymbol}
\newcommand*{\LoopConditionsDef}{\LoopConditionsOfLoop = (\ExitCondition,\LoopLandmarkCounter)}
\newcommand*{\ExitCondition}{\LoopConditions^{\mathrm{exit}}}
\newcommand*{\LandmarkGraph}{\graph_\Domain}
\newcommand*{\LMGraphDef}{\LandmarkGraph = (\nodes, \DirectOrderings, \ConditionalOrderings, \LoopConditions)}
\newcommand*{\LMGraphDefGuide}{\LandmarkGraph = (\nodes, \DirectOrderings, \ConditionalOrderings, \LoopConditions, \GuideSet)}

\newcommand*{\DirectOrderings}{\edges^\mathrm{O}}
\newcommand*{\ConditionalOrderings}{\edges^\mathrm{L}}
\newcommand*{\LMHeurName}{\text{\emph{generalized landmark counting heuristic}}~\LMHeur}
\newcommand*{\LMHeur}{\mathrm{LM}^{\graph}}
\newcommand*{\LMHeurGuided}{\LMHeur_{\Guide}}

\newcommand*{\HAdd}{\SearchHeuristic_{add}}
\newcommand*{\maxValueH}{\mathrm{maxLM}}
\newcommand*{\NumberCompounds}{\mathrm{numD}}
\newcommand{\HeuristicPointerLM}{\texttt{nextLM}}
\newcommand{\HeuristicPointerLoop}{\texttt{loopStart}}
\newcommand{\HeurValueLM}{\SearchHeuristic_{LM}}
\newcommand{\HeurValueGuide}{\SearchHeuristic_{\Guide}}
\newcommand*{\AcceptVariable}[2]{{\mu}_{#1,#2}}
\newcommand*{\SelectStateFunction}[1]{{w}_{#1}}
\newcommand*{\SelectDescriptor}{\SelectDescriptorFunc{\StateDescriptor}}
\newcommand*{\SelectDescriptorFunc}[1]{\mathrm{select}_{#1}}

\newcommand*{\SelectedLandmarkStatesFunction}[3]{{\SelectedLandmarkStatesVariable}^{#1}(#2,#3)}
\newcommand*{\SelectedLandmarkStatesVariable}{J}
\newcommand*{\LandmarkStateIndex}{l}
\newcommand*{\LandmarkStateFunctionIndex}{u}
\newcommand*{\LandmarkGraphExperiment}[1]{\graph^{#1}}
\newcommand*{\LandmarkGraphSketch}{\LandmarkGraphExperiment{\mathrm{Sk}}}
\newcommand*{\LandmarkGraphExample}{\LandmarkGraphExperiment{\mathrm{Ex}}}
\newcommand*{\LandmarkGraphDisc}{\LandmarkGraphExperiment{\mathrm{Disc}}}
\newcommand{\Preprocessing}{\psi}
\newcommand{\FeatureH}{{\mathrm{hold}}^{\BooleanFeature}}
\newcommand{\FeatureN}{{\mathrm{num}}^{\NumericFeature}}
\newcommand{\FeatureT}{{\mathrm{distT}}^{\NumericFeature}}
\newcommand{\FeatureP}{{\mathrm{distP}}^{\NumericFeature}}

\newcommand{\stateIndex}{j}
\newcommand{\trajectoryStateIndex}{\stateIndex_\Trajectory}
\newcommand{\trajectoryStateIndexPrime}{\trajectoryStateIndex'}
\theoremstyle{plain}

\newtheorem{definition}{Definition}

\newtheorem{example}{Example}

\title{Revisiting Landmarks: Learning from Previous Plans to Generalize over Problem Instances}

\author{Issa Hanou, Sebastijan Duman{\v{c}}i{\'{c}}, Mathijs de Weerdt \\ \texttt{i.k.hanou@tudelft.nl} \\
Delft University of Technology}
\date{\today}

\bibliography{landmarks}

\begin{document}

\maketitle

\begin{abstract}
    Landmarks have greatly improved AI planning by identifying the must-reach states for any plan that solves an instance. However, because landmarks are ground atoms, they are instance- and object-specific and thus do not capture general knowledge about the problem domain. Moreover, landmark graphs often contain repetitive chains of landmarks, thereby increasing graph size. We propose a new framework for discovering landmarks that automatically generalize across a domain. These generalized landmarks are learned from a set of solved instances and describe intermediate goals for planning problems where traditional landmark extraction algorithms fall short. Our generalized landmarks extend beyond the predicates of a domain by using state functions that are independent of the objects of a specific problem and apply to all similar objects, thus capturing repetition. Based on these functions, we construct a directed generalized landmark graph that defines the landmark progression, including loops for repetitive subplans. To use generalized landmarks for planning, we present a simple counting heuristic that exploits loop behavior and local landmark guides to find plans more efficiently. For three domains, we construct the generalized landmark graph by hand and show that the heuristic scales across large and complex instances. The learning algorithm iteratively identifies the earliest next landmark in training trajectories by selecting the state functions that describe it. Our results show that we can learn generalized landmark graphs from a few small instances, which are also effective at solving some larger instances in the same domain, provided that a loop indicating repetition is identified. Generalized landmarks thus provide a way to automate the use of domain knowledge, through hand-constructed graphs or learning from example plans.
\end{abstract}

\section{Introduction}
\label{sec:lm:intro}
Real-world problems involve many objects, often leading to symmetries, resulting in a vast planning search space. 
To prune the search space,~\textcite{Porteous2001} introduced landmarks, which are facts that must hold at some point in all plans that solve a planning problem.
For example, to bring a box to another room, the agent must first hold it.
Heuristics that use landmarks, such as counting those already achieved in previous states, have greatly improved the planning approaches for many problems~\cite{Richter2008,Karpas2009}.
However, these landmarks have one main limitation: they only apply only to a single problem instance and thus mustomputed individually.
This results in two main shortcomings: i) they are dependent on the specific objects of that instance and ii) they do not reuse any knowledge from solving previous instances. 

These drawbacks arise because landmarks are currently extracted from problem definitions by propagating the action preconditions backward from the goal~\cite{Porteous2001}, and we refer to these as \emph{traditional landmarks} for clarity. 
For example, take an instance of the \Delivery{} domain, such as shown in~\autoref{fig:lm:delex}. 
Traditional landmarks for this problem (\autoref{fig:lm:delex:trad}) tell you that to deliver package~$\PackageVar_1$ to cell~$\CellVar_1$, the truck~$\TruckVar_1$ first must carry package~$\PackageVar_1$. 
Similarly, to deliver package~$\PackageVar_2$, the truck~$\TruckVar_1$ needs to carry it, and these traditional landmarks are separate from one another while they hold the same information.
Moreover, truck~$\TruckVar_1$ could take different paths to deliver~$\PackageVar_1$ after picking it up: visiting either cell~$\CellVar_3$ or cell~$\CellVar_5$ on the way to cell~$\CellVar_1$, for example. 
Although this symmetry can be captured using a traditional disjunctive landmark such as~$at(\TruckVar_1,\CellVar_3) \lor at(\TruckVar_1, \CellVar_5)$~\cite{Porteous2002}, this does not reveal the underlying symmetry of different paths, only these two specific cells.
We illustrate four other symmetries that cannot be captured by traditional landmarks at all.
First, if the target cell were not cell~$\CellVar_1$ but cell~$\CellVar_2$ instead, half of the traditional landmarks in~\autoref{fig:lm:delex:trad} no longer hold.
Second, if we add another package called~$\PackageVar_4$ in cell~$\CellVar_9$, we must recompute all traditional landmarks.
Third, if the packages are renamed, we must recompute all traditional landmarks again, while the plan would effectively remain unchanged. If the grid is extended by one row and one column while the objects remain in their original positions, these traditional landmarks are no longer applicable, and we cannot reuse the knowledge gained from solving the instance in~\autoref{fig:lm:delex}.
Because traditional landmarks are always grounded atoms (i.e., a predicate instantiated with specific objects), current extraction methods cannot find any that apply to different objects or instances.

\begin{figure}[t]
    \begin{subfigure}{.46\columnwidth}
        \centering
        \begin{tikzpicture}[]
            \def\cellsize{1.3cm}
            \def\iconscale{2.3}
            \foreach \x in {0,1,2} {
                \foreach \y in {0,1,2} {
                    \draw[thick] (\x*\cellsize, \y*\cellsize) rectangle ++(\cellsize, \cellsize);
                }
            }
            \node at (0.5*\cellsize, 0.55*\cellsize) {\scalebox{\iconscale}{\faFlagCheckered}};
            \node at (0.5*\cellsize, 0.15*\cellsize) {$\CellVar_1$};
            \node at (1.5*\cellsize, 0.15*\cellsize) {$\CellVar_2$};
            \node at (2.5*\cellsize, 0.15*\cellsize) {$\CellVar_3$};
            \node at (0.5*\cellsize, 1.15*\cellsize) {$\CellVar_4$};
            \node at (1.5*\cellsize, 1.15*\cellsize) {$\CellVar_5$};
            \node at (2.5*\cellsize, 1.15*\cellsize) {$\CellVar_6$};
            \node at (0.5*\cellsize, 2.15*\cellsize) {$\CellVar_7$};
            \node at (1.5*\cellsize, 2.15*\cellsize) {$\CellVar_8$};
            \node at (2.5*\cellsize, 2.15*\cellsize) {$\CellVar_9$};
            \node at (0.5*\cellsize, 1.6*\cellsize) {\scalebox{\iconscale}{\faTruck}};
            \node[text=white] at (0.4*\cellsize, 1.65*\cellsize) {$\mathbf{\TruckVar_1}$};
            \node at (2.5*\cellsize, 1.6*\cellsize) {\scalebox{\iconscale}{\textcolor{col4}{\faBox}}}; 
            \node[text=textcol4] at (2.5*\cellsize, 1.45*\cellsize) {$\mathbf{\PackageVar_1}$}; 
            \node at (1.5*\cellsize, 2.6*\cellsize) {\scalebox{\iconscale}{\textcolor{col2}{\faBox}}}; 
            \node[text=textcol2] at (1.5*\cellsize, 2.45*\cellsize) {$\mathbf{\PackageVar_2}$}; 
            \node at (2.5*\cellsize, 0.6*\cellsize) {\scalebox{\iconscale}{\textcolor{col3}{\faBox}}}; 
            \node[text=textcol3] at (2.5*\cellsize, 0.45*\cellsize) {$\mathbf{\PackageVar_3}$}; 
        \end{tikzpicture}
        \caption{\Delivery{} instance with~3~packages.}
        \label{fig:lm:delex}
    \end{subfigure}
    \begin{subfigure}{.53\columnwidth}
        \centering
        \begin{tikzpicture}[node distance=0.4cm]
            \node[draw] (a) {At package};
            \node[draw, below=of a] (b) {Pickup package};
            \node[draw, below=of b] (c) {At target};
            \node[draw, below=of c] (d) {Drop package};
            \draw[->, >=Stealth] (a) -- (b);
            \draw[->, >=Stealth] (b) -- (c);
            \draw[->, >=Stealth] (c) -- (d);
            \draw (d.east) edge[->, >=Stealth, bend right=45] node[right, align=left, text width=2cm]{$\mathrm{Number}\; \mathrm{of}\; \mathrm{undelivered}$} node[right, align=left, text width=2cm, pos=0.3]{$\mathrm{packages} > 0$} (a.east);
        \end{tikzpicture}
        \caption{Generalized landmark graph for \Delivery{}.}
        \label{fig:lm:delex:g}    
    \end{subfigure}
    \begin{subfigure}{\columnwidth}
        \newcommand{\nodeDistanceLandmarks}{0.3cm}
        \centering
        \begin{tikzpicture}[node distance=\nodeDistanceLandmarks]
            \node[draw] (g) {$at(\TruckVar_1,\CellVar_8)$};
            \node[draw, right=of g] (h) {$carry(\TruckVar_1,\PackageVar_2)$};
            \node[draw, right=of h] (i) {$at(\TruckVar_1,\CellVar_2)$};
            \node[draw, right=3.5cm of i, thick] (d) {$at(\TruckVar_1,\CellVar_1)$};
            \node[draw, right=of d] (z) {$at(\PackageVar_2,\CellVar_1)$};
            \draw[->, >=Stealth] (g) edge[] (h);
            \draw[->, >=Stealth] (h) edge[] (i);
            \draw[->, >=Stealth] (i) edge[] (d);
            \draw[->, >=Stealth] (d) edge[] (z.west); 
            \node[draw, above=of g] (a) {$at(\TruckVar_1,\CellVar_6)$};
            \node[draw, right=of a] (b) {$carry(\TruckVar_1,\PackageVar_1)$};
            \node[draw, anchor=south west, at={(b.south east)}, xshift=\nodeDistanceLandmarks, text width=4cm, align=center] (c) {$at(\TruckVar_1,\CellVar_3) \lor at(\TruckVar_1,\CellVar_5)$};
            \node[draw, above=of z] (x) {$at(\PackageVar_1,\CellVar_1)$};
            \draw[->, >=Stealth] (a) edge[] (b);
            \draw[->, >=Stealth] (b) edge[] (c);
            \path[->, >=Stealth] (c.south) edge[] (d);
            \path[->, >=Stealth] (d) edge[] (x.west);
            \node[draw, below=of g] (e) {$at(\TruckVar_1,\CellVar_3)$};
            \node[draw, right=of e] (f) {$carry(\TruckVar_1,\PackageVar_3)$};
            \node[draw, right=of f, text width=5cm, align=center, anchor=north west, at={(f.north east)}] (c2) {$at(\TruckVar_1,\CellVar_5) \lor at(\TruckVar_1,\CellVar_7) \lor at(\TruckVar_1,\CellVar_9)$};
            \node[draw, below=of z] (y) {$at(\PackageVar_3,\CellVar_1)$};
            \draw[->, >=Stealth] (e) edge[] (f);
            \path[->, >=Stealth] (f.east) edge[] (c2);
            \path[->, >=Stealth] (c2.north) edge[] (d);
            \draw[->, >=Stealth] (d) edge[] (y.west);
            \node[above=0.2cm of a] (empty) {};
            \node[draw, left=of g] (a0) {$at(\TruckVar_1,\CellVar_4)$};
            \draw[->, >=Stealth] (a0) edge[] (a);
            \path[->, >=Stealth] (a0) edge[] (e);
            \path[->, >=Stealth] (a0) edge[] (g);
            \node[above=of a0] (space) {};
        \end{tikzpicture}
        \caption{Simplified graph of traditional landmarks for the \Delivery{} instance in~\autoref{fig:lm:delex}.}
        \label{fig:lm:delex:trad}       
    \end{subfigure}    
    \caption{The \Delivery{} domain. (a) shows a problem instance where the truck needs to deliver three packages to the (same) target cell. (b) gives an example of a generalized landmark graph. (c) shows the traditional landmarks for the instance in (a).}
    \label{fig:lm:del}
\end{figure}

We address the lack of generalization (to instances, objects, etc.) of traditional landmarks by introducing a more expressive language using first-order functions.
We use this to define \textbf{generalized landmarks} as landmarks that generalize across a set of problems in the same domain rather than to a single instance. 
While humans can see that any object needs to be picked up before it can be placed at a different location (which is the same action for any package and any location), traditional landmarks are always grounded to the objects of a specific problem instance. Therefore, they cannot capture this generality. Generalized landmarks do not specify a specific object, such as package~$\PackageVar_1$, but simply capture the notion of carrying \emph{any} package. For example, a generalized landmark can say that to put some package~$\PackageVarPrime$ in a location~$\LocationVarPrime$, the truck must carry package~$\PackageVarPrime$, and before that, it must be empty. Furthermore, since this generalized landmark is no longer bound to a specific package, we no longer need separate landmarks for different packages, and we can use repetition to capture the same landmark pattern across different objects. Generalized landmarks hold for \emph{any} object, and thus, also for any number of objects within a problem instance. 

This paper proposes a framework for generalized landmarks expressed as first-order functions in contrast to grounded atoms. 
We construct generalized landmarks using state functions, which are first-order functions over the predicates in a domain and either hold or do not hold in any state.
These functions allow us to abstract beyond atomic predicates and object instantiations, making generalized landmarks insensitive to symmetries within a problem instance. 
Moreover, these are insensitive to differences between problems, as we generalize landmarks to an entire domain of problems and compute them once, reusing them across all instances.

\begin{example}\label{ex:lm:des-simple}
    \autoref{fig:lm:delex:g} shows an example graph of generalized landmarks for the \Delivery{} domain, restricted to one truck with capacity one (based on instances from the DLplan library~\cite{Drexler2022b}).
    For each package, the same four steps must be repeated.
    Because the truck is restricted to carrying one package at a time, these steps must be done sequentially for each package. 
    The first-order functions that constitute the generalized landmarks are functions that take a state~$\stateVar$ and return a truth value. 
    For example, function~$\StateDescriptor_1$ returns \texttt{true} if there is an empty truck;~$\StateDescriptor_2$ returns \texttt{true} if a truck is at a cell with no packages;~$\StateDescriptor_3$ returns \texttt{true} if a truck is at its target cell; and~$\StateDescriptor_4$ returns \texttt{true} if an empty truck is at a target cell.
    In the \Delivery{} domain, each package has a target cell where it should be delivered.
\end{example}

\begin{sloppypar}
    The generalized landmark graph shows a loop with a condition:~$\mathrm{Number\,of\,undelivered\,packages>0}$.
    This condition also takes the form of a first-order function, and when it does not hold, the next landmark must be achieved, or the goal is reached.
    In the \Delivery{} example, the loop condition is a function that compares each package's target cell with its current location to determine the number of undelivered packages. 
    If this is zero (i.e., all packages are at their target cells), the loop is complete, and the goal is reached; if it is greater than zero, the loop must be traversed again.
\end{sloppypar}

To make planning more efficient by reusing the generalized landmarks' knowledge, we propose a \textbf{generalized landmark counting heuristic} that can be used in any planner setup.
The heuristic employs information from the landmarks themselves and from the loop condition.
Moreover, each loop is annotated with a function of the initial state that computes the exact number of times the loop can be traversed for a specific instance.
This allows us to quickly determine the total number of landmarks to accept, and we use it to efficiently prune the search space.
Generalized landmarks form an abstract plan that serves as a long-horizon guide to the goal.
Additionally, we can provide a state function that serves as a \emph{local guide heuristic}: for each landmark, it guides the short-horizon search to the next one.

Our experiments show that we can efficiently solve difficult instances across three chosen domains of varying complexity by hand-constructing generalized landmark graphs for these domains. 
By hand-constructing these generalized landmark graphs, we capture domain knowledge and inherent patterns. 
This allows us to use generalized landmarks to translate expert experience into automated planner support.  

To achieve this generalization across problem instances systematically, we propose a method to \emph{learn} generalized landmarks from a set of previously successful plans. 
Since generalized landmarks apply to an entire domain, we no longer extract them from the problem instance, as is done for traditional landmarks. 
Instead, we provide an algorithm that discovers them from data; more precisely, from several plans that have already succeeded before.
Based on the state trajectories of these plans and a set of state functions, we identify the functions that describe generalized landmark states across all given plans.
By focusing on generalized landmarks that identify must-reach states, we traverse a smaller search space than when detecting actions or state transitions, such as in policy sketches~\cite{Drexler2024}. 

This paper presents a framework for generalized landmarks that generalize across problem instances in a domain, and a heuristic for using them in planning applications.
Generalized landmarks provide four advantages over traditional ones: 
i) they generalize over all instances in a domain,
ii) they only have to be computed once,
iii) they are learned from small instances and scale to larger instances,
and iv) they capture more general relations and repetition in a domain.
Finally, we develop a method to discover generalized landmarks from just a few small solved instances.

The structure of this paper is as follows.
First, we discuss relevant background information and related work on landmarks and other planning abstractions. 
We then formally establish generalized landmarks.
Next, we present our novel landmark heuristic.
We discuss the custom landmark graphs for three well-known planning domains: \Delivery{}, \Ferry{}, and \Childsnack{}.
Given these custom graphs, we show that with our heuristic, we can solve more instances while requiring fewer state expansions than the baseline heuristic.
Afterward, we discuss the discovery process and demonstrate the current status of learning graphs of generalized landmarks from previous plans.

\section{Background}
\label{sec:lm:back}
This section introduces the concepts and notation that we use in this paper. 
A \emph{planning task} $\planningTask$ is described by a tuple $\langle \Predicates, \Actions, \Objects, \InitialState, \Goal \rangle$, where
\begin{description}
    \item[] $\Predicates$ is a set of predicates,
    \item[] $\Actions$ is a set of actions,
    \item[] $\Objects$ is a set of objects,
    \item[] $\InitialState$ is the initial state, and
    \item[] $\Goal$ expresses the goal facts.
\end{description} 
A \emph{state} is a conjunction of facts, or \emph{atoms}. An atom is a grounded predicate, which is a predicate~in~$\Predicates$ with instantiated parameters to objects from~$\Objects$. 
An action has \emph{preconditions} and \emph{effects} such that an action is applicable in a state~$\stateVar$ if all preconditions are true~in~$\stateVar$, and applying an action in state~$\stateVar$ alters that state according to the effects. 
A solution to a planning task is a \emph{plan}~$\plan$, which is a sequence of actions that is successively applicable from the initial state~$\InitialState$ and results in a state~$s^*$ such that~$\Goal \subseteq s^*$. 
The \emph{state trajectory}~$\Trajectory = (\InitialStateVar, \dots, \FinalStateVar)$ is the sequence of states visited by executing plan~$\plan$ from the initial state~$\InitialState = \InitialStateVar$, such that the last state~$\FinalStateVar$ achieves the goal~$\Goal \subseteq \FinalStateVar$.
We say \emph{a state is reached by a plan} when the state is present in the trajectory of the plan's execution. 
A planning task belongs to a \emph{domain}~$\Domain = \langle \Predicates, \Actions, \ObjectTypeSet \rangle$, where
\begin{description}
    \item[] $\Predicates$ is the set of predicates shared by all planning tasks in the domain,
    \item[] $\Actions$ is the set of actions shared by all planning tasks in the domain, and
    \item $\ObjectTypeSet$ is the set of object types shared by all planning tasks in the domain.
\end{description}
Each planning task in a domain has specific objects~$\ObjectLetter \in \Objects$ which each belong to an object type~$\ObjectType \in \ObjectTypeSet$ that belongs to the domain. The parameters of actions and predicates in the domain are annotated with an object type, restricting the objects that can be used to ground the actions and predicates during planning. 

An atom is a \emph{traditional landmark} if, for every plan~$\plan$ that solves planning task~$\planningTask$, there is a state~$\stateVar \in \Trajectory$ traversed by~$\plan$ such that the atom holds in that state~\cite{Hoffmann2003}.
Most traditional landmark approaches in planning use heuristics.
Because different plans may reach the same state achieving a traditional landmark, heuristics should account for these different paths through the search tree. 
Many traditional landmark heuristics are path-dependent and evaluate a state only when it is first reached, which is sufficient because traditional landmarks must hold along all paths that reach a state~\cite{Richter2010}.
While traditional landmarks can be actions or facts (atoms), we focus solely on fact landmarks. 

A traditional landmark is \emph{accepted} in a state if it is true in that state, and all traditional landmarks ordered before this were accepted in the previous state of the trajectory. Once a traditional landmark is accepted, it remains accepted in all successor states. 
We say a traditional landmark is accepted in a state or a state \emph{achieves} a traditional landmark; both terms are used interchangeably throughout this paper. 
We use this notion of accepted and achieved for generalized landmarks.

\begin{sloppypar}
\begin{example}
    Consider the \Delivery{} example from~\autoref{fig:lm:del}. This domain has three object types: $\Truck$, $\Package$, and $\Cell$, where $\Truck$ and $\Package$ are usually associated with a supertype $\Item$. We have four predicates: $at(\Item, \Cell)$, $carrying(\Truck, \Package)$, $empty(\Truck)$, and $adjacent(\Cell_1,\Cell_2)$. We can also use predicates to indicate the object types: $truck(\Truck)$, $package(\Package)$, and $cell(\Cell)$. The domain has four actions: $move(\Truck, \Cell_1, \Cell_2)$, $pickup(\Truck, \Package, \Cell)$, and $drop(\Truck, \Package, \Cell)$. The specific instance in~\autoref{fig:lm:delex} has the following objects~$\Objects$: $\TruckVar_1$ a $\Truck$ (and thus an~$\Item$), three $\Package$ objects (also $\Item$s):~$\PackageVar_1$, $\PackageVar_2$, and $\PackageVar_3$; and nine $\Cell$ objects. In the initial state (shown in~\autoref{fig:lm:delex}), we have (amongst others) the atoms $empty(\TruckVar_1)$, $adjacent(\CellVar_1,\CellVar_2)$, and~$at(\TruckVar_1, \CellVar_4)$. Given the traditional landmark graph from~\autoref{fig:lm:delex:trad}, a traditional landmark for this instance is an atom such as~$at(\TruckVar_1,\CellVar_1)$ with the specific objects of truck~$\TruckVar_1$ and cell~$\CellVar_1$, which is accepted in every goal state. \emph{Disjunctive landmarks} such as $at(\TruckVar_1, \CellVar_3) \lor at(\TruckVar_1, \CellVar_5)$ are disjunctions of atoms, of which one must hold in every plan that reaches the goal~\cite{Porteous2002}. This landmark can be achieved by different states, where the truck is either at~$\CellVar_3$ or~$\CellVar_5$.
\end{example}
\end{sloppypar}

\section{Related Work on Landmarks and Abstractions}
\label{sec:lm:rel}
Many studies have considered traditional landmarks in different contexts. 
This section provides an overview of related work on traditional landmarks and highlights other areas of research that tie in closely to generalized landmarks to show the novelty and benefits of generalized landmarks. 

Traditional landmarks were proposed by~\textcite{Koehler1991} in terms of a goal agenda. and later work proposed ordering traditional landmarks~\cite{Porteous2001,Hoffmann2003}. Soon, traditional disjunctive landmarks were proposed~\cite{Porteous2002}, which are extracted by symmetrically reducing the problem of finding traditional disjunctive atomic landmarks~\cite{Gregory2004}. That work identified four types of symmetries in a plan or problem instance, yet each depends on the instance and its objects. Traditional landmarks can thus capture symmetries using disjunctions, but we generalize these across a domain by moving away from object instantiations. 

Other studies have also focused on exploiting problem symmetries to improve planning.~\textcite{Illanes2019} constructed a framework for solving families of problems by creating a class of problems that generalizes over the objects. Their method computes a policy for this problem class that can be used to solve individual instances. On the other hand,~\textcite{Ridder2014} defined equivalence classes of objects in the lifted planning space to reduce symmetries and limit the number of applicable actions. In both cases, the method explicitly generalizes over objects, whereas our method identifies general relations in the problem (domain) structure. Exploiting these relations helps planners solve more realistic problems by stepping beyond traditional fact landmarks.

More recently, traditional landmarks have been used in other areas of planning besides heuristic search. 
To counter some of the limitations of traditional landmarks we have pointed out, lifted landmarks were proposed, which naturally capture a special case of traditional grounded disjunctive landmarks~\cite{Wichlacz2022}. 
Lifted planning provides an abstraction by parameterizing predicates and actions over a finite universe of objects, thus not sticking to the object instantiations. 
However, unlike generalized landmarks, they are still limited to the predicates of a domain and thus cannot capture relations involving the combination or counting of predicates.

Some studies looked at a more general version of the traditional landmark graph. The illustrated landmark graph was proposed as an abstract task specification used in imitation and reinforcement learning approaches~\cite{Watson2024}. Each path in this graph represents a way to complete the task, and the authors demonstrated the benefits of a multi-path graph. Although these illustrated landmark graphs were human-generated, they show the usefulness of higher-level landmarks to areas like imitation and reinforcement learning, where generalized landmarks may also be applied. As the name suggests, illustrated landmark graphs are based on human illustrations and thus cannot be generated automatically, contrary to generalized landmarks.

\textcite{Amitai2021} proposed a Temporal Plan Network formulation for temporal planning, using diverse planning to create many plans for one problem instance, which are then merged into one network. We do something similar, where we infer the high-level states that can be grouped, except we extend this over problem instances. On the other hand, prior work on diverse planning has used traditional disjunctive landmarks to generate diverse plans~\cite{Bryce2014}. Although our work does not use diverse planning, we see a connection: using it to generate training data could be a future step toward improving our learning method.

We now discuss the related areas of generalized planning and task decomposition (such as hierarchical planning), and how generalized landmarks relate to these.

\subsection{Generalized planning}
Generalized planning focuses on finding a plan that works for multiple environments, or instances of a problem class~\cite{Hu2011}.
Generalized plans can be represented by planning programs that use pointers to objects in a planning instance.~\textcite{SegoviaAguas2022} translated traditional landmark graphs for each planning problem into pointer landmarks, selecting pointer assignments that must be achieved by a solution using the LAMA algorithm. This method still requires computing a traditional landmark graph for every problem instance in the problem set instead of generalizing to the generalized planning problem. 
Furthermore,~\textcite{Chen2026} proposed a generalized planner \textsc{Moose} that constructs lifted first-order rules over the state and goal conditions, and decides the action(s) to take.
They computed these rules using goal regression to find the relevant facts with respect to the goal.
Additionally, recent work proposed a state-centric formulation of generalized planning that learns to predict the successor state rather than the action sequence~\cite{Gupta2026}. 
While their method solved out-of-distribution instances, it relies on a state embedding to learn the transition model, providing a neural prediction from which the eventual plan is extracted.
However, generalized landmarks yield more interpretable abstractions, and our learning idea remains symbolic in nature.

Another method proposed for generalized planning uses \emph{policy sketches}~\cite{Drexler2024}. 
Using domain features~\cite{Bonet2018}, which are functions over a state, they can describe more extensive relations than domain predicates. 
A policy sketch is a set of sketch rules over a set of Boolean and numerical features of the form~$\FeatureCondition \to \FeatureEffect$ that express how the values of the features are supposed to change, effectively defining the subgoals in a problem. 
Here,~$\FeatureCondition$ is a set of Boolean feature conditions, which is~$\BooleanFeature$ or~$\neg\BooleanFeature$ for Boolean features and~$\NumericFeature=0$ or~$\NumericFeature>0$ for numeric features in the feature set~$\FeatureSet$, and~$\FeatureEffect$ consists of feature effects, which are expressions~$\BooleanFeature$,~$\neg\BooleanFeature$, or~$\BooleanFeature?$ for Boolean features, and~$\NumericFeature\uparrow$,~$\NumericFeature\downarrow$, or~$\NumericFeature?$ for numeric features. The effects~$\BooleanFeature?$ and~$\NumericFeature?$ indicate that for those rules, the effect on this feature does not matter.
Policy sketches are very similar to generalized landmarks: each sketch rule identifies a subgoal to be reached, and their use of features captures expressive domain relations. 
However, generalized landmarks differ from policy sketches in that they describe states and do not convey action information, whereas policies specify the state transitions of an action sequence.
As generalized landmarks are thus constructed in the state space, while policy sketches are constructed in the action space, we expect that generalized landmarks are easier to find. Moreover, we can easily apply generalized landmarks in a heuristic that can be used with different search algorithms, while policy sketches are more abstract and require a specialized width-based planner.

\subsection{Task decompositions}
Generalized landmarks are a form of subgoals, and a generalized landmark graph can be seen as an abstract plan. 
We could, in fact, plan between generalized landmarks to create a hierarchical structure of high-level (abstract) generalized landmarks and low-level plans between them. Such a hierarchy is somewhat related to Hierarchical Task Network (HTN) planning, which decomposes a set of \emph{abstract~tasks} hierarchically into executable actions~\cite{Bercher2019}. Although generalized landmarks form a high-level plan description, HTNs can define many more layers in the hierarchy, and the depth-width ratio is thus very different.

The main problem in HTN planning is to find the task decomposition into \emph{methods}, which remains difficult in some domains is still done manually~\cite{Li2024a}. One approach learns \emph{domain~landmarks} for a set of problems, which are conditions that occur in at least one plan for every problem~\cite{FineMorris2022}. While this is very similar to our method, they (and other related work on domain landmarks in HTN planning) use natural language processing to find domain landmarks, employing a hierarchical clustering method to find the domain landmarks. Then, each individual plan trace is used to identify the methods for achieving each high-level domain landmark. Although these domain landmarks can capture numeric properties, in a numeric planning problem, they are still bound to the predicates and functions of the domain. 

\textcite{Elimelech2022} extracted abstract skills from previously solved tasks, where skills are traces of states. This work was proposed for task decomposition in robotics and showed a great benefit of reusing information from previous plans in transfer learning. The skills are translated to an abstract domain for easy transfer to new domains. Later work demonstrated the use of abstract skills to find plans in new domains by matching skills in the library to the current state, thereby solving a Constraint Satisfaction Problem~\cite{Elimelech2023}. Although this work used a manually constructed library of skills, a follow-up can automatically learn the skills~\cite{Elimelech2023a}. Finally, they created a planning approach using abstract strategies based on skills that provide long-term guidance similar to generalized landmarks, but skills must be grounded in the actions of the actual PDDL domain as it is a trace of states in an abstract space~\cite{Elimelech2024}. However, these skills are formulated in an abstract domain, requiring parametric abstraction keys, thus limiting the interpretability of the skills. Similar to generalized landmarks, abstract skills are used to minimize the planning effort as opposed to finding an optimal plan, though the two frameworks have very different approaches to identifying and learning. 

Another method used a Large Language Model (LLM) to decompose a complex task into subtasks, and then constructed a behavior rule library that expresses the subroutines and subgoal hierarchy of previously solved tasks~\cite{Liu2025}.
This allows agents to translate human instructions to planning-compatible code based on abstract reusable behavior.
However, this still requires human instructions (through the LLM), whereas generalized landmarks can be learned from previous executions. 

\subsection{Generalized Landmarks}
Generalized landmarks and learning from execution traces are thus very related to work on task decompositions, as generalized landmarks indeed decompose a planning problem into subgoals.
We provide a novel perspective by stepping away from domain objects and predicates and using first-order functions to define generalized landmark state descriptions. 
Generalized landmarks provide a benefit over HTN methods and abstract skills by still describing a state, which is more tangible and easier to check for.
Although generalized landmarks seem very similar to lifted landmarks, we choose the name \textbf{generalized landmarks} to emphasize the generalizability and flexibility of using first-order functions instead of domain predicates, as used in lifted landmarks. 
These functions have proven useful for identifying state transitions in policy sketches, and we now apply them to states to reduce the search space and enable the application of search heuristics.

\section{The Graph of Generalized Landmarks}
Generalized landmarks provide an abstract plan for an entire domain by identifying the sequential and repetitive patterns required to achieve the goal, which are captured in the graph representation. 
This section formally introduces the concept of \emph{generalized landmarks} and the associated concepts to create a graph of generalized landmarks. First, we introduce the scope and then define state functions and the components of the graph.

Generalized landmarks apply to an entire domain, assuming that the instances share a domain goal.
This assumption is used to categorize the instances that can be solved and builds on prior work on domain goals, showing that this is hardly a restriction~\cite{Grundke2025}.
Generalized landmarks thus apply to a domain annotated with a domain goal, and the instances generated according to this domain goal~\cite{Grundke2025code}.
The goal predicates~$\GoalPredicates$ introduced in the domain goal are included in the domain formulation~\cite{Grundke2024}.
For example, in \Delivery{}, we have predicates~$\Predicates=\{at,carrying,empty,adjacent\}$.
The problem instances considered in this domain require packages to be $at$ at a certain location, and define the goal as a conjunction of literals in the form~$at(\PackageVar,\LocationVar)$.
This is translated using the goal predicates~$\GoalPredicates=\{\goalPredicateFunc{at}\}$ such that the domain goal is~$\forall (\PackageVar, \LocationVar):\,  at(\PackageVar, \LocationVar) \to \goalPredicateFunc{at}(\PackageVar, \LocationVar)$.

The graph of generalized landmarks is defined in terms of two types of state functions. 
The landmarks themselves are defined using \textbf{state descriptors} (\autoref{def:sdes}).
The loops in the graph further use \textbf{state values} (\autoref{def:sval}).
The complete set of \emph{state functions}~$\StateFunctionDef$ thus includes a set of state descriptors~$\StateDescriptorSet$ and a set of state values~$\StateValueSet$. 
We provide the formal representation of the state functions used as examples throughout this section in~\Cref{sec:lm:funcs}.

\begin{definition}[State Descriptor]\label{def:sdes}
    A \textbf{state descriptor}~$\StateDescriptor$ is a function that takes a state as input and returns a truth value. For a state~$\stateVar$ and a state descriptor~$\StateDescriptor$, we use~$\Accepts{\stateVar}{\StateDescriptor}$ to indicate that the descriptor returns \texttt{true} in this state.
\end{definition}

\begin{definition}[State value]\label{def:sval}
    A \textbf{state~value}~$\StateValue(\stateVar) \in \mathbb{R}_{\geq0}$ is a function that takes a state~$\stateVar$ and returns a nonnegative integer value. 
\end{definition}

We define \textbf{generalized landmarks} (\autoref{def:lm}) as a set of state descriptors.
While one state descriptor can be a conjunction of individual descriptors (i.e., state descriptors are not bounded in complexity), a generalized landmark is a set of state descriptors, as each marks an individual component of the landmark.
\autoref{ex:lm:des} gives an example of the state descriptors to define the landmarks in the example graph from~\autoref{fig:lm:delex:g}.

\begin{definition}[Generalized Landmark]\label{def:lm}
    For a domain~$\Domain$, a \textbf{generalized~landmark}~$\GeneralizedLandmark$ is a set of state descriptors such that in the state trajectory~$\Trajectory$ of a plan that solves an instance of~$\Domain$, there is at least one state~$\stateVar \in \Trajectory$ such that~$\GeneralizedLandmark$ holds. 
    A generalized landmark holds if every descriptor~$\StateDescriptor \in \GeneralizedLandmark$ returns \texttt{true} in that state, which we indicate with~$\Accepts{\stateVar}{\GeneralizedLandmark}$.
\end{definition}

\begin{example}\label{ex:lm:des}
    For the \Delivery{} example, we now formally define the state descriptors described in~\Cref{sec:lm:intro} that compose the generalized landmarks in the graph of~\autoref{fig:lm:delex:g}.
    These descriptors can also be formulated more precisely to capture the full extent of the landmarks.
    For example, we define the $\mathrm{At\,Package}$ generalized landmark with~$\StateDescriptor_1'$ returning \texttt{true} when an empty truck is at a cell where there is a package, and it is not its target cell.
    A generalized landmark combines state descriptors to capture the different steps to achieve that, all of which must hold together.
    For~$\mathrm{At\,package}$,~$\StateDescriptor_1'$ is combined with the simple counting of empty objects to ensure that when the truck reaches the package, it is still empty.
    $\mathrm{Pickup\,package}$ is achieved with~$\StateDescriptor_2'$ returning \texttt{true} if a truck is carrying a package which is not at its target cell.
    Descriptor~$\StateDescriptor_3'$ achieves $\mathrm{At\,target}$ when returning \texttt{true} if the truck carries a package and is at the target cell of that package.
    Finally, the $\mathrm{Drop\,package}$ landmark is defined by~$\StateDescriptor_4'$ that returns \texttt{true} when there is an empty truck at a cell where there is also a package for which it is the target cell.
    For the exact formulation, see~\Cref{sec:lm:funcs}.
\end{example}

To capture repetitive subplans, we explicitly define \textbf{loops of generalized landmarks}. 
While a graph of generalized landmarks can be constructed without loops, loops are more efficient in scaling to instances with more objects.
However, we must ensure proper landmark progression when traversing a loop.
We thus need to know exactly when the loop can be traversed, and when to continue achieving the next landmarks.
Moreover, we need to prevent `duplication' of landmarks.
For example, in \Delivery{}, the truck that just delivered a package can pick it up again, move away, return to the target cell, and `redeliver' the same package.
These challenges are addressed by introducing a \emph{loop condition}, which is enforced on the last landmark of the loop, which we refer to as the \emph{loop landmark} (see~\autoref{ex:lm:loop}).

\begin{sloppypar}
    \begin{example}\label{ex:lm:loop}
        Consider the example graph in~\autoref{fig:lm:delex:g}, we have a loop between~$\GeneralizedLandmark_4~(\mathrm{Drop\,package})$ and~$\GeneralizedLandmark_1~(\mathrm{At\,package})$, and the node with the outgoing edge to a previously-ordered landmark is thus the loop landmark~$(\mathrm{Drop\,package})$.
        The condition on the edge is $\mathrm{Number\,of\,undelivered\,packages > 0}$.
        When no \emph{new} package can be delivered, the loop traversal is complete.
        Then, the next landmark can be traversed; or in this case, the goal is achieved.
    \end{example}
\end{sloppypar}

Loops thus provide the scaling factor for generalized landmarks and are bounded by the loop condition, which consists of two components.
First, an \emph{exit condition} on the edge connecting a loop landmark to the next ordered landmark is used to determine when the loop can no longer be traversed (e.g., all packages are delivered).
If this condition holds, then the next landmark should be achieved, or the goal is reached. 
Second, a \emph{loop landmark counter} determines the number of times a loop must be traversed for a specific instance.
This counter is computed in the initial state and indicates how often a loop landmark must be reached. 
This also allows some loops to be skipped altogether for a specific instance.

More formally, for a loop~$\LoopDef{\GeneralizedLandmark_\lmNodeIndexOne}{\GeneralizedLandmark_\lmNodeIndexTwo}$ from generalized landmark~$\GeneralizedLandmark_\lmNodeIndexOne$ to previously-ordered landmark~$\GeneralizedLandmark_\lmNodeIndexTwo$, the loop condition~$\LoopConditionsOfLoop$  (\autoref{def:cond}) is a pair that gives the exit condition~$\ExitCondition$ to determine when to continue in the landmark chain and the loop landmark counter~$\LoopLandmarkCounter$ identifying how many times a loop landmark can be achieved.
The exit condition is constructed using state descriptors, while the loop landmark counter uses state values.
An example of the loop conditions is given in~\autoref{ex:lm:cond}.
All the introduced functions are formalized in~\Cref{sec:lm:funcs}.

\begin{definition}[Loop condition]\label{def:cond}
     For a loop~$\LoopDef{\GeneralizedLandmark_\lmNodeIndexOne}{\GeneralizedLandmark_\lmNodeIndexTwo}$ , the \textbf{loop condition}~$\LoopConditionsDef$ consists of:
     \begin{itemize}
         \item[i)] the $\ExitCondition$, a set of state descriptors such that if state~$\stateVar$ accepts loop landmark~$\GeneralizedLandmark_\lmNodeIndexOne$~(i.e.,~$\Accepts{\stateVar}{\GeneralizedLandmark_\lmNodeIndexOne}$) and each~$\StateDescriptor \in \ExitCondition$ holds~(notated as~$\Accepts{\stateVar}{\ExitCondition}$), the loop must be exited; and
         \item[ii)] the $\LoopLandmarkCounter$, a set of state values that identify in the initial state the number of times the loop landmark~$\GeneralizedLandmark_\lmNodeIndexOne$ can be achieved, such that each~$\StateValue \in \LoopLandmarkCounter$ returns this nonnegative number. 
     \end{itemize}
\end{definition}

\begin{sloppypar}
\begin{example}\label{ex:lm:cond}
    In the landmark graph in~\autoref{fig:lm:delex:g}, we have the loop~$\LoopDef{\mathrm{Drop\,package}}{\mathrm{At\,package}}$. 
    We define the loop condition using a state descriptor~$\StateDescriptor_5$ which holds if all packages are at their target cell, and a state value~$\StateValue_1$ which counts the number of packages that are not at their target cell. 
    When no packages are left to deliver, the exit condition is \texttt{true}, so we have~$\ExitCondition = \{\StateDescriptor_5\}$.
    Moreover, in the initial state, the state value~$\StateValue_1$ returns the total number of packages to be delivered, which is the number of times that the loop landmark `Drop package' can be achieved, so~$\LoopLandmarkCounter = \{\StateValue_1\}$. 
    So, we know right away that the loop can only be traversed three times in the instance in~\autoref{fig:lm:delex}.
\end{example}
\end{sloppypar}

The graph of generalized landmarks provides a compact and interpretable representation, formalized in \autoref{def:graph}.
This captures the repetition and generalization of plans for problems in this domain. 
Using the formal graph definition, we now also formalize the loop landmark (\autoref{def:loop-lm}), which is the starting point of the loop, as it is the last landmark in a chain.

\begin{definition}[Graph of Generalized Landmarks]\label{def:graph}
    For a domain~$\Domain$, a \textbf{graph of generalized landmarks}~$\LMGraphDef$ has a set of generalized landmark~nodes~$\nodes$, a set of ordered edges~$\DirectOrderings$, and a set of loop edges~$\ConditionalOrderings$. Each loop edge~$\edgeDef{\edge}{\GeneralizedLandmark_{\lmNodeIndexOne}}{\GeneralizedLandmark_{\lmNodeIndexTwo}} \in \ConditionalOrderings$ is associated with a loop~$\LoopDef{\GeneralizedLandmark_\lmNodeIndexOne}{\GeneralizedLandmark_\lmNodeIndexTwo}$ with loop condition~$\LoopConditionsDef \in \LoopConditions$.
\end{definition}

\begin{definition}[Loop landmark]\label{def:loop-lm}
    Given a graph of generalized landmarks~$\LMGraphDef$, a loop landmark is a landmark~$\GeneralizedLandmark_\lmNodeIndexOne \in \nodes$ which has at least one outgoing edge to some landmark~$\GeneralizedLandmark_\lmNodeIndexTwo \in \nodes$ such that~$\lmNodeIndexTwo < \lmNodeIndexOne$ and~$\edgeDef{\edge}{\GeneralizedLandmark_\lmNodeIndexOne}{\GeneralizedLandmark_\lmNodeIndexTwo} \in \ConditionalOrderings$, and either has another edge the next landmark~$\GeneralizedLandmark_{\lmNodeIndexOne+1}$ or it is the last landmark in the chain.
\end{definition}

Generalized landmarks provide long-horizon guidance, and we use the same framework and state function introduced previously to define short-horizon heuristics as \emph{landmark guides}.
For example, in \Delivery{}, we know from~\autoref{fig:lm:del} that the first subgoal is for the truck to reach a cell with a package (that is not delivered). 
However, if our grid is not just~3x3 but~10x10, this information provides little guidance.
So, we provide the option of a \emph{guided graph of generalized landmarks} (\autoref{def:guide}) that uses a set of state values to provide a local heuristic for each generalized landmark.
Ideally, the value should be zero in a state where the landmark is accepted; in one state before this landmark state, the value should be one, and so on, such that the value decreases (or at least never increases) as the state is closer to accepting the landmark.
Usually, we only have an estimate of this value, which still works reasonably.
For example, the common domain-independent heuristic~$\HAdd$ belongs to the class of planning as heuristic search heuristics, which estimate the cost to the goal by relaxing the negative effects of actions~\cite{Bonet2001}.
\autoref{ex:lm:guide} gives an example of the landmark guides for \Delivery{}.

\begin{definition}[Guided Graph of Generalized Landmarks]\label{def:guide}
    For a domain~$\Domain$, a \textbf{guided graph of generalized landmarks}~$\LMGraphDefGuide$ has a set of guiding state values~$\GuideSet$, such that for each generalized landmark~$\GeneralizedLandmark_\lmNodeIndexOne \in \nodes$ we have exactly one guide~$\Guide_\lmNodeIndexOne \in \GuideSet$ such that in each state~$\stateVar$ that accepts landmark~$\GeneralizedLandmark_\lmNodeIndexOne$, we have~$\Guide_\lmNodeIndexOne(\stateVar)=0$ and for each state~$\stateVarPrime$ that is reachable in~$\GuideDistance$ actions to achieving~$\GeneralizedLandmark$, we have~$\Guide_\lmNodeIndexOne(\stateVarPrime) \leq \GuideDistance$.
\end{definition}

\begin{example}\label{ex:lm:guide}
    In our example in~\autoref{fig:lm:del}, the guide~$\Guide_1$ to the first landmark~$\mathrm{At\,package}$ is the distance between the truck and the nearest undelivered package, using the \texttt{adjacent} predicate in the domain, and comparing package locations with their targets to determine whether it is delivered. 
    Similarly, the guide~$\Guide_3$ to the third landmark~$\mathrm{At\,target}$ uses the \texttt{adjacent} predicate and the target cell of the package being carried.
    For the second landmark, the guide~$\Guide_2$ simply counts the \texttt{empty} atoms in the current state, which is zero if a package was picked up.
    Contrary, the guide~$\Guide_4$ of the last landmark counts the \texttt{carrying} atoms.
\end{example}

\subsection{Example of the Landmark State Functions}
\label{sec:lm:funcs}
This section shows the actual construction of the \Delivery{} landmark graph used throughout this paper. 
We first formalize the state descriptors, then the loop condition, and then the local guide.
The simple state descriptors described in~\autoref{ex:lm:des-simple} to construct a generalized landmark graph for \Delivery{} are formally defined as follows.
\begin{sloppypar}
    \begin{itemize}
        \item[$\StateDescriptor_1$] $|\{\, \Item \;|\; empty(\Item) \,\}|$  
        \item[] Count the empty objects (trucks)
        \item[$\StateDescriptor_2$] $|\{\, \Item \;|\; at(\Item,\Cell) \,\land\, \forall \Item_2 \,:\, at(\Item_2,\Cell) \to truck(\Item_2)  \,\}|$
        \item[] Count the objects $at$ the cell where there are only trucks 
        \item[$\StateDescriptor_3$] $| \{\, \Item \;|\; \goalPredicateFunc{at}(\Item) \,\land\, \exists \Item_2 \,:\,at(\Item_2,\Cell) \,\land\, empty(\Item_2) \,\}|$
        \item[] Count the objects that have a target cell defined, and there is a $truck$ object $at$ that target 
        \item[$\StateDescriptor_4$] $| \{\, \Item \;|\; \goalPredicateFunc{at}(\Item) \,\land\, \exists \Item_2 \,:\,at(\Item_2,\Cell) \,\land\, empty(\Item_2) \,\}|$ 
        \item[] Count the objects that have a target cell defined, and there is an $empty$ item $at$ that target
    \end{itemize}
\end{sloppypar}
These constitute the landmarks for \Delivery{}: $\GeneralizedLandmark_1 = \{\neg \StateDescriptor_2\}$, $\GeneralizedLandmark_2 = \{\neg \StateDescriptor_1, \StateDescriptor_2\}$, $\GeneralizedLandmark_3 = \{\StateDescriptor_3\}$, and $\GeneralizedLandmark_4 = \{\StateDescriptor_1, \StateDescriptor_4\}$

The more complicated state descriptions described in~\autoref{ex:lm:des} to define the landmarks exactly per function are formulated as follows.
\begin{sloppypar}
    \begin{itemize}
        \item[$\StateDescriptor_0'$] $|\{\, \Item \;|\; empty(\Item) \,\}|$  
        \item[] Count the number of empty (truck) objects
        \item[$\StateDescriptor_1'$] $|\{\, \Item \;|\; empty(\Item) \,\land\, \exists \Item_2 \,:\, at(\Item_2,\Cell) \to \big( \goalPredicateFunc{at}(\Item_2,\Cell_2) \setminus at(\Item_2,\Cell_2) \big) \,\}|$  
        \item[] Count the empty objects (trucks) that are at a cell at which there are also packages that are not at their target cell
        \item[$\StateDescriptor_2'$] $|\{\, \Item \;|\; \goalPredicateFunc{at}(\Item,\Cell) \,\land\, \neg at(\Item, \Cell) \,\land\, carrying(\Item_2,\Item) \land\, truck(\Item_2)  \,\}|$
        \item[] Count the objects that have a target cell ($\goalPredicateFunc{at}$) but are not there and are being carried by a truck
        \item[$\StateDescriptor_3'$] $| \{\, \Item \;|\; carrying(\Item, \Item_2) \,\land\, \exists \Item_2 \,:\, \big( \goalPredicateFunc{at}(\Item_2,\Cell) \setminus at(\Item_2,\Cell) \big)$
        \item[] $\quad \quad \quad \quad \quad \,\land\, \exists \Item_3 \,:\, at(\Item_3, \Cell)  \to truck(\Item_3)\,\}|$
        \item[] Count the objects that are carrying something which is not at its target cell, but a truck is at that cell
        \item[$\StateDescriptor_4'$] $| \{\, \Item \;|\; \goalPredicateFunc{at}(\Item) \,\land\, at(\Item,\Cell) \,\land\, \exists \Item_2 \,:\,at(\Item_2,\Cell) \,\land\, empty(\Item_2) \,\}|$ 
        \item[] Count the objects that are at their target cell, and there is an empty object (truck) at that target
    \end{itemize}
\end{sloppypar}

Now, we formalize the loop in the running example for \Delivery{}. We introduce a state function that counts the objects that have a target cell defined but are not at that target, as described in~\autoref{ex:lm:loop}. This function is used as the exit condition~($\StateDescriptor_5$) when the set is empty. It also gives the state value~($\StateValue_1$) of the loop landmark counter, which gives the size of this set.
\begin{equation}\label{eq:func:loop}
    | \{\, \Item \,|\, \goalPredicateFunc{at}(\Item,\Cell) \,\land\, \neg at(\Item,\Cell)\,\}|
\end{equation}

\begin{figure}[b]
    \centering
    \begin{tikzpicture}[lm/.style={text width=5.5cm, align=left}, loop/.style={text width=4cm, align=right}, node distance=1cm]
        \node[lm, draw] (v1) {$\GeneralizedLandmark_1 = \{\StateDescriptor_0', \StateDescriptor_1'\}$ \;\quad(at package)};
        \node[lm, draw, below=of v1] (v2) {$\GeneralizedLandmark_2 = \{\neg\StateDescriptor_0', \StateDescriptor_2'\}$ (pickup package)};
        \node[lm, draw, below=of v2] (v3) {$\GeneralizedLandmark_3 = \{\neg\StateDescriptor_0', \StateDescriptor_3' \}$ \quad\quad(at target)};
        \node[lm, draw, below=of v3] (v4) {$\GeneralizedLandmark_4 = \{ \StateDescriptor_0', \StateDescriptor_4' \}$ \;\;(drop package)};
        \node[above=0.5cm of v1] (v0) {};
        \draw[->] ([xshift=-2cm]v0.south) -- node[right] {$\Guide_1$} ([xshift=-2cm]v1.north);
        \draw[->] ([xshift=-2cm]v1.south) -- node[right] {$\Guide_2$} ([xshift=-2cm]v2.north);
        \draw[->] ([xshift=-2cm]v2.south) -- node[right] {$\Guide_3$} ([xshift=-2cm]v3.north);
        \draw[->] ([xshift=-2cm]v3.south) -- node[right] {$\Guide_4$} ([xshift=-2cm]v4.north);

        \path[->] (v4.west) edge[bend left] (v1.west);
        \node[loop, left=1cm of v2] (t) {$\LoopSymbol$: Deliver one package};
        \node[loop, below=0.1cm of t] (tl) {$\LoopConditionsDef$};
        \node[loop, below=0.1cm of tl] (tc1) {$\ExitCondition = \{\StateDescriptor_5\}$};
        \node[loop, below=0.1cm of tc1] (lc) {$\LoopLandmarkCounter = \{ \StateValue_1 \}$};
    \end{tikzpicture}
    \caption{Example graph of generalized landmarks for \Delivery{} domain, showing the state descriptors on the right.}
    \label{fig:lm:delex:fullgraph}
\end{figure}

The following state values are used to define the landmark guides, associated with the four landmarks, as described in~\autoref{ex:lm:guide}.
Some of these state values compute a distance to a state that satisfies the latter part.

\begin{itemize}
    \item[$\Guide_1$] Distance between a state where an $empty$ truck is $at$ a cell and a state such that a truck is $at$ a cell, such that a package is $at$ this cell, and it is not the target cell for the package.
    \item[$\Guide_2$] Distance between a state with where something is $empty$ and a state with a non-empty truck.
    \item[$\Guide_3$] Distance between a state where there is a non-empty truck at a cell and a state where the truck is carrying a package and at the target cell of that package. 
    \item[$\Guide_4$] Distance between a state with where something is $empty$ and a state with a non-empty truck.
\end{itemize}

Finally, \autoref{fig:lm:delex:fullgraph} shows the resulting landmark graph that we manually constructed for the \Delivery{} domain.
This graph is available within our codebase~\cite{LandmarkCode2025} and is used in an experiment in~\Cref{sec:lm:learn:exp}.
 
\section{Searching with Generalized Landmarks}
\label{sec:lm:heur}
To use generalized landmarks in planning, we design a heuristic that counts the accepted landmarks, which we name the~$\LMHeurName$.
Similarly to~\textcite{Richter2008}, we estimate the distance from state~$\stateVar$ to the goal by computing the number of landmarks that still need to be achieved from state~$\stateVar$ onward. 
\autoref{alg:lm:heur} gives the pseudocode for the~$\LMHeur$ heuristic. 
Before the search, we calculate the number of times each landmark must be achieved using the loop landmark counter~$\LoopLandmarkCounter$ (\autoref{alg:lm:heur:init}).
Given a graph of generalized landmarks~$\LMGraphDefGuide$ (where~$\GuideSet$ may be empty) and a specific instance, we evaluate the state values~$\StateValue$ of the landmark counter~$\LoopLandmarkCounter$ on the initial state, which returns a nonnegative number (\autoref{def:sval}).
Using this number, we compute the maximum value of~$\LMHeur$~(\autoref{alg:lm:heur:max}), which we denote by~$\maxValueH$, by rolling out the loops and counting the total number of landmarks to be achieved in this instance (\autoref{eq:lm:max}). If a loop is skipped, fewer landmarks have to be achieved in total. 
\begin{equation}\label{eq:lm:max}
\begin{aligned}
    \maxValueH \gets |\nodes| &+ \sum_{\substack{\forall \LoopDef{\GeneralizedLandmark_\lmNodeIndexOne}{\GeneralizedLandmark_\lmNodeIndexTwo} \in \ConditionalOrderings \\ \forall \StateValue \in \LoopLandmarkCounter \\ \StateValue(initState) > 0}} \StateValue(initState) \cdot (\lmNodeIndexOne - \lmNodeIndexTwo + 1) \\
    &- \sum_{\substack{\forall \LoopDef{\GeneralizedLandmark_\lmNodeIndexOne}{\GeneralizedLandmark_\lmNodeIndexTwo} \in \ConditionalOrderings \\ \forall \StateValue \in \LoopLandmarkCounter \\ \StateValue(initState) = 0}} (\lmNodeIndexOne - \lmNodeIndexTwo + 1)
\end{aligned}
\end{equation}

During the search, we track the next generalized landmark to be achieved, as well as the total number of accepted landmarks. 
This reference is updated as landmarks are accepted, and loops are traversed. 
We do not reevaluate states (\autoref{alg:lm:heur:visit}) but use the information from the previous state (\autoref{alg:lm:heur:visited}).
When evaluating a potential next state, we check if the next generalized landmark to be achieved is accepted in this state (\autoref{alg:lm:heur:accept}). 
We immediately check if this landmark is a loop landmark (\autoref{alg:lm:heur:loop}), and if so, \autoref{alg:lm:heur:exit} checks the loop exit condition.
Otherwise, \autoref{alg:lm:heur:next4} updates the landmark pointer by one.  
If the exit condition does \emph{not} hold yet, we update the reference to the next landmark to be achieved (first in the loop) in \autoref{alg:lm:heur:next1}.
This way, we guide the planner to re-achieve the ordered landmarks in the loop while keeping track of the total number of loop traversals. 
If the exit condition is met, we increase the pointer by one (\autoref{alg:lm:heur:next3}).
Finally, we also check how many of the next landmark's state descriptors are already accepted in the current state, which serves as a short guide (value~$\NumberCompounds$) to the complete landmarks (\autoref{alg:lm:heur:numd}).
Our total heuristic value is the precomputed maximum value minus the currently accepted landmarks, plus the value~$\NumberCompounds$ (\autoref{alg:lm:heur:res}).

\begin{algorithm*}[t]
    \begin{algorithmic}[1]
        \raggedright
        \State Initialize the loop landmark counters~$\LoopLandmarkCounter$, the pointer~\HeuristicPointerLM{} and the empty \texttt{visited} array \lline\label{alg:lm:heur:init}
        \State Initialize the landmark counter \texttt{lmCount}, tracking the achieved landmarks in each state\lline\label{alg:lm:heur:lmcount}
        \State Calculate the maximum value~$\maxValueH$ of the heuristic based on the landmark graph~$\LandmarkGraph$ and the current task~$\planningTask$ \lline\label{alg:lm:heur:max}
        \Function{ComputeHeuristic}{$state,\, prevState, \,\LMGraphDefGuide, \,\planningTask \in \Domain$}\lline\label{alg:lm:heur:func}
            \If{$state \in \texttt{visited}$}\lline\label{alg:lm:heur:visit}
                \State \Return $\maxValueH - \texttt{lmCount}[state]$ \lline\label{alg:lm:heur:visited}
            \EndIf
            \State Copy counters, the pointer~\HeuristicPointerLM{} and \texttt{visited} list from $prevState$ \lline\label{alg:lm:heur:copy}
            \State Mark $state$ as \texttt{visited} \lline\label{alg:lm:heur:visited2}
            \If{$\Accepts{state}{\GeneralizedLandmark_{\HeuristicPointerLM}}$} \lline\label{alg:lm:heur:accept}
                \If{If a loop~$\LoopDef{\GeneralizedLandmark_{\HeuristicPointerLM}}{\GeneralizedLandmark_{\HeuristicPointerLoop}}$ exists from $\GeneralizedLandmark_{\HeuristicPointerLM}$ to a previously-ordered landmark $\GeneralizedLandmark_{\HeuristicPointerLoop}$} \lline\label{alg:lm:heur:loop}
                    \If{The exit condition $\ExitCondition$ of the loop does \textbf{not} hold in $state$} \lline\label{alg:lm:heur:exit}
                        \State Accept landmark $\GeneralizedLandmark_{\HeuristicPointerLM}$ in $state$ and update loop traversal counter \lline\label{alg:lm:heur:update}
                        \State $\HeuristicPointerLM \gets \HeuristicPointerLoop$ \lline\label{alg:lm:heur:next1}
                        \If{$\Guide_{\HeuristicPointerLM}(state)=0$} \lline\label{alg:lm:heur:guide}
                            \State$\HeuristicPointerLM \gets +1$ \lline\label{alg:lm:heur:next2}
                        \EndIf
                    \Else\lline\label{alg:lm:heur:else1}
                        \State $\HeuristicPointerLM \gets +1$ \lline\label{alg:lm:heur:next3}
                    \EndIf
                \Else\lline\label{alg:lm:heur:else2}
                    \State Accept landmark $\GeneralizedLandmark_{\HeuristicPointerLM}$ in $state$ \lline\label{alg:lm:heur:accept2}
                    \State $\HeuristicPointerLM \gets +1$ \lline\label{alg:lm:heur:next4}
                \EndIf
            \EndIf
            \State $\NumberCompounds \gets |\GeneralizedLandmark_{\HeuristicPointerLM}| - \big| \{ \StateDescriptor \in \GeneralizedLandmark_{\HeuristicPointerLM} \,|\, \Accepts{state}{\StateDescriptor} \} \big|$ \lline\label{alg:lm:heur:numd}
            \State Compute guide value~$\Guide_{\HeuristicPointerLM}$ on~$state$ \lline\label{alg:lm:heur:guideV}
            \State \Return $\Big(\maxValueH + \NumberCompounds - \texttt{lmCount}[state],\;\; \Guide_{\HeuristicPointerLM}(state)\Big)$ \lline\label{alg:lm:heur:res}
        \EndFunction
    \end{algorithmic}
    \caption{Pseudocode for the~$\LMHeurName$.}
    \label{alg:lm:heur}
\end{algorithm*}

\begin{example}\label{ex:lm:heur}
    Consider again the example problem shown earlier in~\autoref{fig:lm:delex}, with initial state~$\stateVar_\InitialStateVarIndex$.
    Given the more specific landmark graph from~\autoref{fig:lm:delex:fullgraph}, we could encounter the following state exploration during the search. Truck~$\TruckVar_1$ moves two cells to the right (to package~$\PackageVar_1$, two timesteps), and thus~$\Accepts{\stateVar_2}{\GeneralizedLandmark_1}$. Next,~$\TruckVar_1$ picks up the package, reaching~$\stateVar_3$ and~$\Accepts{\stateVar_3}{\GeneralizedLandmark_2}$. We go to cell~$\CellVar_1$ and drop the package, thus in state~$\stateVar_7$, we have~$\Accepts{\stateVar_7}{\GeneralizedLandmark_4}$. Because the exit condition does not hold,~$\NotAccepts{\stateVar_7}{\StateDescriptor_5}$, we must traverse the loop and the next landmark to achieve is landmark~$\GeneralizedLandmark_1$. After picking up and delivering package~$\PackageVar_3$ in state~$\stateVar_{13}$, where again~$\Accepts{\stateVar_{13}}{\GeneralizedLandmark_4}$, so the exit condition is not met and we still have~$\NotAccepts{\stateVar_{13}}{\StateDescriptor_5}$. We traverse the loop again, and in state~$\stateVar_{21}$, we accept landmark~$\GeneralizedLandmark_4$ for the third time, the exit condition now holds, and we have finished the landmark progression. 
\end{example}
    
In general, we restrict landmark progression to accept only one landmark per state. 
However, when a state~$\stateVar$ accepts the loop landmark, and the guide to the next landmark to achieve is already zero in this state, we also immediately accept that first landmark~(\autoref{alg:lm:heur:guide}), but no further landmarks are achieved.
For example, in \Delivery{}, if a package is delivered to a cell that already contains another undelivered package, we immediately advance in the landmark chain.
By allowing this exception in the heuristic, we can handle fully defined landmark graphs and show all intermediate steps of the abstract plan, in which two landmarks are sometimes accepted in a single state. 

After checking whether the next landmark is accepted (and a possible loop is taken), the guide value~$\HeurValueGuide$ (if used) is computed for the next landmark to be accepted (\autoref{alg:lm:heur:guideV}). 
This value is used in a lexicographical evaluation combined with the value~$\HeurValueLM$ computed for~$\LMHeur$ (\autoref{alg:lm:heur:res}).

We implemented a priority queue that prioritizes the heuristic value~$\HeurValueLM$, then the weighted path cost $\SearchPathCost$, then an extra parameter~$\HeurValueGuide$, and finally the size of the search tree at the time of enqueuing.
The default priority queue in A*'s \texttt{expand} function is~$(\SearchEvalFunction,\, \SearchHeuristic,\, |searchTree|)$, where~$\SearchEvalFunction$ is the weighted path cost~$\SearchPathCost$ plus weighted heuristic value~$\SearchHeuristic$.
Because we do not prioritize the~$\SearchEvalFunction$ value, only its individual components, the search algorithm is no longer classified as A*~\cite{Asai2017}.
Our ordering is similar to the novelty ordering in best-first width search~\cite{Lipovetzky2017}, where nodes are ordered lexicographically: first by the novelty measure, then by one or more goal-directed heuristics for tie-breaking.
Our first aim is to satisfy more landmarks, then break ties on path cost as a measure of distance traversed and the guide heuristic as distance to go. 
Because the graph of generalized landmarks provides long-horizon support~$\HeurValueLM$, this is used as the first component. 
As~\textcite{Lipovetzky2017} argued, sorting first on the heuristic then on novelty is too greedy. 
While~$\HeurValueLM$ is also a heuristic value, it acts more like a novelty measure providing long-term guidance, since finding a solution with more satisfied landmarks is a `more novel' state. 

Our first tie-breaking component is the path cost~$\SearchPathCost$, as taken from the original~$\SearchEvalFunction$. 
However, as many nodes will have the same number of satisfied landmarks~$\HeurValueLM$ and path cost~$\SearchPathCost$, we must break further ties. 
Therefore, we use the short-horizon local guide heuristics~$\HeurValueGuide$, which generally provide more diverse values for different states. 
Finally, the use of the search tree size as the last component enforces a last-in-first-out ordering~\cite{Asai2017}.
Because the~$\HeurValueLM$ value is based on previously achieved landmarks, it provides a sense of depth tiebreaking, which was shown to provide good performance~\cite{Asai2017}.
To further prioritize loop traversal, we implemented a \emph{pruning mechanism} that empties the priority queue when a loop is taken or exited. 

As we use information from previously accepted landmarks, our heuristic depends on the previous state and is thus path-dependent. 
However, we do not actually require information about the previous state, only access to the landmarks accepted in that state.  
In the initial state, only the first landmark is checked, though no loops can be taken here yet.

\section{Evaluating Custom Generalized Landmarks}
\label{sec:lm:eval}
Our goal is to show that the~$\LMHeurName$ outperforms a well-known and generally good domain-independent heuristic~$\HAdd$.
To validate the use of generalized landmarks in a planner, we implemented the~$\LMHeur$ heuristic in SymbolicPlanners.jl~\cite{ZhiXuan2022a}, which enables easy heuristic implementation.
However, we recognize that this is not a state-of-the-art planner.
The state functions used to construct the generalized landmarks, their guides, and the loop conditions are all functions over a state returning either a Boolean or a numeric value. 
The implementation of these functions is not restricted to any particular form.
The state functions used in our experiments are based on description logic~\cite{Baader2003} and evaluated by the DLplan library~\cite{Drexler2022b}.
They can be replaced with a different form, as long as an evaluator is provided to evaluate the functions on a specific state.
We manually constructed the (guided) graph of generalized landmarks for two different domains from the IPC benchmark with domain goals~\cite{Grundke2025code}, as well as the \Delivery{} domain, which is used as an example throughout this paper and is taken from the DLplan benchmarks.
We use these manual graphs to show the benefits of generalized landmarks and the efficiency of our heuristic. 
First, we describe the domains and landmark graphs; then, we present our setup and results.

\begin{sloppypar}
    \paragraph{Delivery (D)}
    The \Delivery{} domain has three actions: $move$, $pickPackage$, and $dropPackage$, the object types~$cell$ and $locatable$, with subtypes~$truck, package$ of~$locatable$, four predicates: $empty$, $carrying$, $at$, and $adjacent$, and goal predicate~$\goalPredicateFunc{at}$.
    The domain goal is formulated as~$\forall (\PackageVar, \LocationVar):\,  at(\PackageVar, \LocationVar) \to \goalPredicateFunc{at}(\PackageVar, \LocationVar)$.
    The domain presents difficulties in the number of packages, their distribution over the grid, and the grid size. 
    We use the generalized landmark graph described in this paper (see~\Cref{sec:lm:funcs}).
\end{sloppypar}

\begin{sloppypar}
    \paragraph{Ferry (F)}
    The \Ferry{} domain is similar to \Delivery{} as it has three actions $sail, board$, and $debark$, and should pick up cars and sail them to their destination:~$\forall (car, loc):\,  at(car, loc) \to \goalPredicateFunc{at}(car, loc)$.
    However, the ferry object is not explicitly included like the truck, and the locations are not connected via an $adjacent$ predicate.
    While the packages are distributed across a large grid, many cars often wait at the same location.
    We construct a graph with four landmarks: go to a car that is not at its goal, pick up the car, sail the car to its goal location, and debark.
    We use a guide in this landmark graph to allow direct loop continuation: after dropping off a car at some location, thereby achieving the fourth landmark, the guide to the first landmark is used, and when evaluated to zero, it allows immediate acceptance of the first landmark again, to pick up a car waiting at this location.
\end{sloppypar}

\begin{sloppypar}
    \paragraph{Childsnack (C)}
    The \Childsnack{} domain has six actions: $moveTray$, $makeSandwich$, $makeSandwichNoGluten$, $putOnTray$, $serveSandwich$, and $serveSandwichNoGluten$.
    The goal is to serve all children (distributed across several tables) a sandwich, taking into account whether they are gluten-allergic. 
    The domain goal is~$\forall (child):\, \goalPredicateFunc{serve}(child)$.
    We define eight landmarks with two sequential loops.
    The first loop creates a gluten-free sandwich, places it on a tray, moves to a table with a child who is allergic to gluten and has not yet received a sandwich, and serves the child the sandwich.
    This is repeated until all children allergic to gluten have been served.
    The number of loop landmarks depends on the number of allergic children, and if this is zero, the loop (and all contained landmarks) is skipped altogether.
    Then, we create a sandwich for all children who are not allergic and serve them, which is a loop that depends on the number of non-allergic children.
    The second loop can also be skipped if an instance has no non-allergic children.
    While this may result in suboptimal plans, as some sandwiches can be served together on the limited number of trays, the domain is difficult due to the number of consecutive, interdependent steps, and these generalized landmarks provide an efficient abstract plan. 
    We do not use guides in this landmark graph.
\end{sloppypar}

\subsection{Experimental Setup}
We used the~A*-planner from SymbolicPlanners.jl and used the~$\HAdd$ heuristic as a baseline.
For the \Delivery{} domain, we used the thirty instances provided by the DLplan benchmark.
For the other domains, we used the provided test set for evaluating our heuristic; the number of available instances is reported in~\autoref{tab:result}.
The experiments were run on the DelftBlue supercomputer~\cite{DHPC2024}, where each heuristic tested each instance of the different domains with~8GB of memory and a time limit of~30~minutes per instance per heuristic.

Several optimizations were implemented to make the heuristic more efficient.
Whenever a loop is taken after the current state achieves a landmark, we empty the search queue (as mentioned in~\Cref{sec:lm:heur}). 
However, with non-perfect landmark graphs, this may result in undesired behavior. 
For example, when a sandwich is served to a child who has already received a sandwich, or when a package is delivered twice. 
As the exit condition is often a descriptor checking whether the difference between the goal predicate and the dynamic predicate is zero, we ensure that each loop traversal leads to a decrease in this difference. 
So, in the heuristic implementation, we keep track of a state value that is associated with the loop's exit condition, and when the exit condition is not yet met, we check whether this state value has changed since the last time the loop landmark was reached.

\subsection{Results}
\autoref{tab:result} shows the number of solved instances per domain for both~$\HAdd$ and our heuristic, where we also compare performance with~($\LMHeur_{\Guide}$) and without~($\LMHeur$) the landmark guides (where applicable). 
The table also reports the average runtime to solve the instances, averaged over the number of instances solved by that heuristic.
We see that generalized landmarks clearly outperform the baseline heuristic of~$\HAdd$ in both the number of instances solved and runtime.
For \Delivery{}, the landmark graph without guides solves all instances, while the guides increase the runtime as more computations must be made.
This holds only for the computation time in seconds, while the expanded states are always fewer when using the guide.
For \Ferry{}, we do see a difference: without the guide, the condition from~\autoref{alg:lm:heur}~\autoref{alg:lm:heur:guide} does not hold, so after dropping a car, the next car cannot be picked up immediately, indicating where the guide is very helpful.
For \Childsnack{}, we see a very clear difference, where~$\HAdd$ only solves four instances within the timeout,~$\LMHeur$ solves all 30 instances using the custom landmark graph.
This is especially interesting, as \Childsnack{} is known to be a difficult domain, with few goal facts that are achieved only after a series of actions.

\begin{table}[t]
    \centering
\caption{Number of solved instances for the different heuristic configurations and average runtime over instances that were solved by all heuristic settings.}
\label{tab:result}
\begin{tabular}{lrrrrrr}
\toprule
 & \multicolumn{3}{c}{\textbf{Solved}} & \multicolumn{3}{c}{\textbf{Avg Runtime (s)}} \\
\multicolumn{1}{r}{\textit{Heuristic}} & $\HAdd$ & $\LMHeur$ & $\LMHeurGuided$ & $\HAdd$ & $\LMHeur$ & $\LMHeurGuided$ \\
Domain (\#instances) &  &  &  &  &  &  \\
\midrule
childsnack (30) & 4 & \bfseries 30 & - & 453.73 & \bfseries 11.69 & - \\
delivery (30) & 29 & \bfseries 30 & \bfseries 30 & 104.58 & \bfseries 11.32 & 12.88 \\
ferry (57) & 49 & 39 & \bfseries 57 & 41.98 & 22.38 & \bfseries 15 \\
\bottomrule
\end{tabular}
\end{table}

\autoref{fig:lm:compare} shows the expanded states and plan length for the successful instances for the different domains.
Again, we see that our heuristic~$\LMHeur$ outperforms the~$\HAdd$ baseline. 
For the \Delivery{} domain, guiding the heuristic toward each landmark reduces the number of expanded states, with up to two orders of magnitude improvement. 
This is because the packages are distributed across a large grid, so a local heuristic to guide the planner toward the next landmark can greatly improve performance.
Moreover, the required number of states is much more stable than~$\HAdd$.
We note that the generalized landmark graphs are not necessarily optimal, so while no major differences in the plan length are visible, it might happen that the plan found by~$\HAdd$ is slightly shorter (generally not more than roughly 5\% of the plan length).

\begin{figure}[h]
    \centering
    \resizebox{\columnwidth}{!}{\input{ipc_domain_goal_comparison.pgf}}
    \caption{Number of expanded states (top) and plan length (bottom) for instances solved by the different heuristics.}
    \label{fig:lm:compare}
\end{figure}

Using custom generalized landmark graphs shows a big improvement in solving the three domains.
The impact varies by domain, but in each case we see fewer expanded states at a similar path length, with more instances solved across all three domains than the baseline.
These results clearly show the benefit of generalized landmarks.
The significance test results in~\autoref{tab:lm:sig} show that in all cases for \Delivery{} and \Ferry{}, the results are significant with alpha=0.05.
Because~$\HAdd$ did not solve many instances of \Childsnack{}, the results cannot be significant, though~\autoref{fig:lm:compare} showed a clear difference.
While the graphs have been constructed by hand, this does not diminish the results, as it provides a way to capture expert domain knowledge, which is highly relevant for real-world problems.

\begin{table}[t]
    \centering
\caption{Significance results from paired Wilcoxon tests and Holm-adjusted p-values. Compares the three domains on the baseline~$\HAdd$ to the landmark heuristic~$\LMHeur$ and guided landmark heuristic~$\LMHeur_{\Guide}$. Column PI gives the number of paired instances.}
\label{tab:lm:sig}
\begin{tabular}{llrrr}
\toprule
Domain & Comparison & PI & Median ratio & Adjusted $p$ \\
\midrule
ferry & $\HAdd$ vs $\LMHeur$ & 36 & 3.258333 & 0.008246 \\
ferry & $\HAdd$ vs $\LMHeur_{\Guide}$ & 49 & 5.279070 & 0.000000 \\
ferry & $\LMHeur$ vs $\LMHeur_{\Guide}$ & 39 & 1.055556 & 0.000081 \\
childsnack & $\HAdd$ vs $\LMHeur$ & 4 & 17117.500000 & 0.125000 \\
delivery & $\HAdd$ vs $\LMHeur$ & 29 & 11.500801 & 0.000000 \\
delivery & $\HAdd$ vs $\LMHeur_{\Guide}$ & 29 & 15.575875 & 0.000000 \\
delivery & $\LMHeur$ vs $\LMHeur_{\Guide}$ & 30 & 1.110804 & 0.000015 \\
\bottomrule
\end{tabular}
\end{table}

\paragraph{Ferry adaptation}
We further analyze the effect of using the guide in the \Ferry{} domain to assess its benefits.
Instead of the guide, we slightly adapt the domain definition by adding a redundant action.
We introduce the predicate~$delivered(car)$, which is never used in a precondition, but does alter the state, so the state is seen as new and not as visited. 
The redundant action~$justDelivered$ then takes as precondition: $emptyFerry$, $at(car1,loc)$, $at(car2,loc)$, and $atFerry(loc)$, with the single effect of setting~$delivered(car1)$ to \texttt{true}.
This action can then be used after delivering a car to a location where the ferry should also pick up its next passenger.
\autoref{fig:ferry} shows the expanded states and plan length when~$\LMHeur$ is used to plan the adapted \Ferry{} instances. 
We see that the difference between the guided and non-guided versions decreases in the adjusted domain compared to the original domain.
Moreover, $\LMHeur_{\Guide}$ solves one instance more than~$\LMHeur$ with the adapted domain (and a lot more with the original domain).
In both versions of the domain,~$\LMHeur$ outperforms the~$\HAdd$ baseline. 
The baseline actually performs worse on the adjusted domain, most likely because the branching factor has increased.
Based on these results, and the results for \Delivery{} if~\Cref{fig:lm:compare}, we conclude that the guide does influence the results.
However, a well-defined domain has a greater impact on the number of instances solved, especially in defining the loop condition for a domain.
The guides do provide a way to overcome domain shortcomings. 

\begin{figure}[h]
    \centering
    \resizebox{\columnwidth}{!}{\input{ferry_empty_results.pgf}}
    \caption{Number of expanded states to solve the original and adapted \Ferry{} domain.}
    \label{fig:ferry}
\end{figure}

\section{Learning Generalized Landmarks}
\label{sec:lm:learn}
This section describes the discovery process of a graph of generalized landmarks.
Graphs of generalized landmarks can be constructed manually by reasoning about a domain, and~\Cref{sec:lm:eval} showed the performance benefits of using such a well-designed graph to solve difficult instances.
Consequently, we want to generate these graphs automatically, so we propose a method to learn them from previously solved instances.
First, we provide the motivation and high-level idea, and then formalize it in~\Cref{sec:lm:learn:one,sec:lm:learn:chain}.
This method is evaluated experimentally in~\Cref{sec:lm:learn:exp}.

Our learning algorithm takes a set of solved instances from a single domain as input and, based on these, generates the state functions.
The training instances are treated as state trajectories based on the provided plan, allowing us to infer the goal and how to reach it.
The benefit of state trajectories is that we can use the states as training data to generate state functions that serve as generalized landmarks, their guides, or the loop conditions. 
Moreover, landmarks are must-reach states, so we search through the state space formed by the trajectories rather than their action plans.
This also provides a smaller search space than using state transitions, such as in policy sketches~\cite{Drexler2024}.
The set of state functions that compose the graph of generalized landmarks is generated using the DLplan library~\cite{Drexler2022b}.
Their algorithm is based on a state space that we constrain to the states in the training trajectories to keep the space manageable.
The domain definition to which the instances belong is used only to provide the predicates, including the goal predicates, and the object types used in the state functions.

The state functions and trajectories are converted to an annotated set of states, from which the landmarks are learned.
Each state in the training trajectories is evaluated using each state function in the available set.
The DLplan library provides functionality to evaluate a state on any function that it generated.
While state descriptors and state values are used differently in our framework, either returning a Boolean or numeric value, they can share the same underlying function. 
For example, say we have a function that returns all packages in \Delivery{} with a defined $\goalPredicateFunc{at}$ goal location but that are not currently $at$ that location.
We create a state descriptor indicating whether the set is empty, and a state value that returns the exact number of packages in the set. 

The main purpose of our learning algorithm is to identify a similar state across trajectories using a subset of state functions.
These states are marked as a landmark found across all trajectories.
We use an iterative process to find the earliest next state in each trajectory for which we have a set of state descriptors that identify it.
This is constrained such that i) the state directly previous to this identified state must not satisfy this state descriptor, ii) it must be strictly later than the previously found state, and iii) each trajectory must have at least one such state.
Once we have identified the landmark state in each trajectory and selected its defining state functions, we also identify a guide function.
Ideally, this must be zero in the state accepting this landmark, one in the state directly preceding it, and increase in the state farther from this landmark.

To discover loops in our graph, we check in each iteration whether the landmark~$\GeneralizedLandmark_\lmNodeIndexOne$ we are currently discovering, which is first accepted in some state~$\stateVar_\lmNodeIndexOne$, is achieved again in a later state~$\stateVarPrime_\lmNodeIndexOne$ for some of the trajectories.
If this is true, we find the earliest previously identified landmark~$\GeneralizedLandmark_\lmNodeIndexTwo$ for which any landmark~$\GeneralizedLandmark_\lmNodeIndexThree$ found between it (such that~$\forall \lmNodeIndexThree: \lmNodeIndexTwo < \lmNodeIndexThree < \lmNodeIndexOne$) is also satisfied in some state between~$\stateVar_\lmNodeIndexOne$ and~$\stateVarPrime_\lmNodeIndexOne$. 
Then, we move on to find a function that gives the number of loop landmarks for each loop in the initial state, and to determine the exit condition for each loop.

\subsection{Discovery of a Single Generalized Landmark}
\label{sec:lm:learn:one}
Given a set of state trajectories~$\TrainingTrajectories$ and the corresponding domain~$\Domain$ unified by domain goal~$\DomainGoal$, we generate a set of state descriptors and state values for this set~$\TrainingTrajectories$.
We define the formal process of discovering generalized landmark~$\GeneralizedLandmark_\lmNodeIndexOne$ below in~\autoref{eq:lm}. 
We use~$\trajectoryStateIndex^{\lmNodeIndexOne}$ as the index of the first landmark state in trajectory~$\Trajectory$ for the currently (to be) discovered landmark~$\GeneralizedLandmark_\lmNodeIndexOne$.
For now, we simplify the notation by omitting the reference index~$\lmNodeIndexOne$ to landmark~$\GeneralizedLandmark_\lmNodeIndexOne$, as the algorithm applies to one landmark at a time. 
Each such iteration considers the trajectory starting from an iteratively increasing index, indicated by the state index of the previous landmark in the respective trajectory~$\Trajectory$, denoted by~$\trajectoryStateIndexPrime$, which is set to negative one initially, as landmarks can also be detected in the initial state.

\begin{example}{Identifying one landmark}
    For example, suppose we have three trajectories for \Delivery{}, where the truck reaches the single package in state~$\stateVar_2$ for the first trajectory~$\Trajectory^1$, in state~$\stateVar_3$ for the second~$\Trajectory^2$, and in~$\stateVar_4$ for the~$\Trajectory^3$, as the truck starts further away from the package in each instance. 
    The first landmark~$\GeneralizedLandmark_1$ (At package) should identify these landmark states: $\stateIndex^{1}_{\Trajectory^1}~=~2,~\stateIndex^1_{\Trajectory^2}~=~3,~\stateIndex^1_{\Trajectory^3}~=~4$.
    So, we need one or more state descriptors that return true in these states. 
\end{example}

We use a binary \textbf{decision variable}~$\SelectDescriptor$, which is \texttt{true} if the state descriptor~$\StateDescriptor$ is included in the landmark.
Similarly, we have a decision variable~$\SelectDescriptorFunc {\neg\StateDescriptor}$, which is \texttt{true} if the state descriptor~$\neg\StateDescriptor$ is included in the landmark (i.e., when~$\StateDescriptor$ must explicitly be \texttt{false}) (\autoref{eq:desSel}).
Because each descriptor can be used as either true or false, we indicate the set of possible descriptors as~$\StateDescriptorSetDomain$, although they cannot be selected simultaneously (\autoref{eq:desDom}).
We select the landmark's state descriptors by minimizing the earliest next landmark state (i.e., a low index~$\trajectoryStateIndex$) for every trajectory~$\Trajectory \in \TrainingTrajectories$ (\autoref{eq:sI}).
This is constrained such that i)~\autoref{eq:oneF} ensures that at least one state descriptor is selected for each landmark; ii) each trajectory must have at least one state that achieves the currently discovered landmark with index~$\trajectoryStateIndex$~(\autoref{eq:oneS}); and iii) in each trajectory, the current landmark state with index~$\trajectoryStateIndex$ must occur later than index~$\trajectoryStateIndexPrime$ of the previously discovered landmark (\autoref{eq:oneS}).

To connect the state descriptors to the landmark states, we need to know which descriptor holds in which state.
We use a binary variable indicating whether a state descriptor~$\StateDescriptor$ returns \texttt{true} in state~$\stateVar$:
\begin{align}
    \AcceptVariable{\StateDescriptor}{\stateVar} =
    \begin{cases}
        1 &\mathrm{if}\quad \Accepts{\stateVar}{\StateDescriptor}\\
        0 &\mathrm{if}\quad \NotAccepts{\stateVar}{\StateDescriptor}
    \end{cases}
    \quad \quad \forall \StateDescriptor \in \StateDescriptorSetDomain,\, \forall \stateVar \in \Trajectory, \forall \Trajectory \in \TrainingTrajectories.
\end{align}
This variable is computed from the set of state descriptors and trajectories by evaluating them using DLplan.
\autoref{eq:des} formally states that each selected state descriptor with~$\SelectDescriptor = 1$ must hold in every landmark state with index~$\trajectoryStateIndex$.

Finally, the landmark state should be the first state where the landmark is encountered.
For example, the landmark `Carry package' should signal the landmark state where the package was just picked up, not after walking while holding the package.
We thus restrict discovery to select the state descriptors such that at least one of them signals a direct change. 
\autoref{eq:change} formally says that at least one state descriptor holds in the currently discovered landmark state (index $\trajectoryStateIndex:\AcceptVariable{\StateDescriptor}{\Trajectory_{\trajectoryStateIndex}}=1$), but not in the state directly prior to that landmark state (index $\trajectoryStateIndex-1:\AcceptVariable{\StateDescriptor}{\Trajectory_{\trajectoryStateIndex-1}}=0$).

\noindent
\begin{minipage}{\linewidth}
    \begin{subequations}
    \begin{align}
        &\min \sum_{\Trajectory \in \TrainingTrajectories} \trajectoryStateIndex  \label{eq:sI}\\
        &\mathrm{s.t.}\quad  \sum_{\StateDescriptor \in \StateDescriptorSetDomain} \SelectDescriptor \geq 1 \label{eq:oneF} \\
        &\hphantom{\mathrm{s.t.}}\quad \trajectoryStateIndex \in \{\trajectoryStateIndexPrime+1,\dots,\Trajectory_{\planLength}\} &\forall \Trajectory \in \TrainingTrajectories \label{eq:oneS} \\
        &\hphantom{\mathrm{s.t.}}\quad \AcceptVariable{\StateDescriptor}{\Trajectory_{\trajectoryStateIndex}} = \SelectDescriptor \quad\quad \forall \SelectDescriptor = 1 \, &\forall \Trajectory \in \TrainingTrajectories \forall \StateDescriptor \in \StateDescriptorSetDomain\label{eq:des} \\
        &\hphantom{\mathrm{s.t.}}\quad \sum_{\StateDescriptor \in \StateDescriptorSetDomain\,:\,\SelectDescriptorFunc{\StateDescriptor}=1}  (\AcceptVariable{\StateDescriptor}{\Trajectory_{\trajectoryStateIndex}} - \AcceptVariable{\StateDescriptor}{\Trajectory_{\trajectoryStateIndex - 1}}) \geq 1 &\forall \Trajectory \in \TrainingTrajectories  \label{eq:change} \\
        &\hphantom{s.t.}\quad \SelectDescriptor, \SelectDescriptorFunc{\neg\StateDescriptor} \in \{0,1\} &\forall \StateDescriptor \in \StateDescriptorSet \label{eq:desSel} \\
        &\hphantom{s.t.}\quad \forall \StateDescriptor \in \StateDescriptorSet\;:\; \StateDescriptor \in \StateDescriptorSetDomain \land \neg\StateDescriptor \in \StateDescriptorSetDomain\;:\; &\SelectDescriptorFunc{\StateDescriptor} + \SelectDescriptorFunc{\neg\StateDescriptor} \leq 1 &\label{eq:desDom}
    \end{align}
    \label{eq:lm}
    \end{subequations}
    \makebox[\linewidth]{\autoref{eq:lm}: Generalized landmark discovery for landmark~$\GeneralizedLandmark_\lmNodeIndexOne$.}
\end{minipage}

\subsection{Creating a Graph of Generalized Landmarks}
\label{sec:lm:learn:chain}
The Constraint Satisfaction Problem in~\autoref{eq:lm} defines one iteration of the generalized landmark discovery process, which can be iterated by using found indices~$\trajectoryStateIndex$ as~$\trajectoryStateIndexPrime$ in the next iteration to find a chain of landmarks. Next, we introduce the discovery process for the graph of generalized landmarks by adding loops.

Although generalized landmarks can be found without loops, loop discovery depends on the current and previously discovered landmarks, so this process is done simultaneously with landmark discovery.
A loop can be discovered if at least two trajectories in the training set show the same repeated behavior. 
We indicate the set of training trajectories that show loop behavior with~$\LoopTrajectories$.
A loop~$\LoopDef{\GeneralizedLandmark_\lmNodeIndexOne}{\GeneralizedLandmark_\lmNodeIndexTwo}$ from landmark~$\GeneralizedLandmark_\lmNodeIndexOne$ to landmark~$\GeneralizedLandmark_\lmNodeIndexTwo$ is found in the iteration where landmark~$\GeneralizedLandmark_\lmNodeIndexOne$ (the loop landmark) is discovered, and landmark~$\GeneralizedLandmark_\lmNodeIndexTwo$ was discovered previously, so~$\lmNodeIndexTwo < \lmNodeIndexOne$.

Loop landmark~$\GeneralizedLandmark_\lmNodeIndexOne$ is discovered by identifying for each trajectory~$\Trajectory \in \TrainingTrajectories$ the earliest landmark state that appears again in the same trajectory.
For example, consider trajectory~$\Trajectory^2$ in~\autoref{tab:ex:disc}. Landmark~$\GeneralizedLandmark_2$ is achieved in state~$\trajectoryStateIndex=2$ and again in state~$\trajectoryStateIndex=6$.
We use~$\LandmarkStateIndex$ to indicate the time this landmark is achieved, so for~$\LandmarkStateIndex=1:\trajectoryStateIndex=2$ and for~$\LandmarkStateIndex=2:\trajectoryStateIndex=6$, for the same landmark. 
To group them, we use the function~$\SelectedLandmarkStatesFunction{\LandmarkStateIndex}{\lmNodeIndexOne}{\Trajectory}$ to give the index of the~$\LandmarkStateIndex$'th landmark state achieving~$\GeneralizedLandmark_\lmNodeIndexOne$ in trajectory~$\Trajectory$. In this case, the second time that~$\GeneralizedLandmark_2$ is achieved in~$\Trajectory^2$ is~$\SelectedLandmarkStatesFunction{2}{2}{\Trajectory^2} = 6$.

\begin{figure}[t]
    \centering
    \begin{subfigure}{.32\textwidth}
        \centering
        \begin{tikzpicture}
            \def\cellsize{0.8cm}
            \def\iconscale{1.8}
            \foreach \x in {0,1} {
                \foreach \y in {0,1} {
                    \draw[thick] (\x*\cellsize, \y*\cellsize) rectangle ++(\cellsize, \cellsize);
                }
            }
            \node at (0.5*\cellsize, 1.5*\cellsize) {\scalebox{\iconscale}{\faTruck}};
            \node[text=white] at (0.4*\cellsize, 1.55*\cellsize) {${\mathbf{\TruckVar_1}}$};
            \node at (1.5*\cellsize, 0.5*\cellsize) {\scalebox{\iconscale}{\faFlagCheckered}};
            \node at (1.5*\cellsize, 1.5*\cellsize) {\scalebox{\iconscale}{\textcolor{col4}{\faBox}}}; 
            \node[text=textcol4] at (1.5*\cellsize, 1.35*\cellsize) {${\mathbf{\PackageVar_1}}$}; 
        \end{tikzpicture}
        \caption{Instance of trajectory~$\Trajectory^1$.}
        \label{fig:ex:1p}
    \end{subfigure}
    \begin{subfigure}{.32\textwidth}
        \centering
        \begin{tikzpicture}
            \def\cellsize{0.8cm}
            \def\iconscale{1.8}
            \foreach \x in {0,1} {
                \foreach \y in {0,1} {
                    \draw[thick] (\x*\cellsize, \y*\cellsize) rectangle ++(\cellsize, \cellsize);
                }
            }
            \node at (0.5*\cellsize, 1.5*\cellsize) {\scalebox{\iconscale}{\faTruck}};
            \node[text=white] at (0.4*\cellsize, 1.55*\cellsize) {${\mathbf{\TruckVar_1}}$};
            \node at (1.5*\cellsize, 0.5*\cellsize) {\scalebox{\iconscale}{\faFlagCheckered}};
            \node at (1.5*\cellsize, 1.5*\cellsize) {\scalebox{\iconscale}{\textcolor{col4}{\faBox}}}; 
            \node[text=textcol4] at (1.5*\cellsize, 1.35*\cellsize) {${\mathbf{\PackageVar_1}}$}; 
            \node at (0.5*\cellsize, 0.5*\cellsize) {\scalebox{\iconscale}{\textcolor{col2}{\faBox}}}; 
            \node[text=textcol2] at (0.5*\cellsize, 0.35*\cellsize) {${\mathbf{\PackageVar_2}}$}; 
        \end{tikzpicture}
        \caption{Instance of trajectory~$\Trajectory^2$.}
        \label{fig:ex:2p}
    \end{subfigure}
    \begin{subfigure}{.32\textwidth}
        \centering
        \begin{tikzpicture}
            \def\cellsize{0.8cm}
            \def\iconscale{1.8}
            \foreach \x in {0,1,2} {
                \foreach \y in {0,1} {
                    \draw[thick] (\x*\cellsize, \y*\cellsize) rectangle ++(\cellsize, \cellsize);
                }
            }
            \node at (0.5*\cellsize, 1.5*\cellsize) {\scalebox{\iconscale}{\faTruck}};
            \node[text=white] at (0.4*\cellsize, 1.55*\cellsize) {${\mathbf{\TruckVar_1}}$};
            \node at (1.5*\cellsize, 0.5*\cellsize) {\scalebox{\iconscale}{\faFlagCheckered}};
            \node at (1.5*\cellsize, 1.5*\cellsize) {\scalebox{\iconscale}{\textcolor{col4}{\faBox}}}; 
            \node[text=textcol4] at (1.5*\cellsize, 1.35*\cellsize) {${\mathbf{\PackageVar_1}}$}; 
            \node at (0.5*\cellsize, 0.5*\cellsize) {\scalebox{\iconscale}{\textcolor{col2}{\faBox}}}; 
            \node[text=textcol2] at (0.5*\cellsize, 0.35*\cellsize) {${\mathbf{\PackageVar_2}}$}; 
            \node at (2.5*\cellsize, 0.5*\cellsize) {\scalebox{\iconscale}{\textcolor{col5}{\faBox}}}; 
            \node[text=textcol5] at (2.5*\cellsize, 0.35*\cellsize) {${\mathbf{\PackageVar_3}}$}; 
        \end{tikzpicture}
        \caption{Instance of trajectory~$\Trajectory^3$.}
        \label{fig:ex:3p}
    \end{subfigure}
    \caption{Three instances for which the optimal trajectories are generated and reported in~\autoref{tab:ex:disc}.}
    \label{fig:ex:disc}
\end{figure}

\begin{table}[t]
    \centering
    \caption{Trajectories used in~\autoref{ex:disc} showing which landmarks from~\autoref{fig:lm:delex:fullgraph} are accepted in each state. The top row gives the state indices. See~\Cref{fig:ex:disc} for the instances that these trajectories solve.}
    \label{tab:ex:disc}
    \begin{tabular}{c|ccccccccccccc}
        Traj. & 0 & 1 & 2 & 3 & 4 & 5 & 6 & 7 & 8 & 9 & 10 & 11 & 12 \\
        \midrule
        $\Trajectory^1$ & $\emptyset$ & $\GeneralizedLandmark_1$ & $\GeneralizedLandmark_2$ & $\GeneralizedLandmark_3$ & $\GeneralizedLandmark_4$ & - & - & - & - & - & - & - & - \\
        $\Trajectory^2$ & $\emptyset$ & $\GeneralizedLandmark_1$ & $\GeneralizedLandmark_2$ & $\GeneralizedLandmark_3$ & $\GeneralizedLandmark_4$ & $\GeneralizedLandmark_1$ & $\GeneralizedLandmark_2$ & $\GeneralizedLandmark_3$ & $\GeneralizedLandmark_4$ & - & - & - & - \\
        $\Trajectory^3$ & $\emptyset$ & $\GeneralizedLandmark_1$ & $\GeneralizedLandmark_2$ & $\GeneralizedLandmark_3$ & $\GeneralizedLandmark_4$ & $\GeneralizedLandmark_1$ & $\GeneralizedLandmark_2$ & $\GeneralizedLandmark_3$ & $\GeneralizedLandmark_4$ & $\GeneralizedLandmark_1$ & $\GeneralizedLandmark_2$ & $\GeneralizedLandmark_3$ & $\GeneralizedLandmark_4$ \\
    \end{tabular}
\end{table}

Suppose there is such a state where landmark~$\GeneralizedLandmark_\lmNodeIndexOne$ is accepted again. 
In that case, we identify a previous landmark~$\GeneralizedLandmark_\lmNodeIndexTwo$ (where~$\lmNodeIndexTwo < \lmNodeIndexOne$) that is accepted again in some state before the earliest next landmark state of~$\GeneralizedLandmark_\lmNodeIndexOne$ such that all landmarks~$\GeneralizedLandmark_\lmNodeIndexThree$ discovered after~$\GeneralizedLandmark_\lmNodeIndexTwo$ (also including~$\GeneralizedLandmark_\lmNodeIndexTwo$) and before landmark~$\GeneralizedLandmark_\lmNodeIndexOne$ occur in some state~$\stateVarPrime$ between these two landmark states. 
This must be true for at least some trajectories~$\LoopTrajectories$ such that~$\LoopTrajectories \subseteq \TrainingTrajectories$, though not all trajectories have to exhibit loop behavior. 
We constrain the discovery so that loop behavior is present in more than one trajectory to prevent overfitting, but not necessarily all (i.e.,~$2 \leq |\LoopTrajectories| \leq |\TrainingTrajectories|$).

\begin{example}\label{ex:disc}
    Take the generalized landmark graph from~\autoref{fig:lm:delex:fullgraph} as an example, and the three training trajectories given in~\autoref{tab:ex:disc}, where the superscript is used to identify the trajectories.
    These are generated from optimally solving the three simple instances shown in~\autoref{fig:ex:disc}.    
    Suppose we are running the discovery procedure and are currently in iteration 4, thus discovering generalized landmark~$\GeneralizedLandmark_4$ as well as the loop~$\LoopDef{\GeneralizedLandmark_4}{\GeneralizedLandmark_1}$. 
    So, currently we have~$\stateIndex'_{\Trajectory^1} = \stateIndex'_{\Trajectory^2} = \stateIndex'_{\Trajectory^3} = 3$ because in each trajectory the previous landmark~$\GeneralizedLandmark_3$ was discovered in the state with index 3. We also have the indices of all states that achieve the landmark~$\GeneralizedLandmark_{\lmNodeIndexOne} = \GeneralizedLandmark_4$. In the first trajectory, there is only one state~$\SelectedLandmarkStatesFunction{\LandmarkStateIndex=1}{\lmNodeIndexOne=4}{\Trajectory^1} = 4$. In the second trajectory, we have~$\SelectedLandmarkStatesFunction{\LandmarkStateIndex=1}{\lmNodeIndexOne=4}{\Trajectory^2} = 4$ and~$\SelectedLandmarkStatesFunction{\LandmarkStateIndex=2}{\lmNodeIndexOne=4}{\Trajectory^2} = 8$. The third trajectory achieves this landmark three times:~$\SelectedLandmarkStatesFunction{\LandmarkStateIndex=1}{\lmNodeIndexOne=4}{\Trajectory^3} = 4$,~$\SelectedLandmarkStatesFunction{\LandmarkStateIndex=2}{\lmNodeIndexOne=4}{\Trajectory^3 = 8}$, and~$\SelectedLandmarkStatesFunction{\LandmarkStateIndex=3}{\lmNodeIndexOne=4}{\Trajectory^3} = 12$.
    To construct the loop, we want to find the previous landmark~$\GeneralizedLandmark_\lmNodeIndexTwo$ that is achieved again before the current loop landmark~$\GeneralizedLandmark_\lmNodeIndexOne=\GeneralizedLandmark_4$ is achieved again, so between the landmark states identified by~$\SelectedLandmarkStatesFunction{\LandmarkStateIndex=1}{\lmNodeIndexOne=4}{\Trajectory}$ and ~$\SelectedLandmarkStatesFunction{\LandmarkStateIndex=2}{\lmNodeIndexOne=4}{\Trajectory}$ for each trajectory, as well as between~$\SelectedLandmarkStatesFunction{\LandmarkStateIndex=2}{\lmNodeIndexOne=4}{\Trajectory}$ and~$\SelectedLandmarkStatesFunction{\LandmarkStateIndex=3}{\lmNodeIndexOne=4}{\Trajectory}$ (only for~$\Trajectory^3)$.
    Judging from~\autoref{tab:ex:disc}, all landmarks~$\GeneralizedLandmark_1$,~$\GeneralizedLandmark_2$, and~$\GeneralizedLandmark_3$ are previous landmarks that the loop edge starting from~$\GeneralizedLandmark_4$ can loop back to (as they are all achieved again in between), where selecting~$\GeneralizedLandmark_1$ creates the longest landmark chain in the loop.
    So, we identify the loop~$\LoopDef{\GeneralizedLandmark_4}{\GeneralizedLandmark_1}$.
\end{example}

We want the chain of landmarks to be as long as possible to capture repeated behavior. So, we take the earliest landmark~$\GeneralizedLandmark_\lmNodeIndexTwo$ to form the chain of landmarks~$[\GeneralizedLandmark_\lmNodeIndexTwo, \dots, \GeneralizedLandmark_\lmNodeIndexOne]$ that does not contain any landmark twice. 
This chain is thus achieved at least twice within the trajectories in~$\LoopTrajectories$.
Next, we identify the loop condition~$\LoopConditionsDef$. 
Without these conditions, we cannot say when to traverse a loop and when to continue along the chain of landmarks. 
These conditions are thus necessary for the usage of generalized landmarks in a planner, and a loop~$\LoopDef{\GeneralizedLandmark_\lmNodeIndexOne}{\GeneralizedLandmark_\lmNodeIndexTwo}$ is only valid if the loop condition~$\LoopConditionsDef$ can be identified.

The loop discovery is formalized in~\autoref{eq:loop}, where we select the state functions for the loop exit condition and the loop landmark counter by introducing \textbf{loop decision variables}~$\SelectStateFunction{}$ (\autoref{eq:loop:var}).
For example,~$\SelectStateFunction{\StateDescriptor}$ says that state descriptor~$\StateDescriptor \in \StateDescriptorSetDomain$ is selected for the current loop condition (\autoref{eq:loop:des}).
The loop landmark counter gives the number of times~$\StateValue(\InitialStateVar)$ that a loop landmark can be achieved, which must thus hold for every~$\StateValue \in \LoopLandmarkCounter$.
Since the counter value varies by instance, we determine it in the initial state~$\InitialStateVar$ to ensure proper landmark progression. 
If an instance does not exhibit loop behavior, this value is one because each landmark must be achieved at least once.
For example, in the example trajectories,~$\StateValue(\InitialStateVar^1)=1$ (no loop),~$\StateValue(\InitialStateVar^2)=2$ (one loop traversal, two loop landmarks, and~$\StateValue(\InitialStateVar^3)=3$.
The value given by this counter is used to denote the index of the last landmark state~$\SelectedLandmarkStatesFunction{\StateValue(\InitialStateVar)}{\lmNodeIndexOne}{\Trajectory}$ in trajectory~$\Trajectory$ for landmark~$\GeneralizedLandmark_\lmNodeIndexOne$ where the loop starting from loop landmark~$\GeneralizedLandmark_\lmNodeIndexOne$ can no longer be traversed, and thus the exit condition must hold.

We want to find the smallest set of state functions to define the loop condition (\autoref{eq:minF}) while making sure that at least one is selected per category (\autoref{eq:one}).
First, the state values~$\StateValue$ composing the loop landmark counter~$\LoopLandmarkCounter$ must identify the number of loop landmark states for all trajectories~$\TrainingTrajectories$ (\autoref{eq:loop:val}).
The function~$\SelectedLandmarkStatesFunction{\LandmarkStateIndex}{\lmNodeIndexOne}{\Trajectory}$ must thus always return a valid index for this trajectory for each of these~$\LandmarkStateIndex \in [1,\dots,\StateValue(\InitialStateVar)]$ landmark states (\autoref{eq:count}). 
In addition, the loop condition is only valid if there is a loop in which all intermediate landmarks~$\GeneralizedLandmark_\lmNodeIndexThree$ between~$\GeneralizedLandmark_\lmNodeIndexTwo$ and~$\GeneralizedLandmark_\lmNodeIndexOne$ are achieved again.
\autoref{eq:doLoop} ensures this, and also guarantees that the loop landmark states returned by~$\SelectedLandmarkStatesVariable$ are distinct for loop trajectories~$\Trajectory \in \LoopTrajectories$: $\SelectedLandmarkStatesFunction{\LandmarkStateIndex-1}{\lmNodeIndexOne}{\Trajectory} < \SelectedLandmarkStatesFunction{\LandmarkStateIndex}{\lmNodeIndexOne}{\Trajectory}$.
Finally, we select the state descriptors for the exit condition~$\ExitCondition$.
The selected state descriptors must hold in the last state that reaches the loop landmark (\autoref{eq:exit}) and must not hold in any of the other states achieving loop landmark~$\GeneralizedLandmark_\lmNodeIndexOne$ (\autoref{eq:exitLoop}): either~$\SelectStateFunction{\StateDescriptor}=0$ or~$\AcceptVariable{\StateDescriptor}{\stateVar}=0$. 

\noindent
\begin{minipage}[t]{\linewidth}
    \begin{subequations}
    \begin{align}
        \min &\sum_{\StateDescriptor \in \StateDescriptorSetDomain} \SelectStateFunction{\StateDescriptor} + \sum_{\StateValue \in \StateValueSet} \SelectStateFunction{\StateValue} \label{eq:minF}\\
        \mathrm{s.t.} &\sum_{\StateDescriptor \in \StateDescriptorSetDomain} \SelectStateFunction{\StateDescriptor} \geq 1,\; \sum_{\StateValue \in \StateValueSet} \SelectStateFunction{\StateValue} \geq 1 \label{eq:one} \\
        &\SelectedLandmarkStatesFunction{\LandmarkStateIndex}{\lmNodeIndexOne}{\Trajectory} \in [0,\dots,|\Trajectory|] 
        &&\begin{aligned}
            &\forall \LandmarkStateIndex \in [1,\dots,\StateValue(\InitialStateVar)],\\
            &\forall \StateValue \in \StateValueSet,\; \forall \Trajectory \in \TrainingTrajectories 
        \end{aligned}
        \label{eq:count} \\
        &\SelectedLandmarkStatesFunction{\LandmarkStateIndex-1}{\lmNodeIndexOne}{\Trajectory} < \SelectedLandmarkStatesFunction{\LandmarkStateIndex}{\lmNodeIndexTwo}{\Trajectory} < \SelectedLandmarkStatesFunction{\LandmarkStateIndex}{\lmNodeIndexThree}{\Trajectory} < \SelectedLandmarkStatesFunction{\LandmarkStateIndex}{\lmNodeIndexOne}{\Trajectory}
        &&\begin{aligned}
            &\forall \lmNodeIndexThree \in \langle\lmNodeIndexTwo,..,\lmNodeIndexOne \rangle,\, \forall \LandmarkStateIndex \in [2,\dots,\StateValue(\InitialStateVar)], \\
            &\forall \Trajectory \in \LoopTrajectories,\,\, \forall \StateValue \in \StateValueSet : \SelectStateFunction{\StateValue}=1
        \end{aligned}
        \label{eq:doLoop}\\
        &\SelectStateFunction{\StateDescriptor} = \AcceptVariable{\StateDescriptor}{\Trajectory_{\LandmarkStateFunctionIndex}} \quad\quad:\, \LandmarkStateFunctionIndex = \SelectedLandmarkStatesFunction{\StateValue(\InitialStateVar)}{\lmNodeIndexOne}{\Trajectory}
        &&\begin{aligned}
            &\forall \Trajectory \in \LoopTrajectories,\, \StateDescriptor \in \StateDescriptorSetDomain,\, \StateValue \in \StateValueSet
        \end{aligned}
        \label{eq:exit}\\
        &\SelectStateFunction{\StateDescriptor} + \AcceptVariable{\StateDescriptor}{\Trajectory_\LandmarkStateFunctionIndex} \leq 1 \,:\, \LandmarkStateFunctionIndex = \SelectedLandmarkStatesFunction{\LandmarkStateIndex}{\lmNodeIndexOne}{\Trajectory}
        &&\begin{aligned}
            &\forall \LandmarkStateIndex \in [1,\dots,\StateValue(\InitialStateVar)],\; \\ 
            &\forall \Trajectory \in \LoopTrajectories, \StateDescriptor \in \StateDescriptorSetDomain, \StateValue \in \StateValueSet
        \end{aligned}
        \label{eq:exitLoop}\\
        &\StateValue(\InitialStateVar) = \big|\,\SelectedLandmarkStatesFunction{}{\lmNodeIndexOne}{\Trajectory}\big| < |\Trajectory| &&\forall \Trajectory \in \TrainingTrajectories,\, \StateValue \in \StateValueSet \label{eq:loop:val}\\
        &\SelectStateFunction{\StateDescriptor} \in \{0,1\}, \SelectStateFunction{\StateValue} \in \{0,1\} &&\forall \StateDescriptor \in \StateDescriptorSetDomain,\,  \StateValue \in \StateValueSet \label{eq:loop:var} \\
        &\SelectStateFunction{\StateDescriptor} + \SelectStateFunction{\neg\StateDescriptor} \leq 1 &&\forall \StateDescriptor \in \StateDescriptorSet \label{eq:loop:des}
    \end{align}
    \label{eq:loop}
    \end{subequations}
    \makebox[\linewidth]{\autoref{eq:loop}: Discovery of loop~$\LoopDef{\GeneralizedLandmark_\lmNodeIndexOne}{\GeneralizedLandmark_\lmNodeIndexTwo}$ between generalized landmarks~$\GeneralizedLandmark_\lmNodeIndexOne$ and~$\GeneralizedLandmark_\lmNodeIndexTwo$.}
\end{minipage}

\section{Experimental Results for Discovery}
\label{sec:lm:learn:exp}
We now evaluate the learning method presented in~\Cref{sec:lm:learn}. 
First, we present details of how we implemented the Constraint Satisfaction Problem problem, and then we report our findings. 
The full implementation is available online~\cite{LandmarkCode2025}.

\subsection{Implementation of Discovery Method}
\label{sec:lm:learn:exp:imp}
Our implementation accepts problems defined in PDDL, including numeric functions.
We take a domain description and a set of its problem instances from the IPC benchmark set with annotated domain goals~\cite{Grundke2025code}.
We use a few instances from the training set to create our training trajectories, using FD~\cite{Richter2010} and ENHSP~\cite{Scala2016} to generate plans for the classic and numeric problems, respectively.
For each PDDL problem domain, we assign a set of small training instances based on the given problem instances, ensuring they share the same domain goal.
To create the training set~$\TrainingTrajectories$, we use planners from the unified planning~\cite{Micheli2025} and SymbolicPlanners.jl~\cite{ZhiXuan2022a} libraries to find a plan for the training instances, which are then converted to state trajectories, a function implemented in SymbolicPlanners.jl. We construct the state space from the training trajectories for the problem set~$\TrainingTrajectories$. 

The state functions that we derive are based on \textbf{features}, introduced by~\textcite{Frances2021}, which are defined using description logic~\cite{Baader2003}. A feature is a function of state that returns either a numeric or a Boolean value. To create a state descriptor, we evaluate the feature function as a Boolean, meaning the numeric feature is either zero or positive. State values return the numeric value of a feature function. 
As an example, we showed this in~\Cref{sec:lm:funcs} in~\autoref{eq:func:loop}.

To start the discovery process, we generate a feature pool~$\FeaturePool{}$ based on the constructed state space of the trajectories~$\TrainingTrajectories$ of a given domain~$\Domain$ using the DLplan library~\cite{Drexler2022b}.
The generation requires the state space of the problem set as input. It returns a feature pool $\FeaturePool{} = \{ \BooleanFeatureSet \cup \NumericFeatureSet \}$ of Boolean features~$\BooleanFeatureSet$ and numeric features~$\NumericFeatureSet$ describing the state space of the domain~$\Domain$. The features in this pool are defined using \emph{description logic}~\cite{Baader2003}, which can be unintuitive to interpret, and we use the DLplan library to evaluate features over the states in the state trajectories, yielding a Boolean value for each feature at each state. We let DLplan generate a finite number of features, which can be either Boolean or numeric, even though we mostly use numeric features. 
We initialize the feature limit to 1000 times the number of non-static (plus the goal) predicates in the domain.
The DLplan library takes a configuration as input that specifies the maximum feature complexity, the time limit, the maximum number of features to generate, and additional settings for specific feature types.

The training set~$\TrainingTrajectories$ and feature pool~$\FeaturePool{}$ comprise the input required for the discovery algorithm.
First, we preprocess the feature pool because the generated pool contains many duplicates and similar features. 
We defined three rules for preprocessing.
\begin{enumerate}
    \item Remove features that have the same feature valuation in every state for some trajectory, as these never change value and are thus irrelevant to landmarks.
    \item Remove features that always have the same feature valuation as another feature in every state.
    \item Remove features that always have the opposite feature valuation as another feature in every state. 
\end{enumerate}
For the latter two conditions, ties are broken in favor of less complex features. The complexity of a feature is given by the feature pool generation method. It refers to the combination of concepts and roles that comprise the feature, which are terms from description logic.
This reduced feature pool is then used to identify the state descriptors defining the generalized landmarks.
In the rest of this section, we refer to the \emph{feature pool}~$\FeaturePool{}$ used for training and to the individual \emph{state functions} it includes. 
The set of preprocessing rules chosen is referred to as~$\Preprocessing$.

\paragraph{Answer Set Programming Model}
We implement the Constraint Satisfaction Problem as an Answer Set Program (ASP) in \texttt{clingo}~\cite{Gebser2012}.
The full implementation of the model is included in our codebase~\cite{LandmarkCode2025}.
The iterative process calls the ASP model for each landmark, given an updated set of previously found landmarks and their associated landmark states.
Each iteration, we minimize the next state that identifies a landmark. 
We process the information from the solution to construct the graph of generalized landmarks. 

\subsection{Experimental Setup}
We test on nine domains in the IPC benchmark set from~\textcite{Grundke2025code} and twelve other domains. The nine IPC benchmarks have distinct sets of training and testing, and we used either 5, 10, 15, or 20 of the easiest training instances when available. The other twelve domains have five handcrafted training instances and 30 testing instances.
The plans to train on are generated with Fast Downward (FD).
The complexity of state functions that can be generated varies between 7, 9, 11, 13, and 15.
We evaluate four different variations of the preprocessing rules described above: using only rule~1 ($\Preprocessing_1$), using only rule~2 ($\Preprocessing_2$), using rules~2~and~3 ($\Preprocessing_3$), or using all three~($\Preprocessing_4$).

The experiments were run on DelftBlue~\cite{DHPC2024}, where each discovery process was allocated 32GB of memory and 7200 seconds of runtime. 
The reported learning runtime also includes pre- and post-processing. 
Then, the successful graphs were used to plan all available testing instances for that domain.
We compared~$\HAdd$ to~$\LMHeur$ both with and without guides (when present). 
Each heuristic was given~1800 seconds and~8GB of available memory to solve a single instance.

\subsection{Experimental Results}
\label{sec:lm:learn:exp:res}
\begin{figure}[!h]
    \centering
    \resizebox{\columnwidth}{!}{\input{discovery_outcomes_by_domain.pgf}}
    \caption{For each domain, for a varying number of training problems, for a varying state function complexity, we show whether the discovery process identified a graph of generalized landmarks with loops. The top shows the domains from the IPC benchmark set, and the bottom shows another set of benchmarks, including~\Delivery{}.}
    \label{fig:lm:disc}
\end{figure}

\autoref{fig:lm:disc} shows the results for 20 domains (leaving out an adaptation of grid), where each graph was trained on just five instances, and we show only preprocessing with single rules~$\Preprocessing_1$ and~$\Preprocessing_2$. The full results are reported in~\autoref{app:lm:disc-full}.
The barman domain failed with more complex state functions, and two other domains failed on all possibilities. 
Only two out of twenty domains never find a landmark and result in an empty graph.
On the other hand, just five domains found a landmark graph with a loop. 
For five out of twenty domains, a landmark graph including a loop is learned from just the set of five training instances.

We evaluate the performance of the learned generalized landmark graphs when planning with the~$\LMHeur$ heuristic.
As the presence of loops is crucial for using generalized landmarks to provide the heuristic an advantage, we only consider the planning capacities of landmark graphs that include at least one loop. 
We further narrowed the tested landmark graphs to those learned with~$\Preprocessing_1$ or~$\Preprocessing_2$; with state function complexity~7,~9, or~11; and with 5 training trajectories.
\autoref{tab:discovery-evaluation} shows the planning results and compares them to~$\HAdd$.
We see that the graphs learned from less complex state functions often solve more instances than graphs learned from more complex functions. 
This probably relates to the learning algorithm's bilevel nature that simultaneously minimizes the next landmark and selects the state functions. 
With less complex functions, it might be easier to detect a loop and find more landmarks.
The newspapers domain is the only one where~$\LMHeur$ was faster than~$\HAdd$ for the same instances and solved more instances. 
For this domain,~$\HAdd$ expanded many states and quickly ended up in memory issues, where~$\LMHeur$ ran into timeouts before requiring more than 8GB of memory space.
While the domain seems relatively simple,~$\HAdd$ has a hard time because, optimally, all papers are first collected and then dropped off one by one, but more often planners will try to pick up and deliver the papers one by one, resulting in longer plans and more expanded states.
The only domain that results in failures is spanner, the only one with dead-end states.
Most domains encountered out-of-memory errors or timeouts. 
A more efficient implementation of the heuristic might resolve some of these.
Ideally, the landmarks should provide sufficient pruning when well-defined to overcome memory overhead.
We hardly see the benefit of the learned guides.
For the delivery domain, the guide often solved one fewer instance, though it required less runtime.
In other domains, the guides often increase runtime and only rarely solve one additional instance.

\begin{longtable}[p]{lp{0.1cm}p{0.2cm}p{0.17cm}lrrrrrrr}
\caption{Evaluator results for discovered landmark graphs, indexed by state function complexity (C), training-problem count, preprocessing, and whether the heuristic uses guides. The $\HAdd$ S subcolumn reports the number of instances solved by $\HAdd$. Runtime averages use instances solved by both the landmark-graph planner~$\LMHeur$ and $\HAdd$. Status can be success (S), failure (F), timeout (T/O), or out-of-memory (O/M). The news domain is reported as `news'.} \label{tab:discovery-evaluation} \\
\toprule
 &  &  &  &  & S & F & T/O & O/M & \multicolumn{1}{c}{$\timeVar$} & \multicolumn{2}{c}{$\HAdd$} \\
$\Domain$ (\#inst) & C & $|\TrainingTrajectories|$ & $\Preprocessing$ & $\LMHeur_{(\Guide)}$ &  &  &  &  &  & S & $\timeVar$ \\
\midrule
\endfirsthead
\caption[]{Evaluator results for discovered landmark graphs, indexed by state function complexity (C), training-problem count, preprocessing, and whether the heuristic uses guides. The $\HAdd$ S subcolumn reports the number of instances solved by $\HAdd$. Runtime averages use instances solved by both the landmark-graph planner~$\LMHeur$ and $\HAdd$. Status can be success (S), failure (F), timeout (T/O), or out-of-memory (O/M). The news domain is reported as `news'.} \\
\toprule
 &  &  &  &  & S & F & T/O & O/M & \multicolumn{1}{c}{$\timeVar$} & \multicolumn{2}{c}{$\HAdd$} \\
$\Domain$ (\#inst) & C & $|\TrainingTrajectories|$ & $\Preprocessing$ & $\LMHeur_{(\Guide)}$ &  &  &  &  &  & S & $\timeVar$ \\
\midrule
\endhead
\midrule
\multicolumn{12}{r}{Continued on next page} \\
\midrule
\endfoot
\bottomrule
\endlastfoot
\multirow[t]{6}{*}{baking (30)} & \multirow[t]{2}{*}{7} & \multirow[t]{2}{*}{5} & \multirow[t]{2}{*}{$\Preprocessing_1$} & $\Guide$ & 4 & 0 & 23 & 3 & 179.64 & 8 & 14.23 \\
 &  &  &  & $\LMHeur$ & 3 & 0 & 22 & 5 & 37.93 & 8 & 14.09 \\
\cmidrule{2-12} \cmidrule{3-12} \cmidrule{4-12}
 & \multirow[t]{2}{*}{9} & \multirow[t]{2}{*}{5} & \multirow[t]{2}{*}{$\Preprocessing_1$} & $\Guide$ & 4 & 0 & 22 & 4 & 180.29 & 8 & 14.23 \\
 &  &  &  & $\LMHeur$ & 4 & 0 & 22 & 4 & 179.21 & 8 & 14.23 \\
\cmidrule{2-12} \cmidrule{3-12} \cmidrule{4-12}
 & \multirow[t]{2}{*}{11} & \multirow[t]{2}{*}{5} & \multirow[t]{2}{*}{$\Preprocessing_1$} & $\Guide$ & 4 & 0 & 24 & 2 & 179.55 & 8 & 14.23 \\
 &  &  &  & $\LMHeur$ & 4 & 0 & 22 & 4 & 178.91 & 8 & 14.23 \\
\cmidrule{1-12} \cmidrule{2-12} \cmidrule{3-12} \cmidrule{4-12}
\multirow[t]{12}{*}{delivery (30)} & \multirow[t]{4}{*}{7} & \multirow[t]{4}{*}{5} & \multirow[t]{2}{*}{$\Preprocessing_1$} & $\Guide$ & 11 & 0 & 19 & 0 & 97.58 & 29 & 13.79 \\
 &  &  &  & $\LMHeur$ & 12 & 0 & 18 & 0 & 204.53 & 29 & 13.81 \\
\cmidrule{4-12}
 &  &  & \multirow[t]{2}{*}{$\Preprocessing_2$} & $\Guide$ & 11 & 0 & 19 & 0 & 98.26 & 29 & 13.79 \\
 &  &  &  & $\LMHeur$ & 12 & 0 & 18 & 0 & 207.60 & 29 & 13.81 \\
\cmidrule{2-12} \cmidrule{3-12} \cmidrule{4-12}
 & \multirow[t]{4}{*}{9} & \multirow[t]{4}{*}{5} & \multirow[t]{2}{*}{$\Preprocessing_1$} & $\Guide$ & 10 & 0 & 20 & 0 & 29.09 & 29 & 16.57 \\
 &  &  &  & $\LMHeur$ & 11 & 0 & 19 & 0 & 90.44 & 29 & 17.02 \\
\cmidrule{4-12}
 &  &  & \multirow[t]{2}{*}{$\Preprocessing_2$} & $\Guide$ & 11 & 0 & 19 & 0 & 28.34 & 29 & 23.13 \\
 &  &  &  & $\LMHeur$ & 11 & 0 & 19 & 0 & 89.26 & 29 & 17.02 \\
\cmidrule{2-12} \cmidrule{3-12} \cmidrule{4-12}
 & \multirow[t]{4}{*}{11} & \multirow[t]{4}{*}{5} & \multirow[t]{2}{*}{$\Preprocessing_1$} & $\Guide$ & 9 & 0 & 8 & 13 & 90.78 & 29 & 13.55 \\
 &  &  &  & $\LMHeur$ & 9 & 0 & 8 & 13 & 86.91 & 29 & 21.88 \\
\cmidrule{4-12}
 &  &  & \multirow[t]{2}{*}{$\Preprocessing_2$} & $\Guide$ & 7 & 0 & 0 & 23 & 28.53 & 29 & 13.49 \\
 &  &  &  & $\LMHeur$ & 7 & 0 & 0 & 23 & 26.92 & 29 & 13.49 \\
\cmidrule{1-12} \cmidrule{2-12} \cmidrule{3-12} \cmidrule{4-12}
\multirow[t]{6}{*}{ferry (57)} & \multirow[t]{4}{*}{7} & \multirow[t]{4}{*}{5} & \multirow[t]{2}{*}{$\Preprocessing_1$} & $\Guide$ & 19 & 0 & 7 & 31 & 269.00 & 49 & 16.36 \\
 &  &  &  & $\LMHeur$ & 18 & 0 & 7 & 32 & 163.65 & 49 & 15.56 \\
\cmidrule{4-12}
 &  &  & \multirow[t]{2}{*}{$\Preprocessing_2$} & $\Guide$ & 18 & 0 & 6 & 33 & 250.82 & 49 & 16.41 \\
 &  &  &  & $\LMHeur$ & 18 & 0 & 8 & 31 & 158.41 & 49 & 15.56 \\
\cmidrule{2-12} \cmidrule{3-12} \cmidrule{4-12}
 & \multirow[t]{2}{*}{9} & \multirow[t]{2}{*}{5} & \multirow[t]{2}{*}{$\Preprocessing_2$} & $\Guide$ & 8 & 0 & 0 & 49 & 30.27 & 49 & 15.25 \\
 &  &  &  & $\LMHeur$ & 8 & 0 & 0 & 49 & 28.73 & 49 & 15.25 \\
\cmidrule{1-12} \cmidrule{2-12} \cmidrule{3-12} \cmidrule{4-12}
\multirow[t]{12}{*}{news (30)} & \multirow[t]{4}{*}{7} & \multirow[t]{4}{*}{5} & \multirow[t]{2}{*}{$\Preprocessing_1$} & $\Guide$ & 13 & 0 & 12 & 5 & 24.96 & 10 & 198.31 \\
 &  &  &  & $\LMHeur$ & 13 & 0 & 10 & 7 & 23.02 & 10 & 198.31 \\
\cmidrule{4-12}
 &  &  & \multirow[t]{2}{*}{$\Preprocessing_2$} & $\Guide$ & 13 & 0 & 10 & 7 & 24.50 & 10 & 198.31 \\
 &  &  &  & $\LMHeur$ & 13 & 0 & 10 & 7 & 23.35 & 10 & 198.31 \\
\cmidrule{2-12} \cmidrule{3-12} \cmidrule{4-12}
 & \multirow[t]{4}{*}{9} & \multirow[t]{4}{*}{5} & \multirow[t]{2}{*}{$\Preprocessing_1$} & $\Guide$ & 13 & 0 & 7 & 10 & 26.98 & 10 & 198.31 \\
 &  &  &  & $\LMHeur$ & 13 & 0 & 7 & 10 & 25.22 & 10 & 198.31 \\
\cmidrule{4-12}
 &  &  & \multirow[t]{2}{*}{$\Preprocessing_2$} & $\Guide$ & 13 & 0 & 5 & 12 & 25.85 & 10 & 198.31 \\
 &  &  &  & $\LMHeur$ & 13 & 0 & 7 & 10 & 23.65 & 10 & 198.31 \\
\cmidrule{2-12} \cmidrule{3-12} \cmidrule{4-12}
 & \multirow[t]{4}{*}{11} & \multirow[t]{4}{*}{5} & \multirow[t]{2}{*}{$\Preprocessing_1$} & $\Guide$ & 9 & 0 & 0 & 21 & 81.84 & 10 & 105.37 \\
 &  &  &  & $\LMHeur$ & 9 & 0 & 0 & 21 & 79.61 & 10 & 105.37 \\
\cmidrule{4-12}
 &  &  & \multirow[t]{2}{*}{$\Preprocessing_2$} & $\Guide$ & 9 & 0 & 0 & 21 & 80.32 & 10 & 105.37 \\
 &  &  &  & $\LMHeur$ & 9 & 0 & 0 & 21 & 78.38 & 10 & 105.37 \\
\cmidrule{1-12} \cmidrule{2-12} \cmidrule{3-12} \cmidrule{4-12}
\multirow[t]{4}{*}{spanner (30)} & \multirow[t]{4}{*}{11} & \multirow[t]{4}{*}{5} & \multirow[t]{2}{*}{$\Preprocessing_1$} & $\Guide$ & 22 & 8 & 0 & 0 & 30.17 & 24 & 14.90 \\
 &  &  &  & $\LMHeur$ & 22 & 8 & 0 & 0 & 28.27 & 24 & 14.90 \\
\cmidrule{4-12}
 &  &  & \multirow[t]{2}{*}{$\Preprocessing_2$} & $\Guide$ & 22 & 8 & 0 & 0 & 32.54 & 24 & 14.90 \\
 &  &  &  & $\LMHeur$ & 22 & 8 & 0 & 0 & 30.89 & 24 & 14.90 \\
\cmidrule{1-12} \cmidrule{2-12} \cmidrule{3-12} \cmidrule{4-12}
\end{longtable}

\subsection{Customized Learning}
\label{sec:lm:learn:exp:cust}
Next, we experimented with further customizing the learning process.
We manually created five training instances for each domain and used Fast Downward (FD) to generate the plans. 
Additionally, we manually altered the plans to contain expected patterns (based on the custom landmark graphs). The resulting plans were often slightly longer.

For \Delivery{}, the five instances are all on a 3x3 grid, with either 2 or 3 packages to be delivered. The custom plans changed the delivery order for some packages. 
The graph learned from the FD plans has four landmarks, each with a guide and a loop from the fourth to the first landmark, and the graph was learned in 26.4 seconds. The graph learned from the manual plans is very similar, but in the third and fourth landmarks, an extra state descriptor is included, and the graph was produced in 28.4 seconds.
For \Childsnack{}, the five instances each had 4 children distributed across the tables, with either a 2/2 or 1/3 allergic/non-allergic distribution. 
The custom plans ensured a sandwich was made, brought to a child, and then the next sandwich was made. 
The graph learned from the FD plans has three landmarks and no loops, and was produced in 73.7 seconds. 
The graph learned from the custom plans has eight landmarks and three loops (2,1), (3,1), and (4,1); and was learned in 547.4 seconds. 
For \Ferry{}, the custom instances ensured that the goal location of one car was not the starting location of another. Each instance has three or four cars. For this domain, no further custom plans were needed.
The graph learned from the FD plans has four landmarks and three loops (2,1), (3,1), and (4,1), and was produced in 22.9 seconds. 

We reran the experiment described in~\Cref{sec:lm:learn:exp:res} for the~$\LMHeur$ heuristic with these landmark graphs. Each heuristic setting was allocated 8GB of memory and 1800 seconds of runtime to solve each single instance. 
\autoref{tab:learned} shows the results from this experiment.
Failures happen when the state progression enters a loop, thereby emptying its priority queue, which was a bad decision. 
For \Delivery{}, we see that the manual plans provided more information, which improved the heuristic performance, but for \Childsnack{}, this did not have an impact. 
Both the manual \Childsnack{} and~FD-computed \Ferry{} graphs contain several nested loops that our heuristic is currently not optimized for, resulting in improper landmark progression.
We want to address this in future work.

\begin{table}[t]
    \centering
    \caption{Solved instances by $\LMHeur$ heuristic with learned landmark graphs, trained on Fast-Downward plans (FD) or on custom-made plans (manual). For the three domains, we show the number of instances that were successfully solved (S), failed (F), timed out (T/O), or ran out of memory (O/M). For each variant of the landmark graph, we show the compute the average planning time in seconds for instances solved by both this $\LMHeur$ variant and $\HAdd$, reporting the average time for both $\LMHeur$ (Time column) and $\HAdd$ ($\HAdd$ column).}
    \label{tab:learned}
    \begin{tabular}{llrrrrrr}
        \toprule
        &  & S & F & T/O & O/M & Time & $\HAdd$ \\
        \midrule
        \multirow[t]{2}{*}{childsnack (30)} & FD & 2 & 0 & 8 & 20 & 95.60 & 60.96 \\
        & manual & 2 & 0 & 0 & 28 & 135.67 & 60.96 \\
        \cmidrule{1-8}
        \multirow[t]{2}{*}{delivery (30)} & FD & 6 & 8 & 16 & 0 & 35.14 & 13.53 \\
        & manual & 10 & 0 & 20 & 0 & 26.99 & 16.18 \\
        \cmidrule{1-8}
        ferry (57) & FD & 11 & 0 & 5 & 41 & 26.12 & 14.74 \\
        \cmidrule{1-8}
        \bottomrule
    \end{tabular}
\end{table}

\subsection{Comparison to Traditional Landmarks and Policy Sketches}
\label{subsec:lm:compare}
This section compares generalized landmarks with traditional landmarks and policy sketches.
We perform these experiments for only one domain, specifically for the  \Delivery{} domain, which we have used as the running example throughout this paper, and for which the discovery procedure returned meaningful results. It was also used in a previous paper on generalized planning, introducing a general policy for the domain, along with a clear interpretation of the state functions. 
This section will answer the following questions.
\begin{enumerate}[label=\textbf{Q\arabic*}]
    \item\label{q:lm:trad:info} Do generalized landmarks uncover information that traditional landmarks cannot capture for the \Delivery{} domain?
    \item\label{q:lm:sk:info} Do generalized landmarks reveal different information about a general planning problem than policy sketches for the \Delivery{} domain?
    \item\label{q:lm:sk:comp} Can generalized landmarks solve more instances than policy sketches for the \Delivery{} domain?
\end{enumerate}
We first introduce the generalized landmark graphs used in this section. 
The graph explained in this paper, already shown in~\autoref{fig:lm:delex:fullgraph}, is also used in these experiments and is referred to as~$\LandmarkGraphExample$. 
Moreover, we also use the discovered generalized landmark graph, which was computed using complexity 11 and 5 training instances, resulting in a graph with four nodes, with one loop from the last to the first. While each landmark has a guide, these are simple state descriptors that will return either true or false, so they are not very helpful. The loop exit condition is having all packages at their target cell. 
We refer to this graph as~$\LandmarkGraphDisc$.

\begin{figure}[t]
    \centering
\begin{tikzpicture}[>=Stealth]
\tikzset{landmark label/.style={font=\fontsize{8}{9}\selectfont,align=center,fill=white,inner sep=1pt}}
\definecolor{edge0}{HTML}{F58518}
\definecolor{edge1}{HTML}{4C78A8}
\definecolor{edge2}{HTML}{E45756}
\definecolor{edge3}{HTML}{B279A2}
\definecolor{edge4}{HTML}{54A24B}
\coordinate (v0) at (3.823cm,5.976cm);
\coordinate (v1) at (1.699cm,1.093cm);
\coordinate (v2) at (6.878cm,3.871cm);
\coordinate (v3) at (1.043cm,4.545cm);
\coordinate (v4) at (1.094cm,2.694cm);
\coordinate (v5) at (5.768cm,4.587cm);
\coordinate (v6) at (6.594cm,2.339cm);
\coordinate (v7) at (5.760cm,0.918cm);
\coordinate (v8) at (2.730cm,2.022cm);
\coordinate (v9) at (4.949cm,5.490cm);
\coordinate (v10) at (5.038cm,2.468cm);
\coordinate (v11) at (2.185cm,3.937cm);
\coordinate (v12) at (4.490cm,0.343cm);
\coordinate (v13) at (3.854cm,3.903cm);
\coordinate (v14) at (3.102cm,0.351cm);
\coordinate (v15) at (2.415cm,5.829cm);
\draw[-{Stealth[length=4mm]},line width=0.45pt,opacity=0.7,edge1] (v1) -- (v0);
\draw[-{Stealth[length=4mm]},line width=0.45pt,opacity=0.7,edge4] (v1) -- (v5);
\draw[-{Stealth[length=4mm]},line width=0.45pt,opacity=0.7,edge0] (v1) -- (v4);
\draw[-{Stealth[length=4mm]},line width=0.45pt,opacity=0.7,edge1] (v2) -- (v0);
\draw[-{Stealth[length=4mm]},line width=0.45pt,opacity=0.7,edge1] (v2) -- (v4);
\draw[-{Stealth[length=4mm]},line width=0.45pt,opacity=0.7,edge0] (v2) -- (v1);
\draw[-{Stealth[length=4mm]},line width=0.45pt,opacity=0.7,edge1] (v3) -- (v0);
\draw[-{Stealth[length=4mm]},line width=0.45pt,opacity=0.7,edge0] (v3) -- (v4);
\draw[-{Stealth[length=4mm]},line width=0.45pt,opacity=0.7,edge3] (v3) -- (v13);
\draw[-{Stealth[length=4mm]},line width=0.45pt,opacity=0.7,edge3] (v3) -- (v9);
\draw[-{Stealth[length=4mm]},line width=0.45pt,opacity=0.7,edge0] (v4) -- (v0);
\draw[-{Stealth[length=4mm]},line width=0.45pt,opacity=0.7,edge4] (v4) -- (v5);
\draw[-{Stealth[length=4mm]},line width=0.45pt,opacity=0.7,edge3] (v4) -- (v13);
\draw[-{Stealth[length=4mm]},line width=0.45pt,opacity=0.7,edge3] (v4) -- (v9);
\draw[-{Stealth[length=4mm]},line width=0.45pt,opacity=0.7,edge0] (v5) -- (v0);
\draw[-{Stealth[length=4mm]},line width=0.45pt,opacity=0.7,edge0] (v5) -- (v6);
\draw[-{Stealth[length=4mm]},line width=0.45pt,opacity=0.7,edge0] (v5) -- (v10);
\draw[-{Stealth[length=4mm]},line width=0.45pt,opacity=0.7,edge1] (v7) -- (v6);
\draw[-{Stealth[length=4mm]},line width=0.45pt,opacity=0.7,edge4] (v7) -- (v5);
\draw[-{Stealth[length=4mm]},line width=0.45pt,opacity=0.7,edge0] (v7) -- (v9);
\draw[-{Stealth[length=4mm]},line width=0.45pt,opacity=0.7,edge1] (v8) -- (v6);
\draw[-{Stealth[length=4mm]},line width=0.45pt,opacity=0.7,edge3] (v8) -- (v4);
\draw[-{Stealth[length=4mm]},line width=0.45pt,opacity=0.7,edge0] (v8) -- (v9);
\draw[-{Stealth[length=4mm]},line width=0.45pt,opacity=0.7,edge3] (v8) -- (v13);
\draw[-{Stealth[length=4mm]},line width=0.45pt,opacity=0.7,edge0] (v9) -- (v6);
\draw[-{Stealth[length=4mm]},line width=0.45pt,opacity=0.7,edge4] (v9) -- (v5);
\draw[-{Stealth[length=4mm]},line width=0.45pt,opacity=0.7,edge3] (v9) -- (v13);
\draw[-{Stealth[length=4mm]},line width=0.45pt,opacity=0.7,edge1] (v11) -- (v10);
\draw[-{Stealth[length=4mm]},line width=0.45pt,opacity=0.7,edge4] (v11) -- (v5);
\draw[-{Stealth[length=4mm]},line width=0.45pt,opacity=0.7,edge0] (v11) -- (v13);
\draw[-{Stealth[length=4mm]},line width=0.45pt,opacity=0.7,edge1] (v12) -- (v10);
\draw[-{Stealth[length=4mm]},line width=0.45pt,opacity=0.7,edge3] (v12) -- (v4);
\draw[-{Stealth[length=4mm]},line width=0.45pt,opacity=0.7,edge3] (v12) -- (v9);
\draw[-{Stealth[length=4mm]},line width=0.45pt,opacity=0.7,edge0] (v12) -- (v13);
\draw[-{Stealth[length=4mm]},line width=0.45pt,opacity=0.7,edge0] (v13) -- (v10);
\draw[-{Stealth[length=4mm]},line width=0.45pt,opacity=0.7,edge4] (v13) -- (v5);
\draw[-{Stealth[length=4mm]},line width=0.45pt,opacity=0.7,edge1] (v14) -- (v5);
\draw[-{Stealth[length=4mm]},line width=0.45pt,opacity=0.7,edge1] (v14) -- (v11);
\draw[-{Stealth[length=4mm]},line width=0.45pt,opacity=0.7,edge1] (v14) -- (v7);
\draw[-{Stealth[length=4mm]},line width=0.45pt,opacity=0.7,edge1] (v14) -- (v1);
\draw[-{Stealth[length=4mm]},line width=0.45pt,opacity=0.7,edge0] (v15) -- (v4);
\draw[-{Stealth[length=4mm]},line width=0.45pt,opacity=0.7,edge0] (v15) -- (v9);
\draw[-{Stealth[length=4mm]},line width=0.45pt,opacity=0.7,edge0] (v15) -- (v13);
\filldraw[fill=blue!12,draw=black!70,line width=0.4pt] (3.823cm,5.976cm) circle (1.4pt);
\node[landmark label, align=left] (label0) at (4.294cm,6.057cm) {\shortstack{at(p2, c1)}};
\draw[gray!65,line width=0.35pt] (v0) -- (label0);
\filldraw[fill=blue!12,draw=black!70,line width=0.4pt] (1.699cm,1.093cm) circle (1.4pt);
\node[landmark label, align=left] (label1) at (2.148cm,1.120cm) {\shortstack{at(t1, c8)}};
\draw[gray!65,line width=0.35pt] (v1) -- (label1);
\filldraw[fill=yellow!55,draw=black!70,line width=0.4pt] (6.878cm,3.871cm) circle (1.4pt);
\node[landmark label, align=left] (label2) at (7.157cm,3.299cm) {\shortstack{OR\\ at(t1, c5)\\ at(t1, c7)\\ at(t1, c9)}};
\draw[gray!65,line width=0.35pt] (v2) -- (label2);
\filldraw[fill=blue!12,draw=black!70,line width=0.4pt] (1.043cm,4.545cm) circle (1.4pt);
\node[landmark label, align=left] (label3) at (1.514cm,4.626cm) {\shortstack{at(p2, c8)}};
\draw[gray!65,line width=0.35pt] (v3) -- (label3);
\filldraw[fill=blue!12,draw=black!70,line width=0.4pt] (1.094cm,2.694cm) circle (1.4pt);
\node[landmark label, align=left] (label4) at (1.848cm,2.721cm) {\shortstack{carrying(t1, p2)}};
\draw[gray!65,line width=0.35pt] (v4) -- (label4);
\filldraw[fill=blue!12,draw=black!70,line width=0.4pt] (5.768cm,4.587cm) circle (1.4pt);
\node[landmark label, align=left] (label5) at (6.211cm,4.657cm) {\shortstack{at(t1, c1)}};
\draw[gray!65,line width=0.35pt] (v5) -- (label5);
\filldraw[fill=blue!12,draw=black!70,line width=0.4pt] (6.594cm,2.339cm) circle (1.4pt);
\node[landmark label, align=left] (label6) at (7.064cm,2.371cm) {\shortstack{at(p3, c1)}};
\draw[gray!65,line width=0.35pt] (v6) -- (label6);
\filldraw[fill=blue!12,draw=black!70,line width=0.4pt] (5.760cm,0.918cm) circle (1.4pt);
\node[landmark label, align=left] (label7) at (6.207cm,0.950cm) {\shortstack{at(t1, c3)}};
\draw[gray!65,line width=0.35pt] (v7) -- (label7);
\filldraw[fill=blue!12,draw=black!70,line width=0.4pt] (2.730cm,2.022cm) circle (1.4pt);
\node[landmark label, align=left] (label8) at (3.200cm,2.055cm) {\shortstack{at(p3, c3)}};
\draw[gray!65,line width=0.35pt] (v8) -- (label8);
\filldraw[fill=blue!12,draw=black!70,line width=0.4pt] (4.949cm,5.490cm) circle (1.4pt);
\node[landmark label, align=left] (label9) at (5.703cm,5.582cm) {\shortstack{carrying(t1, p3)}};
\draw[gray!65,line width=0.35pt] (v9) -- (label9);
\filldraw[fill=blue!12,draw=black!70,line width=0.4pt] (5.038cm,2.468cm) circle (1.4pt);
\node[landmark label, align=left] (label10) at (5.508cm,2.500cm) {\shortstack{at(p1, c1)}};
\draw[gray!65,line width=0.35pt] (v10) -- (label10);
\filldraw[fill=blue!12,draw=black!70,line width=0.4pt] (2.185cm,3.937cm) circle (1.4pt);
\node[landmark label, align=left] (label11) at (2.633cm,3.970cm) {\shortstack{at(t1, c6)}};
\draw[gray!65,line width=0.35pt] (v11) -- (label11);
\filldraw[fill=blue!12,draw=black!70,line width=0.4pt] (4.490cm,0.343cm) circle (1.4pt);
\node[landmark label, align=left] (label12) at (4.960cm,0.370cm) {\shortstack{at(p1, c6)}};
\draw[gray!65,line width=0.35pt] (v12) -- (label12);
\filldraw[fill=blue!12,draw=black!70,line width=0.4pt] (3.854cm,3.903cm) circle (1.4pt);
\node[landmark label, align=left] (label13) at (4.608cm,3.936cm) {\shortstack{carrying(t1, p1)}};
\draw[gray!65,line width=0.35pt] (v13) -- (label13);
\filldraw[fill=blue!12,draw=black!70,line width=0.4pt] (3.102cm,0.351cm) circle (1.4pt);
\node[landmark label, align=left] (label14) at (3.550cm,0.424cm) {\shortstack{at(t1, c2)}};
\draw[gray!65,line width=0.35pt] (v14) -- (label14);
\filldraw[fill=blue!12,draw=black!70,line width=0.4pt] (2.415cm,5.829cm) circle (1.4pt);
\node[landmark label, align=left] (label15) at (2.897cm,5.856cm) {\shortstack{empty(t1)}};
\draw[gray!65,line width=0.35pt] (v15) -- (label15);
\node[anchor=west,font=\footnotesize\bfseries] at (8.75cm,4.58cm) {Edge type};
\draw[edge0,line width=1pt] (8.75cm,4.17cm) -- (9.17cm,4.17cm);
\node[anchor=west,font=\footnotesize] at (9.30cm,4.17cm) {greedy necessary};
\draw[edge1,line width=1pt] (8.75cm,3.69cm) -- (9.17cm,3.69cm);
\node[anchor=west,font=\footnotesize] at (9.30cm,3.69cm) {natural};
\draw[edge2,line width=1pt] (8.75cm,3.21cm) -- (9.17cm,3.21cm);
\node[anchor=west,font=\footnotesize] at (9.30cm,3.21cm) {necessary};
\draw[edge3,line width=1pt] (8.75cm,2.73cm) -- (9.17cm,2.73cm);
\node[anchor=west,font=\footnotesize] at (9.30cm,2.73cm) {obedient-reasonable};
\draw[edge4,line width=1pt] (8.75cm,2.25cm) -- (9.17cm,2.25cm);
\node[anchor=west,font=\footnotesize] at (9.30cm,2.25cm) {reasonable};
\end{tikzpicture}
    \caption{LAMA's extracted traditional landmark graph for the \Delivery{} instance from \autoref{fig:lm:delex}.}
    \label{fig:lama:del}
\end{figure}

First, we compare these generalized landmarks to traditional landmarks and consider the difference in the information they convey to qualitatively answer~\ref{q:lm:trad:info}. 
We used the LAMA planner~\cite{Richter2010} to extract the traditional landmarks.
\autoref{fig:lama:del} shows the complete traditional landmark graph created by LAMA for the \Delivery{} instance from~\autoref{fig:lm:delex}, for more information on the edge types we refer to~\textcite{Richter2010}. 
The traditional landmark graph in~\autoref{fig:lama:del} has more nodes than the four nodes in both~$\LandmarkGraphExample$ and~$\LandmarkGraphDisc$.
This makes it harder to read, while the flow of generalized landmarks is much easier to see.
On the other hand, the nodes' information is very clear compared to the state functions as discussed in~\Cref{sec:lm:funcs}. 
The goal of the generalized landmarks is also quite clear: the last landmark provides an indication, and together with the loop condition, we see that when the empty truck is at the goal and no packages can be delivered, we are done.
In contrast, the traditional goal landmarks are the nodes that only have incoming edges~(e.g., $at(\PackageVar_2,\CellVar_1)$) that are hard to spot among all the landmarks, as the structure is less clear. Moreover, the dependencies between traditional landmarks are more complex to see than the generalized ones.

\begin{sloppypar}
To evaluate the difference in information disclosed by both landmark graphs, we look for the information of generalized landmarks in LAMA's traditional landmark graph. For generalized landmark~$\GeneralizedLandmark_1$, the empty truck being at a cell cannot be found directly in~\autoref{fig:lama:del}; we have the separate nodes~$empty(\TruckVar_1)$, $at(\PackageVar_1, \CellVar_6)$, and~$at(\TruckVar_1, \CellVar_6)$, which are all ordered before node~$carrying(\TruckVar_1, \PackageVar_1)$. So, although this information is technically included, it is quite hard to find. Next, generalized landmark~$\GeneralizedLandmark_2$ says a truck is carrying a package, but the truck is not at a cell with packages. We see the nodes~$carrying(\TruckVar_1, \PackageVar_1)$, $carrying(\TruckVar_1,\PackageVar_2)$ and~$carrying(\TruckVar_1, \PackageVar_3)$, which are all ordered before~$at(\TruckVar_1,\CellVar_1)$ (the goal cell), but the information that the package is not at the cell is not present. Generalized landmark~$\GeneralizedLandmark_3$, truck at the goal cell of the package being carried, is found in the order~$carrying(\TruckVar_1, \PackageVar_1)$ before~$at(\TruckVar_1,\CellVar_1)$, although the fact that this cell is indeed the goal is unknown. The generalized landmark~$\GeneralizedLandmark_4$ says that an empty truck is at the goal location, which is not a traditional landmark produced by LAMA. 
\end{sloppypar}

With this analysis, we answer~\ref{q:lm:trad:info} and conclude that the graph of generalized landmarks uncovers more information than traditional landmarks. Generalized landmarks identified higher-level relations that can be used by the planner; for example, the truck is at the goal cell of the package being carried instead of some specific cell or carrying a different package, while traditional landmarks might have identified more relations, but they are simpler, such as all the different atom relations. 

The other method we compare with is the policy sketches~\cite{Drexler2024}. 
To do so, we constructed a generalized landmark graph (\autoref{fig:sketchLm}) based on the policy rules used to define a policy sketch~\cite{Bonet2021}, which are given in~\autoref{eq:sk}. This generalized landmark graph uses the same type of state functions as generalized landmarks (the DLplan features), and we refer to this graph as~$\LandmarkGraphSketch$. The state descriptors used here are~$\FeatureH$, which is \texttt{true} when a truck is holding a package;~$\FeatureP$ gives the distance to the nearest undelivered package;~$\FeatureT$ gives the distance to the nearest goal cell; and~$\FeatureN$ gives the number of undelivered packages.

\begin{subequations}
\begin{align}
    \{\neg \FeatureH,\FeatureP>0\} &\to \{\FeatureP\downarrow,\FeatureT?\} \label{eq:sk1}\\ 
    \{\neg \FeatureH,\FeatureP=0\}  &\to \{\FeatureH\} \label{eq:sk2}\\
    \{\FeatureH,\FeatureT>0\} &\to \{\FeatureT\downarrow\} \label{eq:sk3}\\
    \{\FeatureH,\FeatureN>0,\FeatureT=0\} &\to \{\neg \FeatureH,\FeatureN\downarrow,\FeatureP?\} \label{eq:sk4}
\end{align}
\label{eq:sk}
\end{subequations}
\makebox[\linewidth]{\autoref{eq:sk}: Policy sketch for the \Delivery{} domain.}

\begin{figure}[h]
    \centering
    \begin{tikzpicture}[node distance=.5cm]
        \node[draw] (lm0) {$\neg \FeatureH \land \FeatureP > 0$};
        \node[draw, below=of lm0] (lm1) {$\neg \FeatureH \land \FeatureP=0$};
        \node[draw, below=of lm1] (lm2) {$\FeatureH \land \FeatureT>0$};
        \node[draw, below=of lm2] (lm3) {$\FeatureH \land \FeatureT=0 \land \FeatureN > 0$};
        \node[draw, below=of lm3] (lm4) {$\FeatureN=0$};
        \draw[->] (lm0) -- (lm1);
        \draw[->] (lm1) -- (lm2);
        \draw[->] (lm2) -- (lm3);
        \draw[->] (lm3) -- (lm4);
        \path[->] (lm3.east) edge[bend right] node[right] {$\FeatureN \downarrow$} (lm0.east);
    \end{tikzpicture}
    \caption{Generalized landmark graph for the policy rules in~\autoref{eq:sk}.}
    \label{fig:sketchLm}
\end{figure}

To determine whether generalized landmarks reveal different information than policy sketches~(\ref{q:lm:sk:info}), we qualitatively compare this graph and the corresponding sketch (\autoref{eq:sk}) to the information in our generated graph. 
Each rule in the policy sketch can be applied many times, effectively creating small loops of similar states. For example, the first rule (\autoref{eq:sk1}) can be executed until a truck reaches an undelivered package. This is similar to achieving the first generalized landmark, except that no information is given before reaching a generalized landmark that the truck is going in the right direction. The first generalized landmark~$\GeneralizedLandmark_1$ is equivalent to the left-hand side of~\autoref{eq:sk2}, and~$\GeneralizedLandmark_2$ reveals similar information as the right-hand side. However,~$\GeneralizedLandmark_2$ has more information; the truck should be at a cell with no packages, while this may be a limitation for an instance with multiple packages starting at the same cell. Generalized landmarks capture the notion of being at the target cell of the package being carried, which is not in the primary definition of the policy sketch but could be included as well~\cite{Bonet2021}. Our graph of generalized landmarks consists of a loop from generalized landmark~$\GeneralizedLandmark_4$ to generalized landmark~$\GeneralizedLandmark_3$, which is something that we cannot directly see in the policy sketch, while the policy sketch has small `loops' within each rule. Therefore, the difference between policy sketches and graphs of generalized landmarks is the loop behavior, which is higher-level for generalized landmarks and lower-level for the policy sketches, answering~\ref{q:lm:sk:info}. 

Finally, we compare the performance in terms of the number of expanded states for the~$\LMHeur$ heuristic on the \Delivery{}generalized landmark graphs from this paper.
We compare the hand-constructed graph~$\LandmarkGraphExample$, the graph learned from hand-constructed instances solved with Fast Downward (FD)~$\LandmarkGraphExperiment{\TrainingTrajectories}_\mathrm{FD}$, the graph learned from hand-constructed plans~$\LandmarkGraphExperiment{\TrainingTrajectories}$, the graph~$\LandmarkGraphDisc$ discovered from five training instances using state function complexity 11, and the graph derived from the policy sketched~$\LandmarkGraphSketch$.
Initial testing resulted in many out-of-memory failures for~$\LMHeur$ with all graphs except for~$\LandmarkGraphExample$, so we allocated~16GB of memory for each heuristic configuration for solving a single instance.
The results are given in~\autoref{tab:lm:delivery}.
The hand-constructed graph~$\LandmarkGraphExample$ is the only one solving all instances, in much fewer states than~$\HAdd$.
The next-best-performing graph is the second-most tailored, as it is learned from hand-constructed plans~$\LandmarkGraphExperiment{\TrainingTrajectories}$. 
The fully learned graph~$\LandmarkGraphDisc$ compared very similarly in terms of number of solved instances, though it required many more state expansions and thus resulted in more memory failures. 
The only graph that failed to find a plan was trained on hand-constructed instances with FD plans. 
Finally, the graph composed from the policy sketches did not perform very well, resulting mostly in timeouts.
So, generalized landmarks solve more instances than policy sketches when used in the landmark heuristic~$\LMHeur$ (\ref{q:lm:sk:comp}).
However, this is obviously not a fair comparison. 
As the planning results using policy sketches reported in~\textcite{Drexler2024} included the domains Barman, Childsnack, Driverlog, Floortile, Grid, Schedule, and TPP, and generalized landmarks were not found with a loop for these domains, we cannot make a direct comparison.

\begin{table}[ht]
	\centering
	\caption{Solved \Delivery{} instances and expanded states for~$\LMHeur$ and~$\LMHeur_{\Guide}$ on each generalized landmark graph. Expanded states are averaged over instances solved by both the landmark heuristic and $\HAdd$; PI is the paired-instance count.} 
	\label{tab:lm:delivery}
	\begin{tabular}{llrrrrrrr}
		\toprule
		&  & Solved & Failure & T/O & O/M & PI & Expanded & $\HAdd$ exp. \\
		$\LandmarkGraph$ & $\LMHeur$ &  &  &  &  &  &  &  \\
		\midrule
		\multirow[t]{2}{*}{$\LandmarkGraphExample$} & $\LMHeur$ & 30 & 0 & 0 & 0 & 29 & 190 & 17723 \\
		& $\LMHeur_{\Guide}$ & 30 & 0 & 0 & 0 & 29 & 136 & 17723 \\
		\cmidrule{1-9}
		\multirow[t]{2}{*}{$\LandmarkGraphExperiment{\TrainingTrajectories}_{\mathrm{FD}}$} & $\LMHeur$ & 7 & 5 & 10 & 8 & 7 & 9696 & 1827 \\
		& $\LMHeur_{\Guide}$ & 6 & 8 & 12 & 4 & 6 & 10874 & 167 \\
		\cmidrule{1-9}
		\multirow[t]{2}{*}{$\LandmarkGraphExperiment{\TrainingTrajectories}$} & $\LMHeur$ & 10 & 0 & 14 & 6 & 10 & 3036 & 1581 \\
		& $\LMHeur_{\Guide}$ & 9 & 0 & 15 & 6 & 9 & 1210 & 1723 \\
		\cmidrule{1-9}
		\multirow[t]{2}{*}{$\LandmarkGraphDisc$} & $\LMHeur$ & 9 & 0 & 9 & 12 & 9 & 163691 & 162 \\
		& $\LMHeur_{\Guide}$ & 10 & 0 & 9 & 11 & 10 & 149674 & 218 \\
		\cmidrule{1-9}
		\multirow[t]{2}{*}{$\LandmarkGraphSketch$} & $\LMHeur$ & 6 & 0 & 24 & 0 & 6 & 138188 & 118 \\
		& $\LMHeur_{\Guide}$ & 6 & 0 & 23 & 1 & 6 & 145906 & 118 \\
		\bottomrule
	\end{tabular}
\end{table}

\section{Conclusion}
This chapter introduces a novel approach for reusing previously successful plans to learn abstractions that help automated planners by reducing the number of states expanded during search.
We revisit the definition of traditional landmarks and introduce generalized landmarks that apply to all problems in a domain. 
The domain comprises the predicates and actions, as well as a common goal representation shared among the problem instances.
Generalized landmarks no longer depend on the atoms of a problem instance but are defined by state functions that capture symmetries and are independent of a problem's specific objects.
Moreover, generalized landmarks allow grouping of repetitive behavior to achieve the goal by including loops in the generalized landmark graph, such as repeating actions for different objects.

Using domain knowledge, we construct graphs of generalized landmarks out of first-order functions over a state.
We show that generalized landmarks provide a compact representation of the subgoals in a domain, thereby yielding an interpretable plan.
When the graph is well-defined, our proposed heuristic significantly outperforms the baseline by leveraging loop information. 
In many real-world applications, human expertise is available to construct these graphs, and the sequential nature of the problem is well-suited for the use of generalized landmark graphs. This again shows the potential of manually derived generalized landmark graphs.

Generalized landmarks share commonalities with policy sketches, as they are based on the same type of state functions, and both methods derive an abstract plan. 
The main difference is that generalized landmarks describe \emph{states} while policies describe \emph{state transitions}. 
Because generalized landmarks are derived from known plans, they can capture how to solve an instance, rather than the instance itself or the general search space. 
Finally, generalized landmarks can be used in a simple heuristic that can be implemented across many planners, whereas policy sketches are applied in specialized iterated width planners.

\subsection{Future Work}
In future work, we want to improve the learning method to extract a generalized landmark graph from a few small solved instances across all domains.
Currently, we generate a set of state functions and identify the same abstract states by finding a subset of state functions that describe that state across instances and objects.
While many different landmark graphs can be described for one domain, one is more useful in heuristic search than another, based on the descriptiveness of the state functions and the presence of one or more loops.
The remaining challenge is how to optimize the learning process to automatically find a representative and efficient landmark graph while interleaving loop discovery.

A first direction for improvement is to allow disjunctive paths in the generalized landmark graph.
The presented framework must form a chain, which is not sufficient to describe all domains, as some domains have simultaneous subgoals that must be achieved. 
For example, in the \texttt{Logistics} domain, packages are transported by trucks within cities, and by planes between cities.
However, the state trajectories from which generalized landmarks are discovered may first transport a package~$\PackageVar_1$ with a truck in city~$ct_2$, then fly a plane carrying package~$\PackageVar_2$ from city~$ct_3$ to city~$ct_4$, and then fly a plane carrying package~$\PackageVar_1$ from city~$ct_2$ to city~$ct_3$.
We want to extend the framework to support these parallel subgoals and their disjunctive paths, so the heuristic can use them and the learning algorithm can handle these training instances.
This would also allow planning with several agents, where disjunctive paths enable simultaneous execution.
Another extension is self-loops in the generalized landmark graph to capture specific patterns. 
While we do allow some loops to be traversed zero times in specific instances, a self-loop would make the framework more intuitive.
For example, in the \texttt{visitall} domain, each state ideally achieves the next landmark, which is most easily captured with a self-loop.

Second, when all generalized landmarks are positive, a similar method can be used to find generalized \emph{negative} landmarks, such as avoid states, which is especially interesting for domains with dead ends. 
Although the \texttt{Spanner} domain showed that we can already solve domains with dead ends to some extent, it would be interesting to study this in more depth.
Especially in more complicated real-world domains where resources are scarce, this might be useful.

Third, the related work section already noted some closely related topics, such as Hierarchical Task Networks (HTN) and abstract skills, where generalized landmarks may be useful for decomposing HTN methods and automatically learning abstract skills that cover more than just domain predicates. 
This highlights the broader potential of generalized landmarks, extending beyond classical planning and connecting to generalized planning, hierarchical planning, and use cases in robotics. 

We see a fourth potential for generalized landmarks to be applied in planning settings to optimize replanning.
Generalized landmarks can be used to indicate the states where earlier parts of a plan can be reused. 
Then, only this part has to be replanned, reducing computation efforts. 
In approaches where recovering from insufficient or low-quality could investigate the required quality of the training trajectories to maintain reasonable landmark qualityan investigate the required quality of the training trajectories to still provide reasonable landmarks.

Finally, in human-AI collaboration settings, the abstract plan composed by generalized landmarks is suitable for more interpretable representations of complex plans, especially in real-world domains. 
Learning the graphs from previous human-made plans could reveal the patterns they may not consciously adhere to. 
Or the generalized landmarks could serve as an explanation of the automated plan produced that should be adopted by human operators, which is much easier to interpret than a plan of several hundred actions.

\section*{Acknowledgements}
    This work is part of the NWO LTP-ROBUST RAIL Lab, a collaboration between the Delft University of Technology, Utrecht University, NS, and ProRail. More information at \url{https://icai.ai/icai-labs/rail/}.

\begin{sloppypar}
    \printbibliography
\end{sloppypar}

\clearpage
\appendix

\section{Domain descriptions}
\label{app:dom}
We use a selection of PDDL domains, some of which were taken from the PDDLgym library~\cite{Silver2020}, some of which were used in the DLplan library~\cite{Drexler2022b}, and some of which were used in the SketchLearner evaluation~\cite{Drexler2024}, some originating from the autoscale benchmark set~\cite{Torralba2021}. For some of the domains, we created an instance generator to control the size of the instances. All the domains and instances used can be found in our repository~\cite{LandmarkCode2025}, and the generators are in the folder of the associated domain.

List of domains:
\begin{itemize}
    \item \texttt{Baking}: this domain comes from the PDDLgym library, and we created a generator. The goal is to bake a set of cakes (requiring egg and flour) and a set of souffles (requiring only egg). For each baked good, the ingredients must be mixed and baked in the oven. For each good, a clean pan must be available, either from the initial state or cleaned after use, for which soap must be available. We changed the domain to have an \texttt{ovenisempty} predicate and used \texttt{ovenisempty(Oven)} in the goal instead of \texttt{not(ovenisfull(Oven))} to avoid negative goals, which are difficult for the DLplan library.
    \item \texttt{Barman}: this domain comes from the autoscale set, which we altered not to have cost-optimization. The goal is to make a set of cocktails with two ingredients each. To make a cocktail, an ingredient has to be poured from a dispenser into a clean shot glass, then added to the shaker. A shaker with two ingredients can be shaken and poured into a shot glass to fulfill the goal of making one cocktail.
    \item \texttt{Blocksworld}: this domain comes from the PDDLgym library. The goal is to create a particular set of stacked blocks on one table by unstacking and stacking blocks from their initial stacks on the table.
    \item \texttt{Childsnack}: this domain comes from the autoscale set. The goal is to serve sandwiches to children seated at different tables. Some children are allergic to gluten, so they must be served a gluten-free sandwich. To reach the goal, we must make (gluten-free) sandwiches, put them on the tray, go to the tables, and serve the sandwiches.
    \item \texttt{Delivery}: this domain comes from the DLplan library benchmark set. The goal is to deliver packages to the goal locations. Generally, one truck can pick up packages and drop them at a different location in a grid-based world.
    \item \texttt{Driverlog}: this domain comes from the autoscale benchmark set. Packages must be delivered by loading them into a truck, driving the truck to the target location, and unloading the packages. A driver is required to drive a truck, and drivers and trucks may also have goal locations.
    \item \texttt{Ferry}: this domain comes from the PDDLgym library, and we created a generator. Given a set of locations and passengers, a ferry boat has to transport the passengers from their origin to their destination locations.
    \item \texttt{Floortile}: this domain comes from the autoscale benchmark set. A set of robots has to paint a set of tiles in a grid; specifically, a rectangular portion of a rectangular grid has to be painted. Robots can only paint one cell up or down from their position, and each robot is equipped with a color, although they can change colors.
    \item \texttt{Footwear}: this domain comes from the PDDLgym library. There are four possible locations to travel to perform a task: the gym (workout), the beach (swim), the office (give a presentation), and the forest (hike). To perform the associated tasks, the agent has to wear the correct footwear: a sneaker, a sandal, a dress shoe, or a boot, respectively. The agent can only change socks and shoes at home, so the goal is to wear the correct footwear and perform the required tasks.
    \item \texttt{Grid}: this domain comes from the autoscale benchmark set. The goal is to move keys to specific locations, but some may be locked, so a key is needed to unlock the location to reach the target location. Keys and locks have different shapes that must match to open a lock. The robot can only hold one key at a time.
    \item \texttt{Grid-no-exchange}: this domain is an adaptation of the \texttt{Grid} domain used in the SketchLearner benchmark set, where a key being held cannot be exchanged with a key at the current location.
    \item \texttt{Logistics}: this domain comes from the PDDLgym library, and we created a generator. Packages have to be transported between cities. Between cities, planes can fly from one airport to another, and within a city, trucks can transport packages between locations.
    \item \texttt{Miconic}: this domain comes from the DLplan benchmark set. The goal is to serve passengers waiting on an elevator, where each has an origin and a destination floor. 
    \item \texttt{Newspapers}: this domain comes from the PDDLgym library, and we created a generator. Given a set of locations that want a paper to be delivered and a set of papers that can be delivered, the goal is to serve all locations waiting for a paper.
    \item \texttt{Satellite}: this domain comes from the IPC learning track. The goal is to take images with satellites in space. To do so, the satellites must be turned in the right direction, and the right instruments on board must be turned on and calibrated.
    \item \texttt{Spanner}: this domain comes from the IPC learning track. The goal is to tighten a set of bolts at the gate, while picking up spanners throughout the yard. The paths are one-way and spanners can be used only once, so if too few are picked up, the instance cannot be solved.
    \item \texttt{TPP}: this domain comes from the autoscale benchmark set. The goal of the Traveling Purchaser Problem is to store specific goods, which can be done at depots connected with roads to the markets where goods can be bought. The quantities of available and requested goods have to be met as well.
    \item \texttt{Transport}: this domain comes from the IPC learning track. The domain is very similar to \Delivery{}, but trucks can have different capacities, which are objects as the domain is non-numerical.
    \item \texttt{Visitall}: this domain comes from the DPplan benchmark set. Given a set of locations that are connected in a predefined manner (usually a grid), the goal is to visit each location. 
\end{itemize}

\section{Full Results for Discovery of Generalized Landmark Graphs}
\label{app:lm:disc-full}

\begin{longtable}{lllrrrrr}
\caption{Complete results for the discovery experiment from~\Cref{sec:lm:learn:exp:res} shown in~\autoref{fig:lm:disc}. Comp shows the allowed state function complexity.} \label{tab:lm:disc-full} \\
\toprule
 & & & $|\FeaturePool{}|$ & Time & $|\nodes|$ & $|\ConditionalOrderings|$ & $|\GuideSet|$ \\
$\Domain$ & $\Preprocessing$ & Comp. & & & & \\
\midrule
\endfirsthead
\caption[]{Complete results for the discovery experiment from~\Cref{sec:lm:learn:exp:res} shown in~\autoref{fig:lm:disc}. Comp shows the allowed state function complexity.} \\
\toprule
 & & & $|\FeaturePool{}|$ & Time & $|\nodes|$ & $|\ConditionalOrderings|$ & $|\GuideSet|$ \\
$\Domain$ & $\Preprocessing$ & Comp. & & & & \\
\midrule
\endhead
\midrule
\multicolumn{8}{r}{Continued on next page} \\
\midrule
\endfoot
\bottomrule
\endlastfoot
\multirow[t]{20}{*}{baking} & \multirow[t]{5}{*}{$\Preprocessing_1$} & 7 & 2273 & 1169.15 & 7 & 1 & 5 \\
 &  & 9 & 2273 & 1143.46 & 7 & 1 & 5 \\
 &  & 11 & 2273 & 1141.07 & 7 & 1 & 5 \\
 &  & 13 & 2273 & 1173.40 & 7 & 1 & 5 \\
 &  & 15 & 2273 & 1168.62 & 7 & 1 & 5 \\
\cmidrule{2-8}
 & \multirow[t]{5}{*}{$\Preprocessing_2$} & 7 & 2273 & 1085.35 & 7 & 0 & 5 \\
 &  & 9 & 2273 & 1082.84 & 7 & 0 & 5 \\
 &  & 11 & 2273 & 1066.54 & 7 & 0 & 5 \\
 &  & 13 & 2273 & 1065.55 & 7 & 0 & 5 \\
 &  & 15 & 2273 & 1097.31 & 7 & 0 & 5 \\
\cmidrule{2-8}
 & \multirow[t]{5}{*}{$\Preprocessing_3$} & 7 & 2273 & 1073.28 & 7 & 0 & 5 \\
 &  & 9 & 2273 & 1081.64 & 7 & 0 & 5 \\
 &  & 11 & 2273 & 1071.70 & 7 & 0 & 5 \\
 &  & 13 & 2273 & 1097.48 & 7 & 0 & 5 \\
 &  & 15 & 2273 & 1080.04 & 7 & 0 & 5 \\
\cmidrule{2-8}
 & \multirow[t]{5}{*}{$\Preprocessing_4$} & 7 & 2273 & 1070.13 & 7 & 0 & 5 \\
 &  & 9 & 2273 & 1081.80 & 7 & 0 & 5 \\
 &  & 11 & 2273 & 1071.51 & 7 & 0 & 5 \\
 &  & 13 & 2273 & 1100.64 & 7 & 0 & 5 \\
 &  & 15 & 2273 & 1083.46 & 7 & 0 & 5 \\
\cmidrule{1-8} \cmidrule{2-8}
\multirow[t]{4}{*}{barman} & $\Preprocessing_1$ & 7 & 1485 & 2088.04 & 5 & 0 & 2 \\
\cmidrule{2-8}
 & $\Preprocessing_2$ & 7 & 1485 & 2168.11 & 5 & 0 & 0 \\
\cmidrule{2-8}
 & $\Preprocessing_3$ & 7 & 1485 & 2134.57 & 5 & 0 & 0 \\
\cmidrule{2-8}
 & $\Preprocessing_4$ & 7 & 1485 & 2110.36 & 5 & 0 & 0 \\
\cmidrule{1-8} \cmidrule{2-8}
\multirow[t]{20}{*}{blocksworld} & \multirow[t]{5}{*}{$\Preprocessing_1$} & 7 & 941 & 12.84 & 0 & 0 & 0 \\
 &  & 9 & 1623.25 & 14.86 & 0 & 0 & 0 \\
 &  & 11 & 1623.25 & 15.12 & 0 & 0 & 0 \\
 &  & 13 & 1623.25 & 15.40 & 0 & 0 & 0 \\
 &  & 15 & 1623.25 & 14.88 & 0 & 0 & 0 \\
\cmidrule{2-8}
 & \multirow[t]{5}{*}{$\Preprocessing_2$} & 7 & 941 & 12.93 & 0 & 0 & 0 \\
 &  & 9 & 1623.25 & 14.99 & 0 & 0 & 0 \\
 &  & 11 & 1623.25 & 15.33 & 0 & 0 & 0 \\
 &  & 13 & 1623.25 & 15.51 & 0 & 0 & 0 \\
 &  & 15 & 1623.25 & 14.89 & 0 & 0 & 0 \\
\cmidrule{2-8}
 & \multirow[t]{5}{*}{$\Preprocessing_3$} & 7 & 941 & 13.07 & 0 & 0 & 0 \\
 &  & 9 & 1623.25 & 15.01 & 0 & 0 & 0 \\
 &  & 11 & 1623.25 & 15.61 & 0 & 0 & 0 \\
 &  & 13 & 1623.25 & 15.51 & 0 & 0 & 0 \\
 &  & 15 & 1623.25 & 15.41 & 0 & 0 & 0 \\
\cmidrule{2-8}
 & \multirow[t]{5}{*}{$\Preprocessing_4$} & 7 & 941 & 12.71 & 0 & 0 & 0 \\
 &  & 9 & 1623.25 & 15.10 & 0 & 0 & 0 \\
 &  & 11 & 1623.25 & 15.41 & 0 & 0 & 0 \\
 &  & 13 & 1623.25 & 15.64 & 0 & 0 & 0 \\
 &  & 15 & 1623.25 & 14.84 & 0 & 0 & 0 \\
\cmidrule{1-8} \cmidrule{2-8}
\multirow[t]{20}{*}{childsnack} & \multirow[t]{5}{*}{$\Preprocessing_1$} & 7 & 347.75 & 50.90 & 2.50 & 1.50 & 1.50 \\
 &  & 9 & 1150.75 & 409.12 & 3.25 & 1.50 & 2.25 \\
 &  & 11 & 1927.25 & 1038.56 & 3.50 & 0 & 2.50 \\
 &  & 13 & 1927.25 & 1025.08 & 3.50 & 0 & 2.50 \\
 &  & 15 & 1927.25 & 1029.22 & 3.50 & 0 & 2.50 \\
\cmidrule{2-8}
 & \multirow[t]{5}{*}{$\Preprocessing_2$} & 7 & 347.75 & 48.62 & 2.50 & 1.50 & 1.50 \\
 &  & 9 & 1150.75 & 397.20 & 3.25 & 1.50 & 2.25 \\
 &  & 11 & 1927.25 & 937.51 & 3.25 & 0.50 & 2.25 \\
 &  & 13 & 1927.25 & 936.79 & 3.25 & 0.50 & 2.25 \\
 &  & 15 & 1927.25 & 937.88 & 3.25 & 0.50 & 2.25 \\
\cmidrule{2-8}
 & \multirow[t]{5}{*}{$\Preprocessing_3$} & 7 & 347.75 & 48.31 & 2.50 & 1.50 & 1.50 \\
 &  & 9 & 1150.75 & 403.61 & 3.25 & 1.50 & 2.25 \\
 &  & 11 & 1927.25 & 940.81 & 3.25 & 0.50 & 2.25 \\
 &  & 13 & 1927.25 & 922.74 & 3.25 & 0.50 & 2.25 \\
 &  & 15 & 1927.25 & 943.67 & 3.25 & 0.50 & 2.25 \\
\cmidrule{2-8}
 & \multirow[t]{5}{*}{$\Preprocessing_4$} & 7 & 347.75 & 48.71 & 2.50 & 1.50 & 1.50 \\
 &  & 9 & 1150.75 & 405.67 & 3.25 & 1.50 & 2.25 \\
 &  & 11 & 1927.25 & 926.83 & 3.25 & 0.50 & 2.25 \\
 &  & 13 & 1927.25 & 933.32 & 3.25 & 0.50 & 2.25 \\
 &  & 15 & 1927.25 & 942.81 & 3.25 & 0.50 & 2.25 \\
\cmidrule{1-8} \cmidrule{2-8}
\multirow[t]{20}{*}{delivery} & \multirow[t]{5}{*}{$\Preprocessing_1$} & 7 & 70 & 3.11 & 2 & 1 & 1 \\
 &  & 9 & 261 & 25.36 & 4 & 1 & 4 \\
 &  & 11 & 856 & 144.83 & 5 & 2 & 5 \\
 &  & 13 & 1456 & 309.45 & 5 & 2 & 5 \\
 &  & 15 & 1456 & 308.01 & 5 & 2 & 5 \\
\cmidrule{2-8}
 & \multirow[t]{5}{*}{$\Preprocessing_2$} & 7 & 70 & 2.79 & 2 & 1 & 1 \\
 &  & 9 & 261 & 19.43 & 4 & 1 & 3 \\
 &  & 11 & 856 & 100.96 & 5 & 1 & 5 \\
 &  & 13 & 1456 & 239.25 & 5 & 2 & 5 \\
 &  & 15 & 1456 & 233.15 & 5 & 2 & 5 \\
\cmidrule{2-8}
 & \multirow[t]{5}{*}{$\Preprocessing_3$} & 7 & 70 & 1.63 & 1 & 0 & 0 \\
 &  & 9 & 261 & 18.66 & 4 & 1 & 2 \\
 &  & 11 & 856 & 113.96 & 5 & 0 & 4 \\
 &  & 13 & 1456 & 243.10 & 5 & 1 & 4 \\
 &  & 15 & 1456 & 237.19 & 5 & 1 & 4 \\
\cmidrule{2-8}
 & \multirow[t]{5}{*}{$\Preprocessing_4$} & 7 & 70 & 2.86 & 2 & 1 & 1 \\
 &  & 9 & 261 & 19.75 & 4 & 1 & 3 \\
 &  & 11 & 856 & 99.44 & 5 & 1 & 5 \\
 &  & 13 & 1456 & 239.89 & 5 & 2 & 5 \\
 &  & 15 & 1456 & 233.49 & 5 & 2 & 5 \\
\cmidrule{1-8} \cmidrule{2-8}
\multirow[t]{20}{*}{driverlog} & \multirow[t]{5}{*}{$\Preprocessing_1$} & 7 & 216 & 2076.51 & 3 & 0 & 0 \\
 &  & 9 & 1301 & 270.97 & 0 & 0 & 0 \\
 &  & 11 & 1301 & 267.38 & 0 & 0 & 0 \\
 &  & 13 & 1301 & 272.62 & 0 & 0 & 0 \\
 &  & 15 & 1301 & 278.33 & 0 & 0 & 0 \\
\cmidrule{2-8}
 & \multirow[t]{5}{*}{$\Preprocessing_2$} & 7 & 216 & 1867.86 & 3 & 0 & 0 \\
 &  & 9 & 1301 & 210.27 & 0 & 0 & 0 \\
 &  & 11 & 1301 & 213.53 & 0 & 0 & 0 \\
 &  & 13 & 1301 & 203.88 & 0 & 0 & 0 \\
 &  & 15 & 1301 & 209.98 & 0 & 0 & 0 \\
\cmidrule{2-8}
 & \multirow[t]{5}{*}{$\Preprocessing_3$} & 7 & 216 & 1901.93 & 3 & 0 & 0 \\
 &  & 9 & 1301 & 206.73 & 0 & 0 & 0 \\
 &  & 11 & 1301 & 213.07 & 0 & 0 & 0 \\
 &  & 13 & 1301 & 204.57 & 0 & 0 & 0 \\
 &  & 15 & 1301 & 216.55 & 0 & 0 & 0 \\
\cmidrule{2-8}
 & \multirow[t]{5}{*}{$\Preprocessing_4$} & 7 & 216 & 2003.56 & 3 & 0 & 0 \\
 &  & 9 & 1301 & 203.70 & 0 & 0 & 0 \\
 &  & 11 & 1301 & 210.04 & 0 & 0 & 0 \\
 &  & 13 & 1301 & 216.59 & 0 & 0 & 0 \\
 &  & 15 & 1301 & 213.98 & 0 & 0 & 0 \\
\cmidrule{1-8} \cmidrule{2-8}
\multirow[t]{20}{*}{ferry} & \multirow[t]{5}{*}{$\Preprocessing_1$} & 7 & 108 & 6.06 & 3 & 2 & 2.25 \\
 &  & 9 & 327.75 & 19.64 & 2.75 & 1.50 & 2 \\
 &  & 11 & 992 & 97.66 & 2.75 & 1.50 & 2 \\
 &  & 13 & 1786.75 & 295.17 & 2.75 & 1.50 & 2 \\
 &  & 15 & 1786.75 & 294.60 & 2.75 & 1.50 & 2 \\
\cmidrule{2-8}
 & \multirow[t]{5}{*}{$\Preprocessing_2$} & 7 & 108 & 5.49 & 3 & 2 & 1.75 \\
 &  & 9 & 327.75 & 16.76 & 3 & 2 & 1.75 \\
 &  & 11 & 992 & 83.42 & 2.75 & 1.50 & 1.75 \\
 &  & 13 & 1786.75 & 248.67 & 3 & 2 & 2 \\
 &  & 15 & 1786.75 & 248.19 & 3 & 2 & 2 \\
\cmidrule{2-8}
 & \multirow[t]{5}{*}{$\Preprocessing_3$} & 7 & 108 & 5.17 & 3 & 2 & 1.75 \\
 &  & 9 & 327.75 & 16.41 & 3 & 2 & 1.75 \\
 &  & 11 & 992 & 83.15 & 2.75 & 1.50 & 1.75 \\
 &  & 13 & 1786.75 & 238.96 & 3 & 2 & 2 \\
 &  & 15 & 1786.75 & 239.36 & 3 & 2 & 2 \\
\cmidrule{2-8}
 & \multirow[t]{5}{*}{$\Preprocessing_4$} & 7 & 108 & 5.52 & 3 & 2 & 1.75 \\
 &  & 9 & 327.75 & 17.82 & 3 & 2 & 1.75 \\
 &  & 11 & 992 & 83.64 & 2.75 & 1.50 & 1.75 \\
 &  & 13 & 1786.75 & 248.38 & 3 & 2 & 2 \\
 &  & 15 & 1786.75 & 247.72 & 3 & 2 & 2 \\
\cmidrule{1-8} \cmidrule{2-8}
\multirow[t]{20}{*}{floortile} & \multirow[t]{5}{*}{$\Preprocessing_1$} & 7 & 91.50 & 5.15 & 1 & 0 & 0.25 \\
 &  & 9 & 277 & 18.13 & 1 & 0 & 0.25 \\
 &  & 11 & 810.50 & 64.49 & 1 & 0 & 0.50 \\
 &  & 13 & 1509 & 144.10 & 1 & 0 & 0.50 \\
 &  & 15 & 1509 & 143.73 & 1 & 0 & 0.50 \\
\cmidrule{2-8}
 & \multirow[t]{5}{*}{$\Preprocessing_2$} & 7 & 91.50 & 4.70 & 1 & 0 & 0.25 \\
 &  & 9 & 277 & 16.04 & 1 & 0 & 0.25 \\
 &  & 11 & 810.50 & 56.82 & 1 & 0 & 0.25 \\
 &  & 13 & 1509 & 116.96 & 1 & 0 & 0.25 \\
 &  & 15 & 1509 & 117.66 & 1 & 0 & 0.25 \\
\cmidrule{2-8}
 & \multirow[t]{5}{*}{$\Preprocessing_3$} & 7 & 91.50 & 4.73 & 1 & 0 & 0.25 \\
 &  & 9 & 277 & 15.86 & 1 & 0 & 0.25 \\
 &  & 11 & 810.50 & 56.83 & 1 & 0 & 0.25 \\
 &  & 13 & 1509 & 116.34 & 1 & 0 & 0.25 \\
 &  & 15 & 1509 & 114.67 & 1 & 0 & 0.25 \\
\cmidrule{2-8}
 & \multirow[t]{5}{*}{$\Preprocessing_4$} & 7 & 91.50 & 4.69 & 1 & 0 & 0.25 \\
 &  & 9 & 277 & 15.44 & 1 & 0 & 0.25 \\
 &  & 11 & 810.50 & 55.99 & 1 & 0 & 0.25 \\
 &  & 13 & 1509 & 115.63 & 1 & 0 & 0.25 \\
 &  & 15 & 1509 & 114.67 & 1 & 0 & 0.25 \\
\cmidrule{1-8} \cmidrule{2-8}
\multirow[t]{20}{*}{footwear} & \multirow[t]{5}{*}{$\Preprocessing_1$} & 7 & 1601 & 72.78 & 2 & 0 & 0 \\
 &  & 9 & 7355 & 1392.82 & 2 & 0 & 0 \\
 &  & 11 & 7355 & 1302.73 & 2 & 0 & 0 \\
 &  & 13 & 7355 & 1397.96 & 2 & 0 & 0 \\
 &  & 15 & 7355 & 1251.95 & 2 & 0 & 0 \\
\cmidrule{2-8}
 & \multirow[t]{5}{*}{$\Preprocessing_2$} & 7 & 1511 & 63.97 & 2 & 0 & 0 \\
 &  & 9 & 6596 & 1102.06 & 2 & 0 & 0 \\
 &  & 11 & 7355 & 1169.97 & 2 & 0 & 0 \\
 &  & 13 & 7355 & 2799.74 & 2 & 0 & 0 \\
 &  & 15 & 7355 & 1371.59 & 2 & 0 & 0 \\
\cmidrule{2-8}
 & \multirow[t]{5}{*}{$\Preprocessing_3$} & 7 & 1601 & 70.58 & 2 & 0 & 0 \\
 &  & 9 & 7355 & 1400.67 & 2 & 0 & 0 \\
 &  & 11 & 7355 & 1317.09 & 2 & 0 & 0 \\
 &  & 13 & 7355 & 1333.94 & 2 & 0 & 0 \\
 &  & 15 & 7355 & 1354.46 & 2 & 0 & 0 \\
\cmidrule{2-8}
 & \multirow[t]{5}{*}{$\Preprocessing_4$} & 7 & 1601 & 69.43 & 2 & 0 & 0 \\
 &  & 9 & 7355 & 1166.39 & 2 & 0 & 0 \\
 &  & 11 & 7355 & 1350.94 & 2 & 0 & 0 \\
 &  & 13 & 7355 & 1361.91 & 2 & 0 & 0 \\
 &  & 15 & 7355 & 1322.57 & 2 & 0 & 0 \\
\cmidrule{1-8} \cmidrule{2-8}
\multirow[t]{20}{*}{grid} & \multirow[t]{5}{*}{$\Preprocessing_1$} & 7 & 348 & 5.13 & 1 & 0 & 1 \\
 &  & 9 & 1753 & 126.39 & 3 & 0 & 2 \\
 &  & 11 & 1753 & 126.08 & 3 & 0 & 2 \\
 &  & 13 & 1753 & 129.58 & 3 & 0 & 2 \\
 &  & 15 & 1753 & 127.67 & 3 & 0 & 2 \\
\cmidrule{2-8}
 & \multirow[t]{5}{*}{$\Preprocessing_2$} & 7 & 348 & 4.33 & 1 & 0 & 1 \\
 &  & 9 & 1753 & 92.85 & 3 & 0 & 2 \\
 &  & 11 & 1753 & 91.67 & 3 & 0 & 2 \\
 &  & 13 & 1753 & 93.05 & 3 & 0 & 2 \\
 &  & 15 & 1753 & 92.40 & 3 & 0 & 2 \\
\cmidrule{2-8}
 & \multirow[t]{5}{*}{$\Preprocessing_3$} & 7 & 348 & 4.18 & 1 & 0 & 1 \\
 &  & 9 & 1753 & 90.53 & 3 & 0 & 2 \\
 &  & 11 & 1753 & 89.60 & 3 & 0 & 2 \\
 &  & 13 & 1753 & 98.24 & 3 & 0 & 2 \\
 &  & 15 & 1753 & 91.64 & 3 & 0 & 2 \\
\cmidrule{2-8}
 & \multirow[t]{5}{*}{$\Preprocessing_4$} & 7 & 348 & 4.23 & 1 & 0 & 1 \\
 &  & 9 & 1753 & 91.48 & 3 & 0 & 2 \\
 &  & 11 & 1753 & 92.14 & 3 & 0 & 2 \\
 &  & 13 & 1753 & 99.07 & 3 & 0 & 2 \\
 &  & 15 & 1753 & 92.74 & 3 & 0 & 2 \\
\cmidrule{1-8} \cmidrule{2-8}
\multirow[t]{20}{*}{grid-no-exchange} & \multirow[t]{5}{*}{$\Preprocessing_1$} & 7 & 327 & 4.47 & 1 & 0 & 0 \\
 &  & 9 & 1344 & 149.26 & 4 & 0 & 3 \\
 &  & 11 & 1344 & 150.15 & 4 & 0 & 3 \\
 &  & 13 & 1344 & 146.75 & 4 & 0 & 3 \\
 &  & 15 & 1344 & 149.03 & 4 & 0 & 3 \\
\cmidrule{2-8}
 & \multirow[t]{5}{*}{$\Preprocessing_2$} & 7 & 288 & 7.27 & 2 & 0 & 1 \\
 &  & 9 & 1562 & 30.82 & 1 & 0 & 0 \\
 &  & 11 & 1344 & 106.72 & 4 & 0 & 2 \\
 &  & 13 & 1344 & 105.56 & 4 & 0 & 2 \\
 &  & 15 & 1344 & 110.53 & 4 & 0 & 2 \\
\cmidrule{2-8}
 & \multirow[t]{5}{*}{$\Preprocessing_3$} & 7 & 288 & 7.11 & 2 & 0 & 1 \\
 &  & 9 & 1344 & 105.30 & 4 & 0 & 2 \\
 &  & 11 & 1344 & 105.32 & 4 & 0 & 2 \\
 &  & 13 & 1344 & 104.50 & 4 & 0 & 2 \\
 &  & 15 & 1344 & 107.65 & 4 & 0 & 2 \\
\cmidrule{2-8}
 & \multirow[t]{5}{*}{$\Preprocessing_4$} & 7 & 288 & 7.23 & 2 & 0 & 1 \\
 &  & 9 & 1344 & 106 & 4 & 0 & 2 \\
 &  & 11 & 1344 & 105.10 & 4 & 0 & 2 \\
 &  & 13 & 1344 & 108.39 & 4 & 0 & 2 \\
 &  & 15 & 1344 & 107.51 & 4 & 0 & 2 \\
\cmidrule{1-8} \cmidrule{2-8}
\multirow[t]{20}{*}{logistics} & \multirow[t]{5}{*}{$\Preprocessing_1$} & 7 & 51 & 2.39 & 1 & 0 & 0 \\
 &  & 9 & 139 & 6.75 & 1 & 0 & 0 \\
 &  & 11 & 452 & 40.80 & 2 & 0 & 0 \\
 &  & 13 & 775 & 99.50 & 3 & 0 & 1 \\
 &  & 15 & 775 & 94.80 & 3 & 0 & 1 \\
\cmidrule{2-8}
 & \multirow[t]{5}{*}{$\Preprocessing_2$} & 7 & 51 & 1.98 & 1 & 0 & 0 \\
 &  & 9 & 139 & 4.79 & 1 & 0 & 0 \\
 &  & 11 & 452 & 28.73 & 2 & 0 & 0 \\
 &  & 13 & 775 & 48.67 & 2 & 0 & 0 \\
 &  & 15 & 775 & 50.10 & 2 & 0 & 0 \\
\cmidrule{2-8}
 & \multirow[t]{5}{*}{$\Preprocessing_3$} & 7 & 51 & 1.99 & 1 & 0 & 0 \\
 &  & 9 & 139 & 4.58 & 1 & 0 & 0 \\
 &  & 11 & 452 & 26.67 & 2 & 0 & 0 \\
 &  & 13 & 775 & 50.46 & 2 & 0 & 0 \\
 &  & 15 & 775 & 49.70 & 2 & 0 & 0 \\
\cmidrule{2-8}
 & \multirow[t]{5}{*}{$\Preprocessing_4$} & 7 & 51 & 1.92 & 1 & 0 & 0 \\
 &  & 9 & 139 & 4.55 & 1 & 0 & 0 \\
 &  & 11 & 452 & 28.17 & 2 & 0 & 0 \\
 &  & 13 & 775 & 50.08 & 2 & 0 & 0 \\
 &  & 15 & 775 & 47.35 & 2 & 0 & 0 \\
\cmidrule{1-8} \cmidrule{2-8}
\multirow[t]{20}{*}{miconic} & \multirow[t]{5}{*}{$\Preprocessing_1$} & 7 & 95.50 & 1.99 & 1 & 0 & 1 \\
 &  & 9 & 286.50 & 8.51 & 1.75 & 0.25 & 1.75 \\
 &  & 11 & 819.25 & 37.34 & 1.75 & 0.25 & 1.75 \\
 &  & 13 & 1752.50 & 130.23 & 1.75 & 0.25 & 1.75 \\
 &  & 15 & 1752.50 & 129.10 & 1.75 & 0.25 & 1.75 \\
\cmidrule{2-8}
 & \multirow[t]{5}{*}{$\Preprocessing_2$} & 7 & 95.50 & 1.66 & 1 & 0 & 0 \\
 &  & 9 & 286.50 & 7.17 & 1.75 & 0.25 & 0.75 \\
 &  & 11 & 819.25 & 29.87 & 1.75 & 0.25 & 1.75 \\
 &  & 13 & 1752.50 & 93.67 & 1.75 & 0.25 & 1.75 \\
 &  & 15 & 1752.50 & 92.27 & 1.75 & 0.25 & 1.75 \\
\cmidrule{2-8}
 & \multirow[t]{5}{*}{$\Preprocessing_3$} & 7 & 95.50 & 1.62 & 1 & 0 & 0 \\
 &  & 9 & 286.50 & 7.08 & 1.75 & 0.25 & 0.75 \\
 &  & 11 & 819.25 & 29.53 & 1.75 & 0.25 & 1.75 \\
 &  & 13 & 1752.50 & 93 & 1.75 & 0.25 & 1.75 \\
 &  & 15 & 1752.50 & 92.58 & 1.75 & 0.25 & 1.75 \\
\cmidrule{2-8}
 & \multirow[t]{5}{*}{$\Preprocessing_4$} & 7 & 95.50 & 1.91 & 1 & 0 & 0 \\
 &  & 9 & 286.50 & 7.19 & 1.75 & 0.25 & 0.75 \\
 &  & 11 & 819.25 & 30 & 1.75 & 0.25 & 1.75 \\
 &  & 13 & 1752.50 & 92.69 & 1.75 & 0.25 & 1.75 \\
 &  & 15 & 1752.50 & 93.46 & 1.75 & 0.25 & 1.75 \\
\cmidrule{1-8} \cmidrule{2-8}
\multirow[t]{20}{*}{newspapers} & \multirow[t]{5}{*}{$\Preprocessing_1$} & 7 & 133 & 7.43 & 3 & 2 & 2 \\
 &  & 9 & 229 & 14.11 & 3 & 2 & 2 \\
 &  & 11 & 316 & 20.28 & 3 & 2 & 2 \\
 &  & 13 & 382 & 26.34 & 3 & 2 & 2 \\
 &  & 15 & 406 & 28.47 & 3 & 1 & 2 \\
\cmidrule{2-8}
 & \multirow[t]{5}{*}{$\Preprocessing_2$} & 7 & 133 & 7.20 & 3 & 2 & 2 \\
 &  & 9 & 229 & 12.46 & 3 & 2 & 2 \\
 &  & 11 & 316 & 18.62 & 3 & 2 & 2 \\
 &  & 13 & 382 & 21.85 & 3 & 2 & 2 \\
 &  & 15 & 406 & 24.70 & 3 & 2 & 2 \\
\cmidrule{2-8}
 & \multirow[t]{5}{*}{$\Preprocessing_3$} & 7 & 133 & 7.09 & 3 & 2 & 1 \\
 &  & 9 & 229 & 12.90 & 3 & 2 & 2 \\
 &  & 11 & 316 & 18 & 3 & 2 & 2 \\
 &  & 13 & 382 & 22.04 & 3 & 2 & 2 \\
 &  & 15 & 406 & 24.24 & 3 & 2 & 2 \\
\cmidrule{2-8}
 & \multirow[t]{5}{*}{$\Preprocessing_4$} & 7 & 133 & 7.08 & 3 & 2 & 2 \\
 &  & 9 & 229 & 13.34 & 3 & 2 & 2 \\
 &  & 11 & 316 & 17.97 & 3 & 2 & 2 \\
 &  & 13 & 382 & 22.84 & 3 & 2 & 2 \\
 &  & 15 & 406 & 22.96 & 3 & 2 & 2 \\
\cmidrule{1-8} \cmidrule{2-8}
\multirow[t]{20}{*}{satellite} & \multirow[t]{5}{*}{$\Preprocessing_1$} & 7 & 298.25 & 2.09 & 0 & 0 & 0 \\
 &  & 9 & 1298.25 & 35.15 & 1 & 0 & 0 \\
 &  & 11 & 1502 & 45.20 & 1.25 & 0 & 0.25 \\
 &  & 13 & 1502 & 46.16 & 1.25 & 0 & 0.25 \\
 &  & 15 & 1502 & 45.14 & 1.25 & 0 & 0.25 \\
\cmidrule{2-8}
 & \multirow[t]{5}{*}{$\Preprocessing_2$} & 7 & 298.25 & 2.07 & 0 & 0 & 0 \\
 &  & 9 & 1298.25 & 32.69 & 1 & 0 & 0 \\
 &  & 11 & 1502 & 41.73 & 1.25 & 0 & 0.25 \\
 &  & 13 & 1502 & 41.94 & 1.25 & 0 & 0.25 \\
 &  & 15 & 1502 & 41.51 & 1.25 & 0 & 0.25 \\
\cmidrule{2-8}
 & \multirow[t]{5}{*}{$\Preprocessing_3$} & 7 & 298.25 & 2.03 & 0 & 0 & 0 \\
 &  & 9 & 1298.25 & 32.86 & 1 & 0 & 0 \\
 &  & 11 & 1502 & 41.04 & 1.25 & 0 & 0.25 \\
 &  & 13 & 1502 & 42.28 & 1.25 & 0 & 0.25 \\
 &  & 15 & 1502 & 41.76 & 1.25 & 0 & 0.25 \\
\cmidrule{2-8}
 & \multirow[t]{5}{*}{$\Preprocessing_4$} & 7 & 298.25 & 2.13 & 0 & 0 & 0 \\
 &  & 9 & 1298.25 & 33.12 & 1 & 0 & 0 \\
 &  & 11 & 1502 & 42.14 & 1.25 & 0 & 0.25 \\
 &  & 13 & 1502 & 42.34 & 1.25 & 0 & 0.25 \\
 &  & 15 & 1502 & 41.66 & 1.25 & 0 & 0.25 \\
\cmidrule{1-8} \cmidrule{2-8}
\multirow[t]{20}{*}{spanner} & \multirow[t]{5}{*}{$\Preprocessing_1$} & 7 & 139 & 1.40 & 0 & 0 & 0 \\
 &  & 9 & 416 & 8.59 & 1 & 0 & 1 \\
 &  & 11 & 1189.25 & 76.07 & 2 & 1 & 2 \\
 &  & 13 & 1977.50 & 201.33 & 2 & 0.75 & 1.75 \\
 &  & 15 & 1977.50 & 201.66 & 2 & 0.75 & 1.75 \\
\cmidrule{2-8}
 & \multirow[t]{5}{*}{$\Preprocessing_2$} & 7 & 139 & 1.23 & 0 & 0 & 0 \\
 &  & 9 & 416 & 7 & 1 & 0 & 1 \\
 &  & 11 & 1189.25 & 61.57 & 2 & 1 & 1 \\
 &  & 13 & 1977.50 & 163.30 & 2 & 1 & 1 \\
 &  & 15 & 1977.50 & 162.32 & 2 & 1 & 1 \\
\cmidrule{2-8}
 & \multirow[t]{5}{*}{$\Preprocessing_3$} & 7 & 139 & 1.02 & 0 & 0 & 0 \\
 &  & 9 & 416 & 6.81 & 1 & 0 & 1 \\
 &  & 11 & 1189.25 & 60.79 & 2 & 1 & 1 \\
 &  & 13 & 1977.50 & 163.78 & 2 & 0.75 & 1 \\
 &  & 15 & 1977.50 & 162.03 & 2 & 0.75 & 1 \\
\cmidrule{2-8}
 & \multirow[t]{5}{*}{$\Preprocessing_4$} & 7 & 139 & 1.27 & 0 & 0 & 0 \\
 &  & 9 & 416 & 6.84 & 1 & 0 & 1 \\
 &  & 11 & 1189.25 & 61.36 & 2 & 1 & 1 \\
 &  & 13 & 1977.50 & 162.75 & 2 & 1 & 1 \\
 &  & 15 & 1977.50 & 162.26 & 2 & 1 & 1 \\
\cmidrule{1-8} \cmidrule{2-8}
\multirow[t]{20}{*}{tpp} & \multirow[t]{5}{*}{$\Preprocessing_1$} & 7 & 68 & 5.01 & 1 & 0 & 0 \\
 &  & 9 & 153 & 11.76 & 1 & 0 & 0 \\
 &  & 11 & 372 & 59.45 & 2 & 0 & 0 \\
 &  & 13 & 863 & 175.04 & 2 & 0 & 0 \\
 &  & 15 & 1970 & 917.41 & 3 & 0 & 0 \\
\cmidrule{2-8}
 & \multirow[t]{5}{*}{$\Preprocessing_2$} & 7 & 68 & 4.18 & 1 & 0 & 0 \\
 &  & 9 & 153 & 8.97 & 1 & 0 & 0 \\
 &  & 11 & 372 & 49.14 & 2 & 0 & 0 \\
 &  & 13 & 863 & 148.70 & 2 & 0 & 0 \\
 &  & 15 & 1970 & 828.33 & 3 & 0 & 0 \\
\cmidrule{2-8}
 & \multirow[t]{5}{*}{$\Preprocessing_3$} & 7 & 68 & 4.07 & 1 & 0 & 0 \\
 &  & 9 & 153 & 9.15 & 1 & 0 & 0 \\
 &  & 11 & 372 & 49.06 & 2 & 0 & 0 \\
 &  & 13 & 863 & 149.94 & 2 & 0 & 0 \\
 &  & 15 & 1970 & 823.76 & 3 & 0 & 0 \\
\cmidrule{2-8}
 & \multirow[t]{5}{*}{$\Preprocessing_4$} & 7 & 68 & 3.97 & 1 & 0 & 0 \\
 &  & 9 & 153 & 9.23 & 1 & 0 & 0 \\
 &  & 11 & 372 & 49.48 & 2 & 0 & 0 \\
 &  & 13 & 863 & 148.82 & 2 & 0 & 0 \\
 &  & 15 & 1970 & 832.61 & 3 & 0 & 0 \\
\cmidrule{1-8} \cmidrule{2-8}
\multirow[t]{20}{*}{transport} & \multirow[t]{5}{*}{$\Preprocessing_1$} & 7 & 75.50 & 2.63 & 1 & 0 & 0.25 \\
 &  & 9 & 248 & 14.58 & 1.75 & 0 & 0.75 \\
 &  & 11 & 786.25 & 109.36 & 2.50 & 1 & 0.75 \\
 &  & 13 & 1455.25 & 312.57 & 2.50 & 1 & 1.50 \\
 &  & 15 & 1455.25 & 308.75 & 2.50 & 1 & 1.50 \\
\cmidrule{2-8}
 & \multirow[t]{5}{*}{$\Preprocessing_2$} & 7 & 75.50 & 2.46 & 1 & 0 & 0 \\
 &  & 9 & 248 & 13.42 & 1.75 & 0 & 0.50 \\
 &  & 11 & 786.25 & 99.51 & 2.50 & 1 & 0.75 \\
 &  & 13 & 1455.25 & 285.24 & 2.50 & 1 & 1.50 \\
 &  & 15 & 1455.25 & 284.04 & 2.50 & 1 & 1.50 \\
\cmidrule{2-8}
 & \multirow[t]{5}{*}{$\Preprocessing_3$} & 7 & 75.50 & 2.51 & 1 & 0 & 0 \\
 &  & 9 & 248 & 12.90 & 1.75 & 0 & 0.50 \\
 &  & 11 & 786.25 & 98.91 & 2.50 & 1 & 0.75 \\
 &  & 13 & 1455.25 & 277.41 & 2.50 & 1 & 1.50 \\
 &  & 15 & 1455.25 & 277.38 & 2.50 & 1 & 1.50 \\
\cmidrule{2-8}
 & \multirow[t]{5}{*}{$\Preprocessing_4$} & 7 & 75.50 & 2.69 & 1 & 0 & 0 \\
 &  & 9 & 248 & 13.48 & 1.75 & 0 & 0.50 \\
 &  & 11 & 786.25 & 101.12 & 2.50 & 1 & 0.75 \\
 &  & 13 & 1455.25 & 280.82 & 2.50 & 1 & 1.50 \\
 &  & 15 & 1455.25 & 280.66 & 2.50 & 1 & 1.50 \\
\cmidrule{1-8} \cmidrule{2-8}
\multirow[t]{20}{*}{visitall} & \multirow[t]{5}{*}{$\Preprocessing_1$} & 7 & 107 & 1.11 & 0 & 0 & 0 \\
 &  & 9 & 322 & 3.30 & 0 & 0 & 0 \\
 &  & 11 & 925 & 9.87 & 0 & 0 & 0 \\
 &  & 13 & 1359 & 17.13 & 0 & 0 & 0 \\
 &  & 15 & 1359 & 16.52 & 0 & 0 & 0 \\
\cmidrule{2-8}
 & \multirow[t]{5}{*}{$\Preprocessing_2$} & 7 & 107 & 1.11 & 0 & 0 & 0 \\
 &  & 9 & 322 & 2.76 & 0 & 0 & 0 \\
 &  & 11 & 925 & 8.83 & 0 & 0 & 0 \\
 &  & 13 & 1359 & 15.41 & 0 & 0 & 0 \\
 &  & 15 & 1359 & 15.61 & 0 & 0 & 0 \\
\cmidrule{2-8}
 & \multirow[t]{5}{*}{$\Preprocessing_3$} & 7 & 107 & 1.02 & 0 & 0 & 0 \\
 &  & 9 & 322 & 2.92 & 0 & 0 & 0 \\
 &  & 11 & 925 & 8.82 & 0 & 0 & 0 \\
 &  & 13 & 1359 & 14.98 & 0 & 0 & 0 \\
 &  & 15 & 1359 & 14.88 & 0 & 0 & 0 \\
\cmidrule{2-8}
 & \multirow[t]{5}{*}{$\Preprocessing_4$} & 7 & 107 & 1.11 & 0 & 0 & 0 \\
 &  & 9 & 322 & 2.88 & 0 & 0 & 0 \\
 &  & 11 & 925 & 9.01 & 0 & 0 & 0 \\
 &  & 13 & 1359 & 15.64 & 0 & 0 & 0 \\
 &  & 15 & 1359 & 18.88 & 0 & 0 & 0 \\
\cmidrule{1-8} \cmidrule{2-8}
\end{longtable}

\end{document}